\documentclass[letterpaper, oneside]{report}
\usepackage{import} 
\usepackage{thesis}
\usepackage{indentfirst}
\usepackage{tabularx}
\usepackage{threeparttable}
\usepackage{threeparttablex}
\usepackage{longtable}
\usepackage{float}
\usepackage{graphicx}
\usepackage{lmodern}
\usepackage{anyfontsize}
\usepackage{rotating}

\usepackage{cite}
\usepackage{amsmath, amssymb, amsfonts}
\usepackage{textcomp}
\usepackage{booktabs}

\usepackage{xcolor}
\usepackage{multirow}
\usepackage{soul}
\usepackage{times}
\usepackage{latexsym}
\usepackage[T1]{fontenc}
\usepackage[utf8]{inputenc}
\usepackage{microtype}
\usepackage{enumitem}
\usepackage{inconsolata}
\usepackage{pifont}

\usepackage{afterpage}
\usepackage{placeins}
\usepackage{adjustbox}
\usepackage{newunicodechar}
\newunicodechar{≥}{$\geq$}
\usepackage[framemethod=TikZ]{mdframed}
\mdfdefinestyle{mystyle}{width=\columnwidth}
\usepackage{multicol}
\usepackage[most]{tcolorbox}
\usepackage{hyperref}

\usepackage{seqsplit}
\usepackage{lineno}
\usepackage{array}
\usepackage{calc}

\newcommand{\blfootnote}[1]{%
  \begingroup
  \renewcommand\thefootnote{}\footnote{#1}%
  \addtocounter{footnote}{-1}%
  \endgroup
}

\newcommand{\thesisname}{New AI-Driven Tools for Enhancing Campus Well-being: A Prevention and Intervention Approach}
\newcommand{\authorname}{Jinwen Tang}
\newcommand{\advisorname}{Yi Shang}
\newcommand{\committeeone}{Dong Xu}
\newcommand{\committeetwo}{Toni Kazic}
\newcommand{\committeethree}{Victoria Shaffer}
\newcommand{\degreename}{Doctor of Philosophy}
\newcommand{\fulldegreename}{Doctor of Philosophy in Computer Science}
\newcommand{\university}{University of Missouri-Columbia}
\newcommand{\monthyear}{May. 2026}

\renewcommand{\listtablename}{LIST OF TABLES}
\renewcommand{\listfigurename}{LIST OF ILLUSTRATIONS}

\begin{document}
\maketitle

\clearpage
\thispagestyle{empty}
\begin{center}
    \vspace*{2cm}
    \textbf{Copyright} \\
    \vspace{1cm}
    © \monthyear\ \authorname \\
    All rights reserved.
\end{center}

\clearpage
\flushleft
The undersigned, appointed by the dean of the Graduate School, have examined the dissertation entitled
\vspace{1\baselineskip}
\begin{center}
    \MakeUppercase{\thesisname}
\end{center}
\vspace{1\baselineskip}
presented by \authorname,\\
a candidate for the degree of \fulldegreename,\\
and hereby certify that, in their opinion, it is worthy of acceptance.

\begin{center}
    \vspace{3\baselineskip}
    \rule{25em}{0.4pt}\\
    Professor \advisorname \\[2.5\baselineskip]
    \rule{25em}{0.4pt}\\
    Professor \committeeone \\[2.5\baselineskip]
    \rule{25em}{0.4pt}\\
    Professor \committeetwo \\[2.5\baselineskip]
    \rule{25em}{0.4pt}\\
    Professor \committeethree \\[2.5\baselineskip]
\end{center}

\vfill
\thispagestyle{empty} 

\chapter*{DEDICATION}
\thispagestyle{empty}

\indent\indent
To my parents, for their unconditional love and unwavering support through every frustrating moment of this journey. Your sacrifices and belief in me made this possible.
To my husband, Qiming Guo, for supporting me both in my academic pursuits and in daily life. Your patience and encouragement meant everything.
To my friends, for the laughter and companionship that made the difficult days easier to bear.
And to my cats, Pangyu, Bubble, and Sushi, for the emotional support and happiness you brought me throughout this process.

\clearpage
\thispagestyle{plain}
\setcounter{page}{2}
\chapter*{Acknowledgements}
\phantomsection
\addcontentsline{toc}{chapter}{ACKNOWLEDGEMENTS}
\setlength{\parindent}{2em}

First and foremost, I would like to express my deepest gratitude to my advisor, Professor Yi Shang, for his unwavering support and guidance throughout my doctoral journey. His mentorship in artificial intelligence, natural language processing, and machine learning has been instrumental in shaping my research. As an advisor, he provided me with countless opportunities to grow as a researcher, guided me through the process of conducting rigorous research, and helped me develop the skills to think critically and communicate my work effectively. My research accomplishments would not have been possible without his dedication and encouragement, and I am truly honored to have had him as my advisor.

I would also like to extend my sincere gratitude to my dissertation committee members, Professor Dong Xu, Professor Toni Kazic, and Professor Victoria Shaffer, for their time, thoughtful feedback, and valuable guidance throughout my doctoral program. Their insightful comments during my qualifying exam and dissertation defense greatly strengthened my work.

I would also like to express my appreciation to Professor Yunxin Zhao, whose guidance and support during the earlier stages of my doctoral program were invaluable. Though she has since retired, her contributions to my development as a researcher have left a lasting impact.

I would like to acknowledge my labmates for their collaboration and camaraderie over the years. The shared experiences, both professional and personal, have made this journey far more rewarding.

I would also like to acknowledge the use of AI language tools, including ChatGPT and Claude, which assisted in drafting and refining portions of this dissertation.

\clearpage
\tableofcontents

\clearpage
\phantomsection
\addcontentsline{toc}{chapter}{\listtablename}
\listoftables

\clearpage
\phantomsection
\addcontentsline{toc}{chapter}{\listfigurename}
\listoffigures

\clearpage

\begin{center}
    \MakeUppercase{\thesisname} \\[1cm]
    \authorname \\[0.5cm]
    Dr. \advisorname, Dissertation Supervisor \\[1cm]
    \textbf{ABSTRACT}
\end{center}
\phantomsection 
\addcontentsline{toc}{chapter}{ABSTRACT}
\vspace{0.5cm}
Campus well-being underpins the academic success and personal fulfillment of students, faculty, and staff, yet many universities lack effective methods for proactively monitoring satisfaction and detecting emerging mental health risks. These gaps highlight an urgent need for both preventive strategies, such as capturing nuanced, actionable feedback to improve the campus environment, and intervention tools that identify and support individuals experiencing psychological distress. By collecting richer data and personalizing engagement, institutions can more quickly and effectively respond to the evolving needs of their communities.

\hspace{2em}
Universities often contend with low response rates and superficial feedback in traditional campus climate surveys, limiting their ability to identify and address pressing issues. To overcome these shortcomings, we developed TigerGPT, a personalized survey chatbot, to engage users in deeper, context-aware conversations. Employing a theory-based engagement approach and leveraging large language models (LLMs), TigerGPT adapts its dialogues based on user roles and real-time inputs. Grounded in principles of conversational design and user engagement theory, it dynamically triggers and combines prompts to probe for deeper insights or clarifications, encourages comfortable sharing on sensitive topics, employs empathetic language and well-chosen emojis to foster trust and warmth, offers flexible topic selection to empower users, and incorporates visual elements such as bolded questions to reduce cognitive load. In a pilot study, TigerGPT demonstrated significant improvements over conventional methods, achieving a 75\% usability rating and an 81\% satisfaction score, with 50\% of participants preferring it over traditional surveys. 

\hspace{2em}
Despite these gains, the initial version suffered from repetitive prompts, limited topic diversity, superficial personalization, and low response quality. To address these issues, we introduced AURA, a reinforcement-learning framework that learns within a session to adapt follow-up question types using an LSDE quality signal (Length, Self-disclosure, Emotion, Specificity). AURA initializes expected gains from 96 prior campus-climate conversations (467 total chatbot-user exchanges) and updates them online via an $\epsilon$-greedy policy over 10-15 exchanges, selecting among follow-up classes (e.g., validate, specify, reflect, probe) to optimize LSDE as the conversation unfolds. In controlled evaluations ($n=20$ conversations per condition), AURA achieved a $+0.12$ mean gain in composite response quality with a statistically significant improvement over non-adaptive baselines ($p=0.044$, $d=0.66$), driven by a $63\%$ reduction in specification prompts and a $10\times$ increase in validation behavior. These results indicate that within-session reinforcement learning reduces repetition, increases response specificity and depth, and mitigates late-phase drop-off across heterogeneous campus users.

\hspace{2em}
While these enhancements substantially bolster campus feedback collection (Objective~1), they also underscore the need for specialized methods to detect and address mental health risks in a timely fashion (Objective~2).
Current approaches to early mental health detection remain fragmented and often rely on short-text analysis or static screenings, overlooking deeper emotional and cognitive markers. To address this gap, we focus on Expressive Narrative Stories (ENS), longer, first-person narratives rich in psychological details, to uncover subtle signals that brief, generic posts typically miss. Our methodology evaluates advanced language models (BERT and MentalBERT) alongside traditional classifiers (SVM, Naive Bayes, Logistic Regression) by removing or replacing topic words and shuffling sentence order. Results reveal that conventional classifiers and MentalBERT depend heavily on explicit mental health terms (P-value < 0.05), which can be problematic when such terms are absent. In contrast, BERT(128) maintains robust accuracy even without keyword cues, demonstrating its effectiveness at capturing nuanced linguistic features. Both BERT and MentalBERT remain resilient under sentence shuffling (P-value < 0.05), a critical capability for analyzing extended narratives, particularly in ENS-based comparisons of individuals with and without self-declared mental health issues. These findings underscore ENS as a valuable resource for more nuanced mental health screening, highlighting how context-aware analyses can deepen our understanding of linguistic patterns in psychological distress.

\hspace{2em}
Although Expressive Narrative Stories (ENS) have improved classification in mental health screening, existing approaches often function as “black boxes,” providing minimal transparency into how they derive conclusions. This lack of explainability can undermine user trust and limit clinical validation. To address this challenge, we developed “PsychoGPT”, a specialized large language model tailored for psychological distress assessment. Built atop GPT-4 with adaptations from DSM-5 and PHQ-8 guidelines, PsychoGPT follows a multi-stage evaluation process: it first produces an initial classification of potential distress, then breaks down PHQ-8 criteria to score individual symptoms, and finally reconciles its own findings with external PHQ ratings for an independent validation step. This staged approach not only yields more granular insights but also enables stakeholders to review the system’s reasoning line by line. Tested on datasets like DAIC-WOZ, PsychoGPT maintains high accuracy while generating clear, interpretable justifications for its decisions, thereby bridging the explainability gap in automated mental health screenings.

\hspace{2em}
While PsychoGPT enhances explainability, large language models can still produce inconsistent or erroneous outputs, particularly with extended, narrative-style data where “hallucinations” pose significant risks for mental health assessments. To address this, we developed a Stacked Multi-Model Reasoning (SMMR) framework, which layers multiple “expert” models to iteratively refine each other’s outputs, rather than relying on a single GPT-based analysis. Early layers tackle simpler or more localized subtasks, while subsequent layers integrate and reconcile these partial findings. By systematically resolving discrepancies between models and consolidating complementary insights, SMMR reduces hallucinations and stabilizes results. Evaluations on the DAIC-WOZ dataset and curated case study narratives demonstrate that SMMR outperforms single-model GPT solutions in accuracy, F1-score, and PHQ-8 scoring accuracy. These improvements mark a key step toward more robust, reliable, and safer AI-driven mental health interventions in campus contexts.

\hspace{2em}
To unify the preventive strengths of TigerGPT and AURA with the advanced mental health detection tools, we propose a cohesive framework that seamlessly connects real-time survey feedback with targeted intervention analytics (Objective~3). This design allows insights from survey chatbot’s adaptive dialogues, such as emerging student concerns or early signs of distress, to flow directly into specialized models like PsychoGPT or SMMR. By aligning both preventive and intervention strategies under one integrated system, administrators can simultaneously enhance campus-wide satisfaction initiatives and more swiftly identify and support individuals experiencing psychological challenges.

\mainbody

\setlength{\parindent}{0.5in} 

\chapter{Introduction}
\label{chap:intro}
\section{Motivation \& Problem Statement}
Campus well-being is a cornerstone of academic success and personal fulfillment for students, faculty, and staff \cite{vogel2008assessment}. Universities dedicate substantial resources to fostering environments that support academic achievement, social engagement, and mental health. However, traditional methods for assessing and improving well-being frequently fall short, leaving gaps that hinder timely and effective action. This dissertation explores how Artificial Intelligence (AI) can address these shortcomings, offering innovative tools for both prevention and intervention to enhance campus well-being.

\subsection{The Importance of Campus Well-Being}

A thriving campus community depends on the satisfaction and mental health of its members. Satisfaction with campus life, including facilities, social opportunities, and academic support, can reduce stress and foster resilience. Meanwhile, robust mental health support ensures that people can fully participate in university life. These dimensions are interconnected: Enhancing satisfaction can prevent mental health challenges, while proactive mental health interventions can improve overall well-being. However, despite their efforts, universities struggle to monitor and address these needs effectively due to limitations in current approaches.

\subsection{Limitations of Traditional Methods}
Traditional methods for assessing campus well-being, such as annual or biannual climate surveys, are plagued by low response rates and insufficient detail \cite{dillman2014internet, galesic2009effects}. These surveys, often lengthy and static, fail to capture the dynamic nature of campus experiences, providing administrators with outdated or incomplete data \cite{galesic2009effects}. For example, response rates for such surveys can dip below 20\%, limiting their representativeness and utility \cite{sahlqvist2011effect}.

On the mental health front, services are predominantly reactive, intervening only after problems escalate \cite{andrews2001shortfall}. With around 50\% of college students reporting overwhelming anxiety each year \cite{blanco2008mental}, this approach is inadequate. Resource constraints, stigma, and reluctance to seek help further exacerbate delays in support \cite{gulliver2010perceived, pescosolido1999people}.

\subsection{AI as a Transformative Solution}
AI offers a powerful means to overcome these limitations through real-time data collection and early detection capabilities. Conversational AI tools, such as chatbots, can engage users in adaptive, empathetic dialogues, collecting nuanced feedback on user experiences \cite{zhou2025building}. Unlike traditional surveys, these tools can ask natural follow-up questions, yielding richer insights and higher engagement \cite{xiao2020tell, tanwar2024opinebot}. Simultaneously, AI-driven analytics can process large datasets—such as social media posts or clinical transcripts—to identify early signs of mental distress \cite{tang2024decoding, tang2024advancing}.

\section{Research Objectives}
The primary aim of this dissertation is to harness Artificial Intelligence (AI) to enhance campus well-being by addressing two key areas: improving student satisfaction (prevention) and supporting mental health (intervention). This research integrates completed and proposed work to develop a comprehensive, AI-driven system that proactively tackles challenges in higher education environments. The specific objectives are as follows:

\textbf{Objective 1: Design, evaluate, and enhance a new conversational AI chatbot to improve campus satisfaction.}
Traditional campus surveys often suffer from low engagement and delayed insights. This objective centers on developing and refining \textit{TigerGPT}, a conversational survey baseline that collects real-time, detailed feedback; in a pilot it achieved a 75\% usability rating and an 81\% satisfaction score. Building on the TigerGPT baseline, we developed \textit{AURA}, an RL-based adaptive conversational survey that uses LSDE feedback and an $\epsilon$-greedy policy with online expected-gain updates across 10--15 exchanges. In controlled tests, AURA improved response quality ($+0.076$) with a significant advantage over non-adaptive baselines.

\textbf{Objective 2: Develop new AI-driven tools for mental health assessment using diverse data sources.}
Early identification of mental health issues is critical for effective intervention. This objective centers on creating AI tools to detect distress indicators (e.g., anxiety, depression) by analyzing linguistic patterns in social media posts (e.g., Reddit Expressive Narrative Stories) and narrative transcripts (e.g., DAIC-WOZ dataset). Completed tools like Decoding, Psycho Analyst, and SMMR leverage advanced models (e.g., BERT, GPT-4, multi-expert frameworks) to achieve high accuracy and reliability, offering scalable solutions for campus mental health monitoring.

\textbf{Objective 3: Propose a new framework for integrating AI tools into campus well-being services.}
To maximize impact, AI tools must be deployed responsibly and systematically. This objective involves designing a framework to integrate the chatbot and mental health assessment tools into campus services. 

These objectives collectively establish a holistic, AI-powered approach to enhancing campus satisfaction and mental health, bridging prevention and intervention in higher education.

\section{Contributions}
This dissertation advances the fields of conversational AI, mental health analytics, and campus well-being by introducing innovative tools and methodologies. It integrates completed work—demonstrating practical applications—with planned advancements to maximize future impact. These contributions align with the research objectives and tackle longstanding challenges in academic environments. The key contributions are detailed below:

\subsection{Completed Work}
\begin{itemize}
    \item \textbf{New Survey Chatbot for Campus Satisfaction}
    \begin{enumerate}
    \item \textbf{TigerGPT: A New Survey Chatbot for Campus Satisfaction \cite{tang2025tigergpt}:} We developed TigerGPT, a conversational AI chatbot powered by ChatGPT-turbo, was developed to collect real-time feedback on campus experiences. In a pilot study, it achieved a 75\% usability rating and an 81\% satisfaction score among users, outperforming traditional survey methods in engagement and response quality. Its adaptive dialogue system enables personalized interactions, delivering richer insights into campus satisfaction.
    \item \textbf{AURA: A New Survey Chatbot Framework Based on Adaptive Understanding through Reinforcement Learning for Assessment \cite{tang2025aura}:}
    We developed AURA, an RL-based enhancement that adapts \emph{within} each conversation by updating follow-up question-type selection after each exchange. AURA uses a four-dimensional LSDE signal (Length, Specificity, Self-disclosure, and Emotion) and an $\epsilon$-greedy policy with online expected-gain updates across 10--15 exchanges, initialized with priors from 96 campus-climate conversations (467 exchanges). In controlled evaluations ($n{=}20$ per condition), AURA improved response quality by $+0.076$ on average and showed a statistically significant advantage over non-adaptive baselines ($p{=}0.044$, $d{=}0.66$), alongside meaningful behavioral shifts (fewer specification prompts; more validation). These results establish AURA as the finalized enhancement for Objective~1, with TigerGPT retained as a baseline for comparison.

    \end{enumerate}
    \item \textbf{New Early Detection Tools for Mental Health}
    \begin{enumerate}
        \item \textbf{New Method to Decode Linguistic Nuances in Mental Health Text Classification Using Expressive Narrative Stories \cite{tang2024decoding}:} This study investigates the application of BERT, MentalBert over traditional models for identifying mental state indicators in Expressive Narrative Stories (ENS) from Reddit. Highlights BERT’s superior understanding of context and logical connections, significantly affecting classification performance. The study emphasizes ENS’s value in mental health analysis, advancing context-aware text classification methods.
        \item \textbf{Advancing Mental Health Pre-Screening: A New Custom GPT for Psychological Distress Assessment \cite{tang2024advancing}:} This study introduces ”Psycho Analyst,” a GPT-4-based model optimized for mental health pre-screening. Enhanced with DSM-5 and PHQ-8, it demonstrates high accuracy in detecting mental disorders, evidenced by F1 and Macro-F1 scores of 0.929 and 0.949. Validated with the DAIC-WOZ dataset, it shows potential for improving mental health diagnostics with precise, cost-effective, and accessible AI tools.
        \item \textbf{A New Layered Multi-Expert Framework for Long-Context Mental Health Assessments \cite{tang2025layered}:} This study presents the ”Stacked Multi-Model Reasoning (SMMR)” framework, addressing long-context challenges in mental health assessments with LLMs. Utilizing a multi-layered approach, SMMR combines several LLMs and specialized models as equal ”experts” to process complex, domain-specific data. The framework begins with isolating discrete subtasks and then integrates these findings through advanced models. Tested with the DAIC-WOZ dataset and 48 psychiatric case studies, SMMR significantly outperforms single-model methods, improving accuracy, F1-score, and reducing PHQ-8 errors, thus enhancing the reliability of mental health assessments

    \end{enumerate}
\end{itemize}

\section{Dissertation Organization}
This dissertation is organized as follows: Background (Chapter~\ref{chap:background}), Preventive Methodologies (Chapter~\ref{chap:preventive}), Intervention Methodologies (Chapter~\ref{chap:intervention}), Evaluation and Synthesis
(Chapter~\ref{chap:evaluation}), A Proposed Framework for Objective 3 (Chapter~\ref{chap:object3}, and Discussion and Future Work (Chapter~\ref{chap:discussion}).

\chapter{Background and Related Work}
\label{chap:background}

\section{Chatbot-Based Approaches for Campus Satisfaction Surveys}
\label{sec:rw-chatbots}

Campus satisfaction assessment has increasingly moved from infrequent, static questionnaires toward interactive conversational systems that can elicit richer, context-specific feedback. Traditional web surveys remain valuable for breadth and comparability, but they often suffer from low participation, limited open-ended elaboration, and long feedback cycles that delay remediation \cite{dillman2014internet, galesic2009effects, sahlqvist2011effect}. In contrast, chatbot-based interfaces can invite follow-up clarification in natural language, reduce friction to participation, and surface actionable detail about services, courses, and campus climate \cite{abbas2021university, belhaj2021engaging}. For administrators, the potential benefit is not only higher volume but also higher \emph{quality} of qualitative evidence that informs timely interventions in areas such as advising, facilities, and mental health.

\subsection{Chatbots in Higher Education}
\label{sec:rw-he-chatbots}
AI-powered chatbots are widely deployed for student services, including enrollment support, financial aid navigation, IT troubleshooting, and course planning \cite{abbas2022online, martinez2024ai}. Reports from university pilots describe improved responsiveness, reduced staff workload, and more individualized guidance when conversational systems are integrated with campus workflows \cite{martinez2024ai, liu2024teaching}. Beyond transactional assistance, chat-based interactions can make institutional communication feel more approachable than formal forms or complex web portals, which lowers the barrier for students to ask sensitive questions or report problems \cite{abbas2021university}. At the same time, studies note heterogeneous acceptance driven by privacy, trust, and transparency concerns \cite{njeguvs2021conversational, zarouali2024comparing}, underscoring the need for clear disclosures, opt-out paths, and human review channels when appropriate.

\subsection{Conversational Surveys and Chatbot-Based Data Collection}
\label{sec:rw-conv-surveys}
Conversational survey systems present prompts in natural language and use follow-ups to clarify or probe in real time \cite{xiao2020tell, tanwar2024opinebot}. Compared with conventional web forms, multiple studies report higher engagement, longer answers, and more emotionally expressive or specific content \cite{kim2019comparing, abbas2021university, belhaj2021engaging}. These gains are typically attributed to immediacy, perceived social presence, and the opportunity to elaborate when the agent asks a targeted follow-up. Acceptance is not uniform, however. Some respondents prefer conventional surveys because they feel more private, predictable, or efficient for short answers \cite{njeguvs2021conversational, zarouali2024comparing}. System design therefore matters: tone, transparency about data use, escalation options, and the intrusiveness of follow-ups all influence willingness to engage.

\subsection{Non-Adaptive and Heuristically Adaptive Baselines}
\label{sec:rw-nonadaptive}

\textbf{Rule-based chatbots.}
Early conversational survey systems were predominantly rule-based, using dialogue trees, frame/slot-filling pipelines, agenda-based state machines, and keyword/pattern triggers \cite{thorat2020review, maeng2021designing, chan2022challenges}. These architectures offer predictable flows and low runtime cost and can branch by demographics or topic keywords to appear personalized. In practice, however, they rely on hand-authored prompts and finite transitions, which limits coverage when inputs are unexpected or multi-intent. Recovery often falls back to generic clarifications, yielding repetitive prompts, premature topic shifts, and early disengagement in open-ended surveys.

\textbf{LLM-mediated, heuristically adaptive systems.}
Recent deployments replace pattern matching with large language models to improve fluency, topical coverage, and empathy cues in survey-style conversations \cite{xiao2020tell, tanwar2024opinebot}. Follow-ups are typically selected from curated prompt banks or few-shot instructions, sometimes conditioned on role, topic, or prior answers—providing \emph{heuristic adaptivity} without explicit online learning. This approach increases perceived personalization over rule-based designs, yet follow-up selection remains reactive at the exchange level rather than optimized from cumulative, within-session quality feedback.

\textbf{TigerGPT as a practical heuristically adaptive baseline.}
Our prior work, \textit{TigerGPT} \cite{tang2025tigergpt}, exemplifies this class: an empathetic, LLM-mediated conversational survey that increased engagement and perceived personalization in pilots (via user-driven topic selection and supportive tone), outperforming static web forms. TigerGPT employs design heuristics and role/topic conditioning to adapt its prompts, but it does not update its questioning strategy via \emph{within-session learning} from real-time quality signals.

\textbf{Motivation for within-session learning.}
Open-ended campus-climate conversations are short (about 10--15 exchanges) and heterogeneous across respondents. Static rules and heuristic adaptivity struggle to sustain depth and disclosure across this diversity. These limitations motivate an approach that can update follow-up selection during the conversation based on real-time quality indicators, which we address with AURA in later sections.

\subsection{What Makes Conversational Responses “High Quality”}
\label{sec:rw-quality}
Survey methodologists emphasize that open-ended response quality is multidimensional rather than captured by a single indicator. Common dimensions include response length and elaboration \cite{dillman2014internet}, emotional expression and engagement \cite{porter2005mail, tausczik2010psychological}, and the specificity or concreteness of content \cite{brysbaert2014concreteness}. These factors correlate with downstream utility for decision making because they reflect not only the volume of text but also the depth, relevance, and personal context of the information provided. In conversational settings, such dimensions can be operationalized through lightweight heuristics or model-based scoring to yield per-exchange signals suitable for online control in short sessions. This literature motivates our use of a four-dimensional signal that captures Length, Self-disclosure, Emotion, and Specificity as the basis for adaptive questioning (details deferred to Section~\ref{sec:AURA}).

\subsection{Individual Differences and Heterogeneous Preferences}
\label{sec:rw-individual}
Respondents differ widely in disclosure comfort, preferred communication style, and responsiveness to follow-up strategies \cite{wheeless1977measurement, gudykunst1988culture, pennebaker1999linguistic}. The same prompt can elicit detailed narratives from one person and minimal replies from another, and preferences may shift within a single conversation as topics change or cognitive load fluctuates \cite{tourangeau2000psychology, lenzner2010cognitive, melville2016conducting}. For example, some individuals respond well to specification prompts that request concrete examples, while others disengage unless the agent first acknowledges emotions or validates concerns. One-size-fits-all heuristics are therefore brittle in practice, especially for sensitive topics related to campus well-being and inclusion. Systems need mechanisms to infer, in real time, which questioning strategies are likely to deepen engagement for the current respondent.

\subsection{Adaptive Dialogue Control: Reinforcement Learning}
\label{sec:rw-rl}
Reinforcement learning (RL) and contextual bandits provide a principled framework for optimizing sequential decision making under uncertainty \cite{sutton1998reinforcement}. In dialogue management, these approaches have been used to learn policies for action selection, initiative management, and question sequencing \cite{su2021dynamic, sun2023contextual}. Survey applications impose distinctive constraints that shape algorithm design. First, conversations are short, typically 10--15 exchanges, which limits the horizon and constrains exploration. Second, action spaces should remain small and interpretable, such as a set of follow-up \emph{question types} rather than arbitrary utterances, to enable analysis, auditing, and safe operation. Third, updates must be stable and low-overhead during live interactions, which favors simple value-estimation or bandit-style methods over deep, sample-hungry policies in this setting. Finally, institutional constraints around privacy and risk make \emph{within-session} adaptation preferable to cross-user personalization; the agent should react to the present respondent rather than pooling sensitive signals across many users.

\subsection{Safety, Ethics, and Human Oversight}
\label{sec:rw-ethics}
Conversational systems that collect well-being and climate feedback must adhere to clear safety and ethics guardrails. Prior work recommends transparent system role descriptions, scope limitations, opt-outs, and escalation resources when sensitive disclosures arise \cite{njeguvs2021conversational}. Tone control and empathetic acknowledgments can improve perceived respect and reduce drop-off, but they should not be conflated with clinical guidance. Human-in-the-loop review may be warranted for flagged content, and data retention policies should be clearly communicated. These considerations influence both the conversational design and the choice of adaptive algorithms, since aggressive probing may increase short-term elaboration at the expense of trust.

\subsection{Positioning of TigerGPT and AURA}
\label{sec:rw-positioning}
Within this landscape, \textit{TigerGPT} serves as a practical, heuristically adaptive baseline that demonstrates the usability benefits of LLM-mediated survey interfaces for campus satisfaction assessment \cite{tang2025tigergpt}. However, its fixed heuristics do not adjust follow-ups based on observed engagement within a session. The \textit{AURA} framework addresses this gap by performing \emph{within-session} adaptation: after each exchange (one round of conversation between user and chatbot), the system computes a multidimensional quality signal and selects the next question type to balance specificity, disclosure, and rapport for the current respondent \cite{tang2025aura}. AURA’s policy learning approach is designed for short horizons, interpretable actions, and low-overhead online updates that fit the constraints of campus survey deployments. Methodological details appear in Section~\ref{sec:AURA}, and comparative evaluations against TigerGPT baselines are presented in Section~\ref{sec:AURA_results}.

\section{AI in Mental Health Applications}
Mental health remains a major concern in public health, especially in university settings where academic pressures, financial worries, and social challenges can intensify conditions like depression or anxiety \cite{gulliver2010perceived, pescosolido1999people}. As noted in Chapter~\ref{chap:intro}, neglecting these issues can harm not only individuals’ well-being but also overall campus satisfaction and retention. However, many people delay seeking professional help due to stigma or limited clinical services \cite{andrews2001shortfall, regier1993facto}, highlighting the need for scalable, proactive interventions.

Recent advances in artificial intelligence (AI) aim to provide more discreet and accessible mental health support \cite{openai2023gpt}. Rather than waiting until someone is in crisis or navigating overburdened counseling offices, AI systems can identify early signs of distress. In a university setting, these tools might be woven into chatbots, anonymous surveys, or social media screening, helping detect people who need assistance sooner. Yet incorporating AI into mental health also raises concerns about data privacy, ethical use of personal information, and the reliability of machine-generated evaluations—challenges that become especially salient when the stakes involve mental health diagnoses or support referrals \cite{tangshang2024advancing, ohse2024zero}.

\subsection{Early Machine Learning Approaches and Short-Text Analysis}
Before large language models gained prominence, many researchers used simpler natural language processing (NLP) and machine learning methods—such as sentiment analysis, keyword spotting, or bag-of-words—to spot mental health indicators in short posts. These techniques showed moderate success for conditions like depression or anxiety, especially on platforms like Twitter or mini-blogs \cite{tausczik2010psychological, wang2013depression}, with traditional classifiers (e.g., support vector machines, logistic regression) reaching reasonable accuracy \cite{yates2017depression, yazdavar2017semi}.

They also extended to specialized datasets, such as the Distress Analysis Interview Corpus (DAIC-WOZ), which connects classification outputs to clinical metrics like PHQ-8 \cite{gratch2014distress, devault2014simsensei, ringeval2019avec}. For example, Belser et al. applied BGRU, HAN, and Transformers for depression detection in DAIC-WOZ, whereas Yadav et al. demonstrated a BGRU-based model surpassing earlier methods \cite{belser2023comparison, yadav2023novel}. Milintsevich et al. similarly employed hierarchical regression to predict depression symptom severity \cite{milintsevich2023towards}. Despite these achievements, many early approaches focused on short text fragments and missed deeper linguistic cues present in lengthier personal accounts or multi-turn interviews \cite{huang2023survey, dahl2024large}.

\subsection{Clinical Records vs. Social Media and the Challenge of Subjectivity}
Traditionally, researchers relied on clinical sources such as Electronic Health Records (EHRs) and physician notes \cite{le2021machine, tran2017predicting}. While these records are often more structured and tie directly to professional diagnoses, privacy laws make them hard to access, and they may not capture day-to-day emotional changes. In contrast, social media content (e.g., Reddit, Weibo) offers abundant, informal writing from people describing experiences in real time \cite{zhang2022natural, graham2019artificial, shatte2019machine}, but it rarely includes official clinical labels.

Nonetheless, user-generated text can provide unique clues about how individuals perceive and express their distress, albeit in ways that may be unstructured or highly context-dependent. Instruments like the PHQ-8 establish diagnostic guidelines, yet real-life descriptions of depression or anxiety vary substantially by personal, cultural, and situational factors \cite{ng2019provider, Marsolek20240969PD}. Interpreting these subtleties calls for more advanced methods that can uncover subtle cues, incomplete mentions, or culturally specific references to psychological struggle.

\subsection{Expressive Narrative Stories and Deeper Linguistic Nuances}
Early research in AI-driven mental health often centered on short social media updates. While valuable for large-scale scanning, these methods mostly overlooked Expressive Narrative Stories (ENS)—longer, first-person accounts of life events and internal reflections that can uncover more of a person’s emotional trajectory \cite{cavazos2016content, pennebaker2003psychological, hart2001redeveloping}. Building on Pennebaker’s idea of expressive writing \cite{pennebaker1986confronting}, ENS can reveal hidden signs of distress, including shifts in language, repeated negative themes, or changing self-perception \cite{pennebaker1988disclosure, davidson2002expressive, rivkin2006effects}.

Despite their potential, many current models still concentrate on short, isolated posts instead of capturing the evolving arcs and emotional subtexts found in ENS \cite{yates2017depression, inamdar2023machine}. For instance, a simple classifier might highlight words like “sad” or “lonely” but miss a whole narrative describing guilt, insomnia, or other factors linked to depression. Studies indicate that analyzing entire narratives often yields more accurate mental health cues \cite{wang2013depression, yazdavar2017semi}, but capitalizing on longer text requires robust models that handle extended context and emotional nuances \cite{huang2023survey, dahl2024large}.

\subsection{LLM-Based Solutions for Mental Health}
Large language models (LLMs) such as GPT-3.5 or GPT-4 have reshaped how text-based data is processed, generated, and summarized across various healthcare contexts \cite{garg2023exploring}. In clinical settings, these models can serve as “virtual assistants” that help with drafting medical notes, synthesizing large volumes of research literature, or responding to patient inquiries in near real-time. This functionality holds particular appeal in high-demand environments—like busy campuses or hospitals—where staff may be overwhelmed and need additional decision support.

\textbf{Domain-Adaptive and Specialized Models:} A key strategy for integrating LLMs into healthcare involves tailoring them to specific medical tasks through fine-tuning or domain adaptation. For example, models like Med-PaLM 2 have been trained on large bodies of medical knowledge, enabling them to handle complex exam questions and provide improved answers to consumer health queries \cite{singhal2025toward}. Similarly, a study by Van et al. \cite{van2023clinical} adapted GPT-style architectures to summarize clinical text, including radiology reports and Q\&A threads, sometimes outperforming human experts in both accuracy and clarity. These results underline how careful customization can transform a general-purpose LLM into a specialized “consultant” that condenses dense medical content into concise, actionable insights.

\textbf{Patient-Facing Chatbots and Mental Health Support:} Another growing area is using LLMs to power patient-facing chatbots. Some applications allow users to ask everyday health questions, receive medication tips, or even engage in guided self-care exercises in a conversational format \cite{garg2023exploring}. In mental health, custom GPT-based models have been explored for tasks like early detection of depression, anxiety, or suicidal ideation, scanning written posts or chat logs for risk factors \cite{tangshang2024advancing, agrawal2024illuminate, guo2024soullmate, guo2024soullmate2}. Researchers have also tested AI-driven “therapy-like” dialogues, where an LLM simulates basic counseling approaches—such as reframing negative thoughts or offering supportive language—though these remain experimental and should not replace professional therapy \cite{guo2024large}.

\textbf{Summarization and Triage:} LLMs can further assist with triaging mental health care, particularly in large communities like universities where counseling resources may be limited. For instance, the model might scan a user’s posts (from forums, campus chatbots, or help lines), summarize the main issues (e.g., stress about academic performance, sleep problems, isolation), and flag them for follow-up if certain severity thresholds are met \cite{guo2024large, ng2019provider}. By aligning these summarization features with recognized standards like the Diagnostic and Statistical Manual of Mental Disorders (DSM-5) or the PHQ-8, custom GPT systems can offer more structured insight to clinicians or campus wellness teams.

Despite these advancements, LLMs still face substantial hurdles when dealing with subjective, multi-turn narratives common in mental health settings \cite{huang2023survey, dahl2024large}. Even powerful models can produce “hallucinations”—confident but false statements—if they lack contextual grounding or domain awareness. Ethical concerns—such as privacy, consent, and data ownership—are magnified in mental health contexts, where personal expression may reveal sensitive details. These issues become especially critical for campus-based tools, which might monitor social media or survey data without direct clinical oversight.

Hence, while large language models hold promise for improving mental health screenings and interventions, practical implementations must address data handling, transparency, and human oversight. In campus environments, combining LLM-driven chatbots or screening systems with robust referral processes ensures that AI can complement (rather than replace) human professionals, ultimately broadening access to mental health resources and helping users feel more supported throughout their academic journey.

\subsection{Multi-Expert Frameworks for Complex, Long-Context Data}
Mental health evaluations often involve lengthy or multi-session content—such as extended personal narratives or multi-turn interviews—which can overwhelm a single large language model. Although GPT-4 and similar architectures handle more context than previous models, three major hurdles persist: (1) mental health data remain subjective, complicating symptom interpretation even with standardized tools (e.g., PHQ-8) \cite{ng2019provider, Marsolek20240969PD}; (2) long transcripts can exceed a model’s capacity, causing hallucinations or contradictions \cite{huang2023survey, dahl2024large}; and (3) specialized instruments only partially address these gaps, as a single LLM “expert” might overlook important details unless it repeatedly re-checks or has deep domain knowledge \cite{tangshang2024advancing, ohse2024zero}.

A growing body of research recommends multi-expert or layered frameworks to overcome these obstacles. Rather than relying on one large model, tasks are divided among specialized experts—some focusing on self-harm detection, others on local sentiment shifts, and so on. Outputs from these modules feed into an advanced aggregator (or “meta-LLM”) that integrates partial findings into a coherent conclusion. For example, Tang et al. \cite{tang2025layered} designed a pipeline for long-form mental health assessments by combining lighter-weight classifiers to flag high-risk phrases with a more advanced LLM that synthesizes these signals. This setup lowers the chance of missing a fleeting mention of despair in a multi-page transcript and provides a built-in cross-check system among experts.

Song et al. \cite{song2024combining} similarly rely on hierarchical variational autoencoders (VAEs) to summarize a user’s lengthy social media timeline, then prompt a clinically oriented LLM for “higher-level” analysis—capturing mood deterioration or depressive episodes with greater accuracy. Meanwhile, Soman et al. \cite{soman2025human} explore retrieval-augmented generation, surfacing only relevant conversation segments to avoid overwhelming a model with the entire dialogue history. These multi-expert pipelines scale more effectively for large data, while also improving interpretability: each specialized module’s outputs are plainly visible, crucial for clinical validation where oversights could mean missing a subtle but dangerous shift in someone’s mental state. Xu et al. \cite{xu2024mental} even show that a targeted classifier can outperform GPT-4 in suicide risk detection, emphasizing the benefits of domain-focused solutions.

Overall, multi-expert frameworks offer a safer, more accurate path for analyzing complex or extended mental health data. They share tasks across specialized parts, then merge those insights through a final integrator, reducing hallucinations and enhancing diagnostic rigor. As campuses and clinics increasingly adopt AI-driven screening, these layered architectures stand out as a robust design blueprint for continuous monitoring, early detection, and, ultimately, more informed and empathetic mental health interventions.

\section{Gaps in Campus Well-being Research}
Previously, campus well-being research faced several notable challenges across both student satisfaction assessment and mental health detection. Researchers exploring chatbots or text-based analytics often did so in isolation, leading to the following gaps:

\begin{enumerate}
\item \textbf{Low Engagement and Limited Personalization in Campus Surveys:}
Traditional survey methods—whether web-based questionnaires or static chatbots—struggled with low response rates and minimal follow-up. These tools could not adapt dynamically to user feedback, nor did they sustain participants’ interest over multiple questions. As a result, institutions lacked timely, in-depth insights into evolving campus issues.

\item \textbf{Absence of Within-Session Adaptation in Conversational Surveys:}
Conventional online surveys offer little personalization, leading to low engagement and shallow responses. Although AI chatbots improve convenience and raise usability and perceived personalization, they still rely on fixed dialogue trees or static prompt templates and cannot adapt within a session to an individual respondent, which produces generic follow-ups and leads to lower response quality (especially in short sessions).

\item \textbf{Short-Text Bias in Mental Health Classification:}
Early mental health detection studies (e.g., analyzing social media) often relied on brief user posts or short self-reports. While practical for large-scale scanning, these approaches did not fully capture the nuanced emotional or contextual cues found in more expressive narratives. Consequently, signs of distress or subtle mental health indicators could be overlooked in short-text-only frameworks.

\item \textbf{Limited Utility of Basic Language Models in Clinical or Campus Contexts:}
Although large language models showed promise, few were specialized for mental health screening, which requires deeper understanding of psychological indicators. Generic or rule-based chatbots offered basic question-and-answer flows but struggled to align with clinical guidelines (e.g., DSM-5) or well-validated tools (like PHQ-8). This gap hindered accurate early detection of emotional distress in more sensitive contexts.

\item \textbf{Handling Complex, Multi-Turn Data:}
Many text analytics systems performed well on simple or single-turn data. However, campus well-being often unfolds in extended narratives—whether in open-ended surveys or multi-turn dialogues about personal challenges. Without a robust way to manage longer contexts or multiple topic shifts, these systems risked missing crucial signals and producing incomplete assessments.
\end{enumerate}

To address these issues, this dissertation introduces several AI-driven methods, each tackling a particular gap:

\begin{itemize}
    \item \textbf{TigerGPT \cite{tang2025tigergpt}.} 
    Responding to Gap \#1, TigerGPT improves upon static surveys by offering dynamic prompts and personalized follow-ups, thereby enhancing user engagement and capturing richer feedback on campus life.

   \item \textbf{AURA (Reinforcement Learning Framework for Survey Chatbot).}
    Addressing Gap \#2, AURA learns \emph{within the session} to adjust follow-up question types as engagement evolves, directly targeting low response quality (brevity, shallow specificity, limited elaboration, weak emotional disclosure) in short 10--15 exchange conversations. In controlled studies, AURA sustained depth and specificity across heterogeneous respondents, improved overall response quality, and strengthened late-phase performance relative to baseline systems.

    \item \textbf{Decoding Linguistic Nuances \cite{tang2024decoding}.} 
    Targeting Gap \#3, this approach refines text classification for mental health by analyzing short yet expressive user narratives in detail, ensuring subtle indicators of distress are not missed.

    \item \textbf{Psycho Analyst \cite{tang2024advancing}.} 
    Addressing Gap \#4, this GPT-4-based model incorporates clinical criteria (DSM-5, PHQ-8) to detect psychological distress more accurately, bridging the gap between generic LLMs and clinically oriented mental health screening.

    \item \textbf{SMMR \cite{tang2025layered}.} 
    Confronting Gap \#5, the Stacked Multi-Model Reasoning framework handles extended transcripts or multi-turn conversations by unifying multiple expert models, offering robust analysis for complex, long-context data typical of real campus interactions.
\end{itemize}

By tackling these five gaps, this dissertation advances both satisfaction monitoring and early detection of psychological distress. Chapter~\ref{chap:preventive} presents the development of survey chatbot for campus feedback (TigerGPT and AURA), while Chapter~\ref{chap:intervention} explores the parallel mental health solutions (Decoding Linguistic Nuances, Psycho Analyst, and SMMR). Although the latter are tested on broader, public datasets, they demonstrate key capabilities with clear potential to be adapted for real campus contexts. Taken together, these lines of research illustrate how AI can enhance both preventive and intervention-based strategies for campus well-being.

\chapter{New Preventive AI Methodologies – Survey Chatbot for Campus Satisfaction}
\label{chap:preventive}
\blfootnote{Portions of this chapter were previously published in \cite{tang2025tigergpt}. © 2025 IEEE. Reprinted, with permission, from J. Tang, S. Chen, and Y. Shang, ``TigerGPT: A Theory-Driven AI Chatbot for Adaptive Campus Climate Surveys,'' \textit{2025 IEEE International Conference on Future Machine Learning and Data Science (FMLDS)}, Los Angeles, CA, USA, 2025, pp. 668--673, doi: 10.1109/FMLDS67896.2025.00110.}

\section{Motivation}
Collecting detailed feedback on campus satisfaction is essential for identifying and addressing issues that affect students, faculty, and staff. Traditional surveys, however, often produce limited and superficial responses. Participants may skip items they find irrelevant or provide minimal detail if they are not prompted to elaborate, which risks obscuring concerns about resource allocation, inclusion, or academic support.

Conversational survey interfaces can improve participation and enrich the depth of responses \cite{xiao2020tell, tanwar2024opinebot}. A chatbot can interpret brief or vague replies, ask clarifying follow-ups, and tailor questions to a user’s role and experiences, turning a one-sided questionnaire into an interactive dialogue. These advantages motivated the development of two completed tools in this dissertation: \textit{TigerGPT}, which demonstrated an effective LLM-mediated conversational survey for campus feedback, and \textit{AURA}, which adds within-session adaptation to sustain depth and specificity across short conversations.

\paragraph{Terminology.}
A \emph{conversation session} denotes the full interaction with a single user. An \emph{exchange} is a paired contribution consisting of one chatbot message and one user message, in either order.

\paragraph{Chapter roadmap.}
Section~\ref{sec:TigerGPTV1} and Section~\ref{sec:TigerGPTv2} present \textit{TigerGPT} \cite{tang2025tigergpt}, highlighting design rationale (empathetic prompts, user-driven topic selection). Section~\ref{sec:AURA} introduces \textit{AURA}, an RL-based framework that learns within the session to select follow-up question types from real-time quality signals.

\begin{itemize}
    \item \textbf{TigerGPT} (\autoref{sec:TigerGPTV1}) --- a new AI-driven adaptive conversational survey chatbot that increased engagement and perceived personalization relative to static web forms.
    \item \textbf{AURA} (\autoref{sec:AURA}) --- a new reinforcement learning framework that performs within-session learning during the survey to maintain depth, specificity, and improve response quality across 10--15 exchanges.
\end{itemize}


\section{New Survey Chatbot - TigerGPT \cite{tang2025tigergpt}}
\label{sec:TigerGPTV1}
As part of our effort to enhance campus well-being through proactive feedback methods, we developed a conversational AI chatbot—often referred to as \textbf{TigerGPT}—to collect detailed, adaptive insights on student and faculty satisfaction. Traditional surveys typically offer limited scope and minimal engagement, whereas a chatbot can dynamically prompt users for clarifications and deeper reflections. Below, we first present the underlying motivation for this work, followed by an overview of the chatbot’s design and preliminary findings.

\subsection{Design of TigerGPT}

\begin{figure*}[t]
    \centering
    \includegraphics[width=0.95\linewidth]{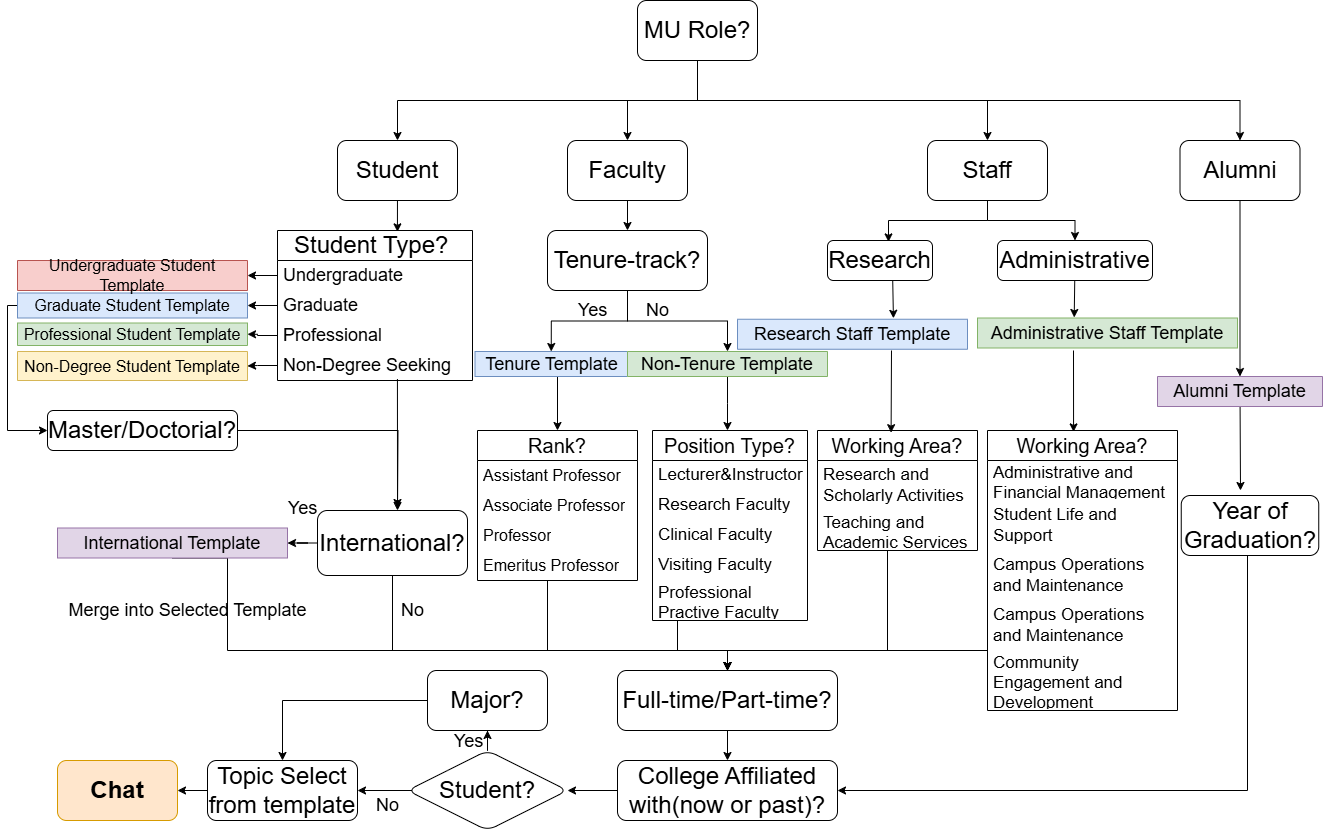} 
    \caption{{Overview of TigerGPT's initial role selection process.} This flowchart illustrates TigerGPT’s selection logic for campus climate surveys at the University of Missouri. Each user’s role (student, faculty, staff, or alumni) plus details such as a student’s type, a faculty member’s rank, or a staff member’s working area guides them to the most suitable survey template and topics.}
    \label{fig:flowchart}
\end{figure*}

\begin{figure}[t]
    \centering
        \includegraphics[width=0.8\linewidth]{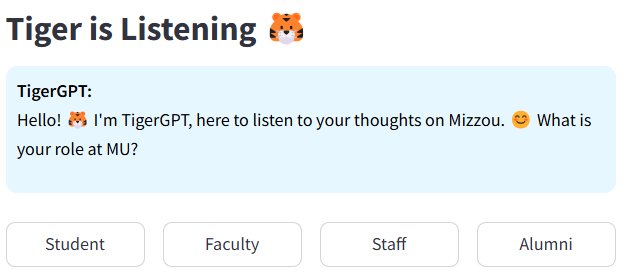}
    \caption{A screenshot of TigerGPT’s initial interface. A welcome screen prompting users to select their role at MU, as student, faculty, staff or alumni, to tailor subsequent interactions accordingly.}
    \label{fig:Preview}
\end{figure}

\begin{figure}
    \centering
    \includegraphics[width=0.8\linewidth]{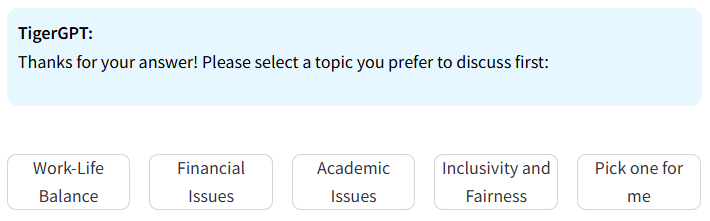}
    \caption{ TigerGPT offers users a variety of survey topics and an option of random selection.}
    \label{fig:subfig_multiple}
\end{figure}

TigerGPT is powered by advanced LLMs, such as OpenAI’s ChatGPT-4-Turbo, and enables dynamic and context-aware conversations. As illustrated in \autoref{fig:flowchart}, the chatbot begins with a brief role selection process, selecting the roles of student, faculty, staff or alumni, along with details such as degree level, international status, and faculty track or working area. This concise setup helps users reduce typing and quickly load the most relevant survey template. Rather than confining the chatbot to a rigid script, however, the chatbot dynamically tailors subsequent questions and responses based on context and conversation history, capable of handling ambiguous or inconsistent inputs. \autoref{fig:Preview} shows a screenshot of TigerGPT’s greetings and initial interface. \autoref{fig:subfig_multiple} shows a screenshot of a variety of survey topics and an option of random selection.

TigerGPT uses Streamlit to implement its web interface and LangChain to orchestrate calls to the OpenAI API, calling LLMs, such as ChatGPT-4-Turbo, to generate responses in real time. Survey participants receive conversation flows that closely match their specific contexts while enjoying a smooth and engaging survey experience.

\begin{figure}[!t]
    \centering
        \includegraphics[width=0.6\linewidth]{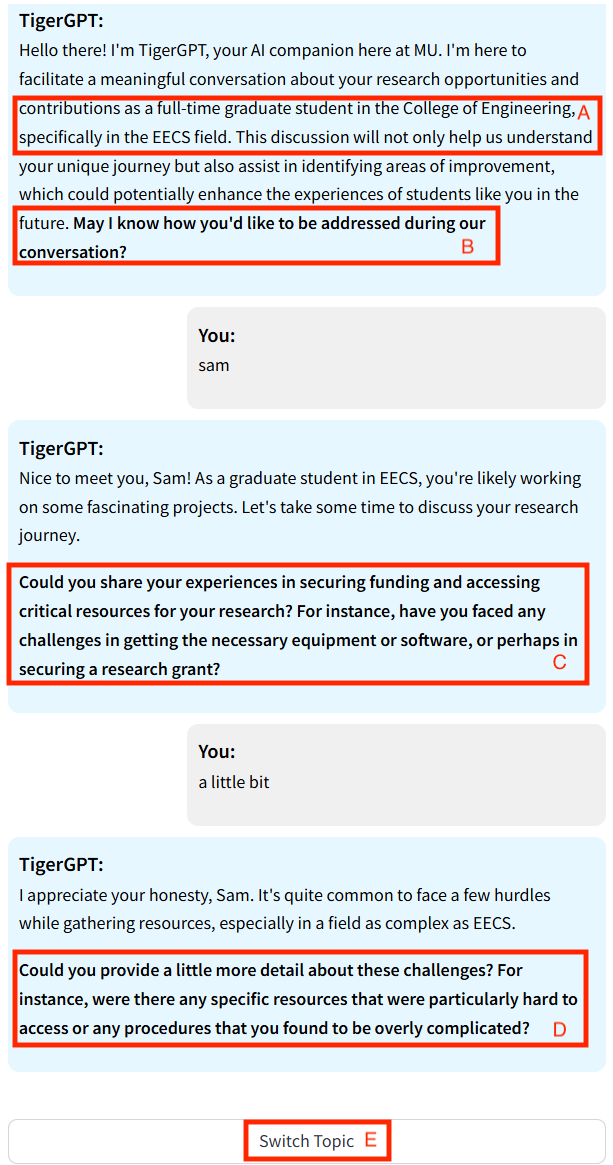}
    \caption{{Sample Conversation with TigerGPT.} This screenshot demonstrates five features: (A) the chatbot’s awareness of user information, (B) a prompt for the user’s preferred name, (C) a bold question supported by examples, (D) a follow-up request for more details, and (E) a “Switch Topic” button allowing the user to change the conversation’s direction.}
    \label{fig:mainFeatures-first}
\end{figure}

TigerGPT is designed based on established principles of conversational design, employing personalization, empathetic
cues, bolded questions, and user-driven topic selection to
encourage sustained interaction and create a
welcoming environment and maintain user attention. Its key features are presented below.

\subsubsection{{Personalized Conversation}} Personalization in chatbots can significantly boost user engagement and satisfaction \cite{kocaballi2019personalization,chen2024recent}. In TigerGPT, each dialogue session is tailored to the user’s role, preferences, and contextual details (e.g., major, research focus), resulting in more targeted prompts and responses. An example of this personalization feature is shown in \autoref{fig:mainFeatures-first}: Label A.

Once the user makes their initial selections (e.g., role, degree program), TigerGPT stores these attributes in a short “user profile” prompt. This profile is then inserted into each conversation request so the chatbot remains aware of the user’s background. By merging the user profile with the active survey template, TigerGPT consistently delivers content that feels both relevant and personalized.

\subsubsection{{Asking for a Participant’s Preferred Name}} 
Using a participant’s chosen name is a personalization technique that fosters rapport and familiarity, making interactions feel more personal and inviting \cite{reghunath2021expression}. By greeting survey participants with their preferred name or referencing it throughout the conversation (see \autoref{fig:mainFeatures-first}: Label B), TigerGPT creates a more human-like presence that can deepen engagement and overall satisfaction.

TigerGPT is prompted to introduce itself and ask for the participant’s preferred name at the  first interaction without asking any other questions. If a name is provided, the system stores it in session memory and integrates it into subsequent prompts (e.g., “Name, could you share more about…?”). If no name is given, TigerGPT defaults to a friendly, generic greeting (e.g., “Hey there!”), ensuring that all interactions remain welcoming and personalized, regardless of the user’s choice.

\subsubsection{Clear, Bold Prompts with Examples} 
Guided by principles of cognitive load reduction \cite{lenzner2010cognitive}, TigerGPT uses \emph{bolded questions}—i.e., questions rendered in visibly heavier or darker text—to help them stand out from other information in the conversation. This visual emphasis makes each prompt quicker to spot and easier to process. It is prompted to ask one open-ended question at a time, using double asterisks to make each query appear bold on-screen and underscore its importance. Each topic also includes a short guidance example illustrating the depth and style of detail users might provide. This approach clarifies expectations, encourages more thoughtful input, and ensures users fully grasp each query (see \autoref{fig:mainFeatures-first}: Label C).

\subsubsection{{Follow-Up Prompts for Gathering Nuanced Information}} Studies on conversational interfaces show that well-placed follow-up questions encourage users to add nuance to their replies, often yielding richer, more detailed feedback \cite{pitts2007improving}. For instance, after a brief or unclear response, a chatbot might ask, “Could you tell me more about that?”, a technique that can lead to more in-depth answers than a single-question approach. TigerGPT adopts this strategy when user inputs appear too short or ambiguous, promoting deeper discussion and a fuller understanding of user needs (see \autoref{fig:mainFeatures-first}: Label~D).

TigerGPT is specifically prompted to offer additional context or examples if a response seems insufficiently detailed. However, to avoid overwhelming users,  TigerGPT only requests further elaboration once per main topic, rather than after every reply. This balanced approach encourages users to expand on their answers without disrupting the conversational flow.

\subsubsection{{User Control via a “Switch Topic” Button}} TigerGPT provides a dedicated button that allows users to switch to a different survey topic they have not yet explored, aligning with usability principles that emphasize “user control and freedom.” By providing a “Switch Topic” button, TigerGPT lets participants gracefully redirect the dialogue and maintain a more user-centric interaction (see \autoref{fig:mainFeatures-first}: Label~E). When this button is clicked, the system checks which topics have not been discussed and presents them to the user, ensuring participants can freely explore multiple survey areas without feeling constrained by a single line of questioning. This mechanism  improves the chatbot experience.

\begin{figure}[!t]
    \centering
        \includegraphics[width=0.6\linewidth]{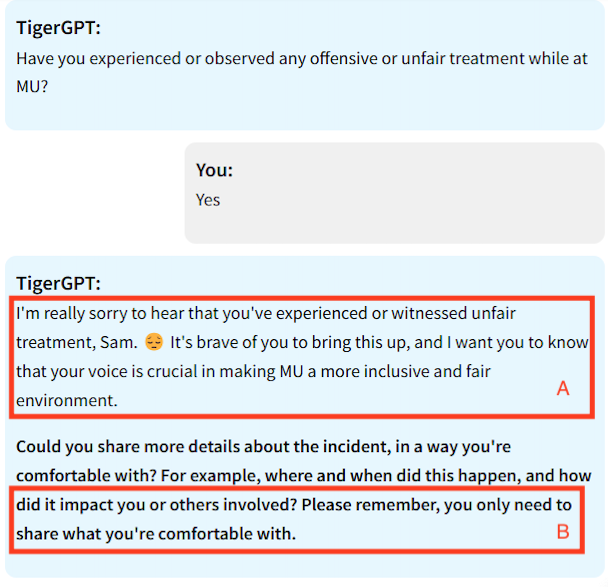}
    \caption{{Sample Conversation for a Sensitive Topic.} This example shows TigerGPT’s empathetic approach to user experiences of unfair treatment. In (A), the chatbot uses active listening and supportive language (with emojis) to validate the user’s feelings. In (B), it invites the user to share more details only at their comfort level, illustrating a respectful framework for discussing sensitive issues.}
    \label{fig:mainFeatures-second}
\end{figure}

\subsubsection{{Active Listening, Empathetic Responses, and Emoji Implementation}} Active listening and empathetic language foster trust and engagement by showing genuine concern for users’ experiences. Research suggests that empathic chatbot responses significantly enhance user satisfaction \cite{zhou2025building}, while well-chosen emojis convey additional warmth and emotional cues, contributing to a more human-like conversation \cite{bai2019systematic}. Together, these elements encourage users to feel heard and comfortable discussing personal topics (see \autoref{fig:mainFeatures-second}: Label A).

To achieve this, TigerGPT maintains a supportive, non-judgmental tone throughout each interaction, acknowledging users’ courage in sharing sensitive information. When users express distress or other strong emotions, the system responds with empathetic phrases. At strategic points, emojis are used to visually reinforce these emotional nuances, further enhancing user engagement and rapport.

\subsubsection{{Encouraging Comfortable Information Sharing}}
Openly discussing sensitive or personal matters can be challenging for participants, especially if they perceive judgment or pressure \cite{melville2016conducting}. The concept of self-disclosure in social psychology emphasizes that individuals are more likely to share personal information when they trust their conversational partner and retain control over how much to reveal \cite{wheeless1977measurement}. Additionally, respect for autonomy in user-centered design \cite{kim2008keeping} underscores the importance of allowing participants to set their own boundaries, thereby creating a safer environment for disclosure (see \autoref{fig:mainFeatures-second}: Label B).

TigerGPT  maintains a supportive, non-judgmental tone throughout each interaction—especially when topics become sensitive. If the user shows hesitation or distress, the chatbot adjusts its questions accordingly, refraining from further probing unless the user explicitly indicates a willingness to continue. Where relevant, TigerGPT also provides information on support resources, avoiding any demands for personal data. This balanced approach encourages deeper sharing while preserving participants’ sense of security.

\subsubsection{{Providing Multiple Topic Options for Personalized Guidance}} To enhance user autonomy and relevance, TigerGPT presents a set of topics aligned with each individual’s profile, along with an option to pick one randomly. This design choice allows participants to guide the conversation toward areas they find most pertinent or interesting, while also offering a more free-form alternative for those who prefer spontaneity (see \autoref{fig:subfig_multiple}). By empowering users to choose the next topic, TigerGPT fosters greater engagement and ensures each session remains tailored to personal preferences.

The detailed evaluation result for this new TigerGPT survey chatbot will be presented in Chapter~\ref{chap:evaluation}.

\section{A Newly Released TigerGPT V2}
\label{sec:TigerGPTv2}
\begin{figure}[!ht]
        \centering
        \includegraphics[width=\textwidth]{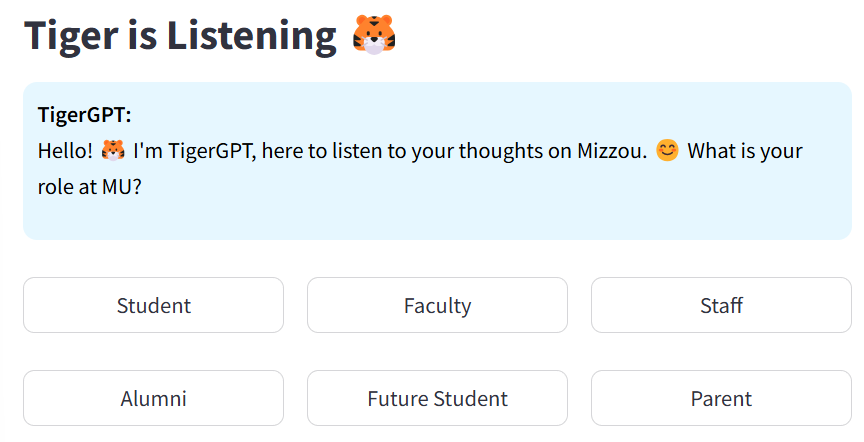}
        \caption{TigerGPT V2 Preview}
    \label{fig:Tiger_V2_preview}
\end{figure}

\begin{figure}[!ht]
        \centering
        \includegraphics[width=\textwidth]{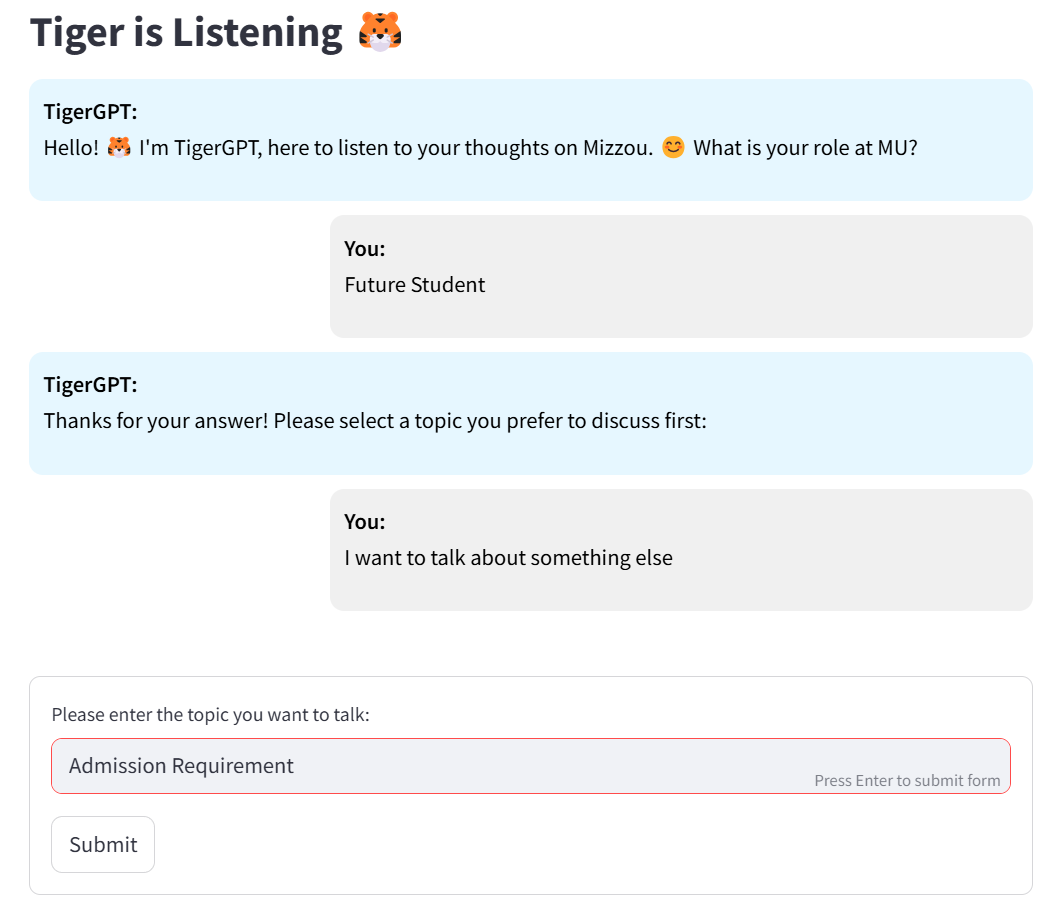}
        \caption{TigerGPT V2 Customized Topic Feature}
    \label{fig:Tiger_V2_Customized}
\end{figure}
Building on the original TigerGPT (Section~\ref{sec:TigerGPTV1}), this upgraded version broadens the chatbot’s audience to include prospective students and parents \autoref{fig:Tiger_V2_preview}. In addition, TigerGPT V2 now lets users move beyond predefined topics by typing any free-form concerns they wish to discuss (\autoref{fig:Tiger_V2_Customized}). This added flexibility ensures that individuals can raise issues or ask questions even if they fall outside the chatbot’s default conversation flows—for example, admissions processes for high school students, financial aid concerns specific to parents, or less common campus life inquiries.

In addition, TigerGPT V2 now compiles user logs into a dynamic summarization report. Rather than requiring manual review of free-form entries, the system automatically synthesizes feedback, organizes major themes, and detects shifting user concerns in real time. Administrators can then quickly identify urgent issues and adapt campus initiatives accordingly (see Chapter~\ref{chap:evaluation} for example of the summarization report). By integrating a more open conversation flow with automated reporting, TigerGPT V2 captures a diverse spectrum of community needs with minimal overhead.

\section{New Reinforcement Learning Framework for Survey Chatbot - AURA}
\label{sec:AURA}
Building on the pilot findings in Section~\ref{sec:TigerGPTV1}, we address Gap~\#2: current chat-based surveys lack within-session adaptation and instead rely on fixed dialogue trees or static prompt templates, yielding generic follow-ups, superficial personalization, and lower response quality (especially in short 10--15 exchange sessions). We introduce \textbf{AURA}, a reinforcement learning framework for adaptive questioning in campus surveys.

\subsection{Overview of the AURA Framework}
\label{sec:overview}

AURA (Adaptive Understanding through Reinforcement learning for Assessment) integrates conversational AI with reinforcement learning to enable real-time personalization in campus-climate surveys. Unlike traditional chatbots that follow fixed question sequences, AURA continuously monitors user engagement and adapts its questioning strategy within each conversation based on observed response-quality patterns. This adaptive loop transforms passive data collection into an interactive optimization process, allowing AURA to infer which questioning strategies maximize engagement within a single conversation and ultimately elicit richer, more detailed feedback than static survey instruments. 

\begin{figure}[h]
    \centering
    \includegraphics[width=\textwidth]{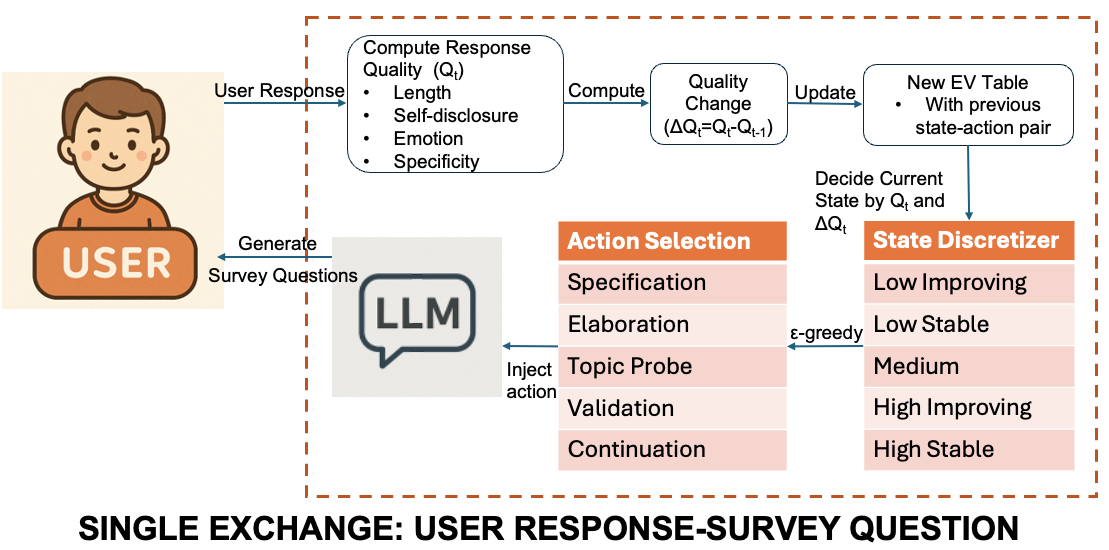}
    \caption{AURA system architecture showing the reinforcement learning cycle 
    within a single conversation exchange. Each user response is scored along 
    four quality dimensions (LSDE), mapped to an engagement state, and used 
    to select the next question type via an $\epsilon$-greedy policy. Observed 
    quality changes update the system's expected-value (EV) estimates, 
    enabling rapid within-session adaptation.}
    \label{fig:AURA_system_architecture}
\end{figure}

Figure~\ref{fig:AURA_system_architecture} illustrates the complete architecture. The system operates as a continuous feedback loop: user responses are assessed for quality, generating reward signals that update action-selection policies in real time. This within-session learning enables the system to discover which questioning strategies work best for each individual participant. 

\begin{figure}[h]
    \centering
    \includegraphics[width=\textwidth]{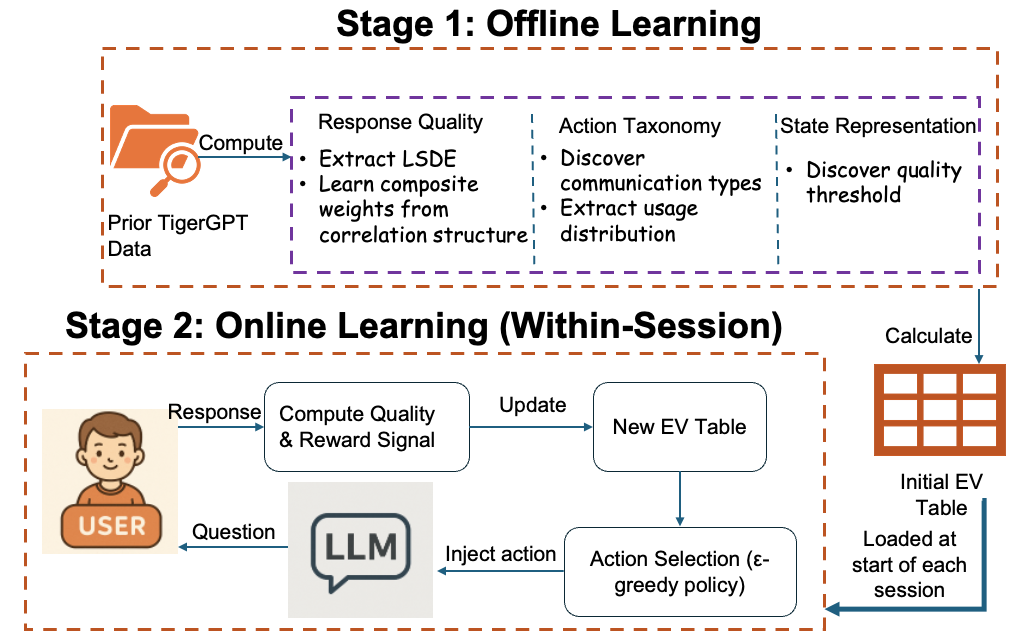}
    \caption{Two-level learning framework. Offline learning (top) extracts patterns from prior conversations to initialize the policy. Online learning (bottom) adapts to individual users within each session through real-time quality feedback, then resets to priors for the next user.}
    \label{fig:AURA_two_level_learning}
\end{figure}

AURA employs a two-level learning strategy that distinguishes it from conventional multi-session reinforcement learning systems (Figure~\ref{fig:AURA_two_level_learning}). At the \textit{offline level}, the system extracts patterns from 96 prior campus-climate conversations to initialize an expected-value (EV) table estimating the quality gain associated with each question type in each engagement state. This offline process establishes population-level priors derived from authentic survey interactions. At the \textit{online level}, during each individual conversation, AURA updates its EV estimates in real time based on the current user's responses, discovering personalized strategies within 10--15 exchanges. Critically, the system resets to priors at the start of each new conversation—no information from previous users is retained—ensuring that adaptive learning remains session-specific and that participant privacy is fully preserved.

This architecture combines the generalizability of population-level patterns with the responsiveness of individual adaptation. By beginning each conversation with empirically grounded priors but adjusting strategy dynamically as user engagement evolves, AURA discovers personalized questioning approaches that static systems cannot achieve. The remainder of this section details the data sources (Section~\ref{sec:data}), quality-assessment methodology (Section~\ref{sec:lsde}), state and action representations (Sections~\ref{sec:actions}--\ref{sec:states}), and reinforcement learning framework (Section~\ref{sec:rl_framework}).

\subsection{Data Collection and Preprocessing}
\label{sec:data}

To initialize AURA's reinforcement-learning policy and establish empirical foundations for engagement measurement, we used campus-climate survey conversations collected from a pilot deployment of TigerGPT \cite{tang2025tigergpt}.  TigerGPT is a semi-adaptive survey chatbot that conducts natural, open-ended dialogues but does not employ reinforcement learning.  The resulting dataset provides authentic examples of engagement in conversational surveys and enables offline estimation of action effectiveness across different engagement states.

\subsubsection{Prior Conversation Dataset}

The dataset comprises 96 campus-climate survey conversations conducted between October 2024 and February 2025 at the University of Missouri.  The chatbot was publicly accessible through a web-based survey link, and all responses were collected anonymously as part of routine feedback activities.  Each conversation included an initial role selection (student, faculty, or staff) followed by optional demographic prompts to personalize dialogue context.  Participants then selected one of five available survey topics—academic issues, financial concerns, work–life balance, campus inclusivity, or a random option—and engaged in multi-exchange dialogues where TigerGPT posed open-ended questions and users provided free-text responses.

\subsubsection{Data Cleaning and Quality Control}

From raw conversation logs totaling 683 exchanges, we applied standard preprocessing procedures:  
(1) removed entries where either the chatbot dialogue or user response was missing, empty, or contained placeholder strings (``nan'', ``N/A''); 
(2) excluded duplicate responses caused by technical errors; and (3) retained only substantive user inputs containing at least one word. 
This yielded a final dataset of 467 valid user responses across 96 conversations. The 31.6\% exclusion rate reflects typical patterns in conversational data collection, where technical issues, participant dropout, or minimal engagement produce incomplete records.

\subsubsection{Dataset Characteristics}

\begin{table}[!ht]
\centering
\caption{Summary Statistics of the Prior Conversation Dataset.}
\label{tab:AURA_dataset_summary}
\begin{tabular}{
    @{\hspace{3mm}}l
    @{\hspace{6mm}}r
}
\toprule
\textbf{Category} & \textbf{Value} \\
\midrule
\multicolumn{2}{l}{\textit{Conversation-level}} \\[1pt]
\quad Total conversations & 96 \\
\quad Mean exchanges per conversation & 4.9\,$\pm$\,5.5 \\
\quad Median exchanges per conversation & 2.5 \\
\quad Range of exchanges & 1–18 \\
\quad Single-exchange conversations & 28\,(29.2\%) \\
\quad Multi-exchange conversations (2+ exchanges) & 68\,(70.8\%) \\
\addlinespace[4pt]
\multicolumn{2}{l}{\textit{Response-level}} \\[1pt]
\quad Total valid responses & 467 \\
\quad Consecutive exchange pairs (for sequential analysis) & 371 \\
\quad Mean response length (words) & 18.3\,$\pm$\,22.2 \\
\quad Median response length (words) & 10.0 \\
\bottomrule
\end{tabular}
\end{table}

Table~\ref{tab:AURA_dataset_summary} summarizes the characteristics of the prior conversation dataset collected from TigerGPT, which serves as the empirical foundation for AURA's reinforcement learning initialization.

The dataset exhibits substantial heterogeneity in conversation length. Single-exchange interactions (29.2\%) occurred when users provided initial feedback but chose not to continue; these responses contribute to cross-sectional quality measurement but cannot inform dynamics of engagement change. Multi-exchange conversations (70.8\%, $N=68$) enable analysis of how response quality evolves across exchanges, producing 371 consecutive exchange pairs used for sequential analysis.

In this context, an \textit{exchange} represents one full exchange between the chatbot and a user. An \textit{exchange pair} links two consecutive user responses via the intervening chatbot question, enabling analysis of how question types affect response quality changes. Since the first response in each conversation has no preceding response for comparison, the 467 total responses yield 371 exchange pairs for sequential analysis ($467 - 96 = 371$).

The longest conversation reached 18 exchanges, though most remained brief (median\,=\,2.5), reflecting typical engagement patterns in voluntary online surveys where respondents balance disclosure with time constraints. This distribution shaped AURA's design requirements: conversations are short enough (mean\,=\,4.9 exchanges) that adaptive systems must learn rapidly, yet long enough (70.8\% with multiple exchanges) to allow repeated state visits and within-session policy refinement.

The remainder of this section details how AURA transforms these prior conversations into an adaptive learning system: Section~\ref{sec:lsde} presents the LSDE quality-scoring methodology applied to all responses, Section~\ref{sec:actions} defines the action space of question types, Section~\ref{sec:states} describes the state representation that discretizes engagement patterns, and Section~\ref{sec:rl_framework} explains the reinforcement learning algorithm that enables within-session personalization.

\subsubsection{Data Usage in AURA Development}

The prior conversation data served three primary functions in AURA's development.

\textbf{Empirical grounding for quality metrics.} 
The 467 user responses provided a reference distribution for establishing normalization parameters and threshold values in the LSDE scoring system (Section~\ref{sec:lsde}). 
For example, the length normalization cap was set at 29~words (the 75th percentile of the observed distribution), ensuring that normalized scores reflect empirical variability rather than arbitrary theoretical bounds.

\textbf{Initial policy estimation.} 
The 371 consecutive exchange pairs enabled the calculation of expected-value (EV) estimates for state–action combinations, capturing population-level patterns in how different question types affect engagement across distinct user states (Section~\ref{sec:rl_framework}). 
These estimates initialize AURA's policy before individual conversations begin, providing evidence-based starting points derived from authentic survey interactions rather than random or uniform priors.

\textbf{Action taxonomy validation.} 
The diversity of question types in the corpus (Section~\ref{sec:actions}) confirmed that the five-action categorization—\textit{specification}, \textit{elaboration}, \textit{topic probe}, \textit{validation}, and \textit{continuation}—captured meaningful functional distinctions in conversational question design. 
Question classification used an LLM-based content analysis (GPT-4o) followed by human validation ($\kappa$\,=\,0.83), establishing both the taxonomic structure and the empirical distribution of action usage in naturalistic campus-climate dialogues.

Importantly, these prior data inform AURA's initialization but do not constrain its behavior during live conversations. Each new user begins with population-level priors, but the system adapts its policy in real time based on individual response patterns, discovering personalized strategies that may diverge substantially from prior averages. The prior corpus thus serves as an empirically grounded prior rather than a fixed reference determining all subsequent decisions.

\subsection{Response Quality Assessment (LSDE)}
\label{sec:lsde}
Effective conversational surveys require real-time measurement of response quality to guide adaptive question selection. We developed a four-dimensional quality metric (Length, Specificity, Self-disclosure, and Emotion; LSDE) that captures distinct aspects of engagement and enables automated assessment during live conversations. This section presents the theoretical foundations for each dimension, describes implementation procedures, and validates the composite scoring approach through empirical analysis of prior campus-climate conversations.

\subsubsection{Foundations and Operationalization}

Survey methodologists emphasize that response quality in open-ended questions is inherently multidimensional: no single indicator fully captures informativeness, engagement, or authenticity \cite{dillman2014internet, tourangeau2000psychology}. Traditional assessment relies on post-hoc human coding, which cannot support real-time adaptive systems. AURA operationalizes four quality dimensions through automated linguistic analysis, enabling exchange-by-exchange measurement during conversation.

\paragraph{Length and Elaboration}

Response length provides a first-order indicator of respondent investment and information content \cite{dillman2014internet, holland2009measuring}. In traditional web surveys, detailed answers ($\approx$50 words) are rated as substantive, but chatbot conversation exchanges are naturally shorter because users elaborate across multiple exchanges.  In our dataset ($N=467$), the median response length was 10 words and the 75th percentile (Q75) was 29 words, providing an empirically grounded scale for this dialogue context.  Length is therefore normalized as
\begin{equation}
L_{\text{norm}} = \min\left(\frac{\text{word\_count}}{29}, 1.0\right)
\end{equation}
so that the top quartile of responses receives full credit for elaboration while maintaining discrimination among briefer exchanges.

\paragraph{Specificity and Concreteness}

Concrete, detailed responses convey richer and more actionable information than vague generalities.  Episodic memory research shows that specific recollection involves three components—entities (who/what), temporal references (when), and spatial references (where) \cite{levine2002autobiographical}.  Authentic accounts contain more such contextual details than fabricated ones ($d=0.72$; \cite{newman2003lying}).  Unlike word-level concreteness metrics \cite{brysbaert2014concreteness}, AURA detects the presence of these episodic components as binary indicators of specificity.

Each response is classified by GPT-4o for presence (1) or absence (0) of \textbf{entities}, \textbf{temporal}, and \textbf{spatial} details, following \cite{levine2002autobiographical}'s framework.  Manual inspection of 50 samples confirmed consistent classification accuracy, after which the protocol was applied to the full corpus:
\begin{align}
S_{\text{total}} &= \text{entities} + \text{temporal} + \text{spatial} \in \{0, 1, 2, 3\}, \\
S_{\text{norm}}  &= \frac{S_{\text{total}}}{3}.
\end{align}
Specificity is extremely sparse in our data (88.7\% of responses have $S_{\text{total}} = 0$, Q75 = 0). Although each component is detected binarily, their sum captures meaningful gradations: responses may contain one component (10.1\%), two components (1.1\%), or all three (0.2\%). This count-based approach preserves distinctions among the 11.3\% of responses containing specific detail while remaining robust to the severe sparsity.

\paragraph{Self-Disclosure and Personal Engagement}

First-person pronouns indicate self-focus and willingness to share personal experiences \cite{pennebaker2003linguistic, tausczik2010psychological}.  In survey contexts, self-disclosure enriches qualitative feedback because responses grounded in personal experience (e.g., ``I felt excluded when\dots'') reveal deeper insight than abstract statements (e.g., ``Some students experience exclusion'').  Because our responses are brief (median = 10 words), we use absolute rather than ratio-based pronoun counts to avoid mechanical dependence on response length: as text length increases, ratio scores decline even when disclosure remains constant, producing the strong Length–Disclosure correlation ($r = 0.856$; Table~\ref{tab:AURA_correlation}).  Absolute counts, capped at the 75th-percentile value of three pronouns, measure disclosure directly while retaining discriminative range. Self-disclosure is therefore normalized as

\begin{equation}
D_{\text{norm}} = \min\left(\frac{\text{pronoun\_count}}{3}, 1.0\right)
\end{equation}
where 3 represents the Q75 of our pronoun distribution. Pronouns include \emph{I, me, my, mine, myself, we, us, our, ours, ourselves}, following LIWC's first-person category.

\paragraph{Emotional Expression and Intensity}

Emotional language reflects psychological investment and topic salience.  According to Russell's \cite{russell1980circumplex} circumplex model of affect, emotion varies along valence (positive–negative) and arousal (intensity).  In surveys, highly affective language—positive or negative—indicates engagement \cite{porter2005mail}.  To measure this automatically, we apply VADER sentiment analysis \cite{hutto2014vader} and use the magnitude of its compound score as a proxy for emotional intensity, independent of polarity.  Although VADER estimates valence rather than arousal directly, research on affective norms shows that extreme valence correlates moderately with arousal ($r \approx 0.40$; \cite{warriner2013norms}), making the absolute compound value a practical indicator of engagement strength in brief conversational responses. Emotional intensity is therefore normalized as:
\begin{equation}
E_{\text{norm}} = \left| \text{VADER}_\text{compound} \right|
\end{equation}
This transformation yields normalized emotion scores on $[0, 1]$, where 0 indicates affectively neutral language and 1 indicates maximal emotional intensity.

\subsubsection{Composite Scoring and Weighting Rationale}
\label{sec:lsde_weighting}

With each dimension normalized to $[0,1]$, we combine them into a composite quality score.  Equal weighting ($0.25$ per dimension) would overrepresent correlated features: Table~\ref{tab:AURA_correlation} shows strong interdependencies among Length, Disclosure, and Emotion ($r=0.52$–$0.86$), whereas Specificity is relatively independent ($\text{mean}\,|r|=0.27$).

\begin{table}[htbp]
\centering
\caption{Correlation Matrix of Normalized LSDE Dimensions ($N = 467$).}
\label{tab:AURA_correlation}
\begin{tabular}{lcccc}
\toprule
            & Length & Specificity & Disclosure & Emotion \\
\midrule
Length      & 1.000  & 0.290       & 0.856      & 0.664   \\
Specificity & 0.290  & 1.000       & 0.248      & 0.259   \\
Disclosure  & 0.856  & 0.248       & 1.000      & 0.523   \\
Emotion     & 0.664  & 0.259       & 0.523      & 1.000   \\
\bottomrule
\end{tabular}
\end{table}

Unequal weights were therefore assigned to balance three goals: (1) limit redundancy among highly correlated dimensions, (2) emphasize dimensions with broad empirical variation, and (3) retain all four theoretical constructs.  The final composite metric is

\begin{equation}
Q_{\text{composite}} = 0.20 L_{\text{norm}} + 0.20 D_{\text{norm}} + 0.35 E_{\text{norm}} + 0.25 S_{\text{norm}}
\end{equation}

\noindent\textbf{Weighting rationale.}
\begin{itemize}
\item \textbf{Length and Disclosure (0.20 each; 0.40 total):} Highly correlated ($r=0.86$); combined weight limits redundancy while reflecting their joint role in elaboration and engagement.
\item \textbf{Emotion (0.35):} Moderately correlated ($r=0.52$–$0.66$) but theoretically central to engagement intensity and empirically well distributed.
\item \textbf{Specificity (0.25):} Statistically most independent but sparse (88.7 \% zeros).  A moderate weight preserves sensitivity for common responses while rewarding information-rich details.
\end{itemize}

\subsubsection{Empirical Validation}

To evaluate the composite metric, we computed quality scores for all 467 prior responses.  Scores spanned the full $[0,1]$ range (mean = 0.328, SD = 0.276), indicating broad variability and confirming that the metric captures meaningful differences in engagement levels (Table~\ref{tab:AURA_quality_dist}).  

\begin{table}[htbp]
\centering
\caption{Distribution of Composite Quality Scores ($N = 467$).}
\label{tab:AURA_quality_dist}
\begin{tabular}{lcc}
\toprule
Category  & Score Range & $n$ (\%) \\
\midrule
Very Low  & 0.0--0.2    & 220 (47.1\%) \\
Low       & 0.2--0.4    & 80 (17.1\%)  \\
Medium    & 0.4--0.6    & 90 (19.3\%)  \\
High      & 0.6--0.8    & 70 (15.0\%)  \\
Very High & 0.8--1.0    & 7 (1.5\%)    \\
\bottomrule
\end{tabular}
\end{table}

Nearly half of the responses (47.1\%) fell into the Very Low category, reflecting the engagement challenges typical of campus-climate surveys in which many participants provide only minimal input (e.g., one-word replies).  This distribution demonstrates that the LSDE metric can distinguish low from high engagement with fine granularity, supporting its use as a reward signal for reinforcement learning models that aim to improve response quality through adaptive question selection.

\subsection{Action Space Extraction}
\label{sec:actions}

Effective reinforcement learning for conversational surveys requires a well-defined \textit{action space} representing the questioning strategies available to the system.  Rather than imposing predetermined categories, we extracted action types empirically from prior TigerGPT conversations and mapped them to established communication theory frameworks.  A GPT-4o–based content analysis labeled each survey question into one of five communicative functions—\textit{specification}, \textit{elaboration}, \textit{topic probe}, \textit{validation}, and \textit{continuation}.  The model prompt instructed classification by primary and optional secondary intents, with structured JSON outputs including confidence and brief reasoning.  Human review of a random subset confirmed consistent labeling ($\kappa = 0.83$).  This process yielded a compact yet comprehensive taxonomy suitable for reinforcement-learning–based policy optimization.

\subsubsection{Taxonomy Development: A Theory-Informed, Data-Driven Approach}

Action taxonomy development followed a hybrid methodology balancing empirical observation with theoretical grounding.  Analysis of 467 chatbot questions across 96 TigerGPT conversations (Section~\ref{sec:data}) revealed recurring functional patterns: some prompts requested concrete examples, others sought elaboration, introduced new topics, or offered acknowledgment.  These patterns were consolidated into five communicative categories and validated against established communication frameworks to ensure conceptual coherence \cite{li2016deep, su2021dynamic}.  

This hybrid strategy avoids the limitations of purely data-driven extraction—which can yield uninterpretable clusters—and purely theory-driven design—which may impose artificial categories unrelated to the data.  AURA’s action space is thus both \textbf{empirically grounded} (all categories appear with sufficient frequency in natural dialogues) and \textbf{theoretically justified} (each aligns with established constructs in communication and dialogue management).  

The five-category structure balances interpretability with statistical power.  With 96 conversations producing 371 consecutive exchange pairs, compound multi-intent schemes would create extreme data sparsity.  For instance, allowing two-intent combinations would expand the action space from 5 to 15 categories, and three-intent combinations to 25—yielding only about 15 examples per category on average, with many categories represented by fewer than five instances.  AURA therefore uses primary-intent classifications for reinforcement learning while retaining secondary intents for post-hoc analysis, prioritizing statistical robustness over taxonomic exhaustiveness.

\subsubsection{Action Classification Methodology}

Each chatbot dialogue in the prior corpus was classified by GPT-4o using a structured prompt defining five communicative categories and specifying primary versus secondary intents.  The model identified the main communicative function of each question (primary intent) and any auxiliary functions (secondary intents).  Approximately 65 \% of questions exhibited multiple intents (e.g., “Thank you for sharing. Could you tell me more about that specific example?” combines \textit{validation} and \textit{specification}), illustrating the prevalence of compound strategies in natural dialogue.

Primary intent was defined as the question’s dominant illocutionary force \cite{searle1976classification}—the central action the question seeks to elicit.  Model outputs included structured JSON records with confidence scores and brief reasoning to support quality control and flag ambiguous cases.

To validate classification accuracy, a random sample of 20 questions was human-coded, showing strong agreement between GPT-4o labels and theoretical definitions.  This human-in-the-loop check confirmed that the automated classification captured meaningful functional distinctions rather than superficial lexical cues.

\begin{table}[htbp]
\centering
\begin{threeparttable}
\caption{Action Type Taxonomy, Theoretical Grounding, and Prior Examples}
\label{tab:AURA_action-taxonomy}
\small
\begin{tabularx}{\textwidth}{p{2cm}p{2cm}p{2.5cm}X}
\toprule
\textbf{Action Type} & \textbf{Speech Act Function} & \textbf{CAT Strategy} & \textbf{Example} \\
\midrule
specification & Directive & Convergence (narrowing focus) & ``Could you share how satisfied you are with your academic program and why?'' \\
\addlinespace
elaboration & Directive & Maintenance (sustaining topic) & ``Could you share more about the challenges you've faced in Greek life?'' \\
\addlinespace
topic probe & Commissive & Divergence (expanding scope) & ``How comfortable do you feel on campus, and have you experienced moments of inclusion or exclusion?'' \\
\addlinespace
validation & Expressive & Maintenance (rapport building) & ``Thank you for sharing that. Your perspective is valuable and helps us understand student experiences.'' \\
\addlinespace
continuation & Directive & Maintenance (flow sustaining) & ``Is there anything else you'd like to share about your experiences at MU?'' \\
\bottomrule
\end{tabularx}
\begin{tablenotes}[flushleft]
\item Note: CAT = Communication Accommodation Theory \cite{giles1991contexts}. 
\item Speech act categories follow \cite{searle1976classification}'s taxonomy of illocutionary forces.
\end{tablenotes}
\end{threeparttable}
\end{table}

\subsubsection{Action Type Definitions and Theoretical Grounding}

Table~\ref{tab:AURA_action-taxonomy} presents the five action types with their theoretical foundations and representative examples from the TigerGPT corpus.  Each action corresponds to distinct constructs from Speech Act Theory \cite{searle1976classification} and Communication Accommodation Theory (CAT) \cite{giles1991contexts}, ensuring that AURA’s taxonomy captures functionally meaningful communicative strategies rather than superficial lexical patterns.

\textbf{Specification} and \textbf{elaboration} both operate as directive speech acts but differ in focus.  Specification requests concrete examples or particular cases, using CAT’s convergence strategy to narrow discussion toward actionable details (e.g., transforming “Classes can be challenging” into “Professor Smith’s EECS 280 lectures often run overtime”).  Elaboration maintains the current topic while seeking richer description without requiring specific instantiation (e.g., “Tell me more about your research experience”).  The distinction is functional: specification targets concreteness, elaboration depth.

\textbf{Topic probe} introduces new experiential dimensions, shifting attention to unexplored areas.  These prompts function at the intersection of commissive and directive acts \cite{searle1976classification}, committing the system to expanding the dialogue scope.  They reflect CAT’s divergence strategy \cite{giles1991contexts}, ensuring broad coverage of campus climate dimensions (academic, social, financial, inclusivity).

\textbf{Validation} acknowledges user contributions without requesting new information.  As expressive acts, validation supports rapport and psychological safety through maintenance strategies \cite{giles1991contexts}.  Although rare in prior data ($n=4$, 0.9\%), validation was retained because evidence shows it facilitates deeper disclosure on sensitive topics \cite{wheeless1977measurement, melville2016conducting}.

\textbf{Continuation} invites further input without specifying direction, granting users conversational autonomy.  Also infrequent ($n=2$, 0.4\%), continuation serves as a flow-maintaining strategy that may prove effective for users preferring open-ended engagement \cite{kocaballi2019personalization}.

\begin{table}[htbp]
\centering
\begin{threeparttable}
\caption{Action Type Distribution in Prior Corpus ($N=467$ conversation exchanges)}
\label{tab:AURA_action-distribution}
\begin{tabularx}{\textwidth}{@{}lrrX@{}}
\toprule
\textbf{Action Type} & \textbf{Count} & \textbf{Percentage (\%)} & \textbf{Characteristics} \\
\midrule
Specification & 291 & 62.3 & Requests concrete examples \\
Elaboration & 110 & 23.6 & Asks for expanded detail \\
Topic probe & 60 & 12.8 & Introduces new dimensions \\
Validation & 4 & 0.9 & Acknowledges contributions \\
Continuation & 2 & 0.4 & General follow-up \\
\midrule
\textbf{Total} & \textbf{467} & \textbf{100} & \\
\bottomrule
\end{tabularx}
\begin{tablenotes}[flushleft]
\normalsize
\item \textit{Note:} Distribution reflects prior TigerGPT behavior. AURA's RL framework enables adaptive rebalancing for individual users.
\end{tablenotes}
\end{threeparttable}
\end{table}

\subsubsection{Action Distribution in Prior Data}
\label{sec:histor_action_dist}

Table~\ref{tab:AURA_action-distribution} summarizes the distribution of action types in the prior TigerGPT corpus.  The data show a pronounced imbalance: \textit{specification} dominates (62.3\%), while \textit{validation} (0.9\%) and \textit{continuation} (0.4\%) are rarely used.  This pattern reflects conventional survey design practices that prioritize information extraction over rapport maintenance or open-ended exploration.

From a reinforcement learning perspective, this imbalance presents both a challenge and an opportunity.  The challenge is data sparsity: rare actions have too few examples for reliable expected-value (EV) estimation, creating cold-start conditions for several state–action pairs.  Specifically, eight of twenty-five combinations lack any prior samples (Section~\ref{sec:rl_framework}), requiring AURA to explore them during live interactions.  The opportunity lies in AURA’s capacity to \textbf{correct prior biases}: the dominance of specification likely reflects template-driven defaults rather than optimal engagement strategies.  By adaptively testing underused actions such as \textit{validation} after emotional responses or \textit{topic probe} during stagnation, AURA can discover more effective questioning patterns for individual users.

An $\epsilon$-greedy exploration policy ($\epsilon=0.30$; Section~\ref{sec:rl_framework}) ensures all actions receive exposure despite imbalance.  Roughly 30\% of actions per conversation are randomly selected, allowing even rare strategies to be evaluated.  When these actions improve subsequent response quality, their expected values increase and they are chosen more often for that user; conversely, ineffective actions decline in frequency.  Through this continual rebalancing, AURA learns personalized questioning strategies that static, rule-based systems cannot achieve.

\subsubsection{Multi-Intent Analysis and Primary Action Selection}

About 65 \% of prior questions contained multiple intents, showing that conversational prompts often blend functions.  For instance, “Thank you for sharing that. Could you tell me about a particular instance when this happened?” combines \textit{validation} with \textit{specification}, illustrating how rapport building and information elicitation frequently co-occur.

For reinforcement learning, we label each question by its \textbf{primary intent}—the dominant communicative function.  This choice resolves a key design trade-off: compound actions capture conversational nuance but generate extreme data sparsity.  With five base types, even two-intent combinations expand the action space to 15 categories and three-intent combinations to 25+, leaving fewer than five samples per class across 96 conversations—insufficient for reliable expected-value estimation.

Using primary intent yields a tractable five-action space (25 state–action pairs) while retaining compound information for post-hoc analysis.  Secondary intents were recorded but excluded from training, allowing later evaluation of whether particular combinations (e.g., \textit{specification} + \textit{validation}) consistently enhance engagement.  This design favors \textbf{within-session learning feasibility} over exhaustive taxonomy: AURA must learn effective strategies within 10–15 exchanges per user.  Future extensions could explore hierarchical policies that first select a primary action, then conditionally choose secondary intents as data volume grows.

\subsection{State Representation Design}
\label{sec:states}
Having established AURA's five-action question taxonomy, we now define the state representation that captures user engagement patterns during conversation. Effective reinforcement learning requires discrete states that both reflect meaningful engagement differences and enable sufficient data for policy learning within short conversations. This section describes the state discretization approach, threshold selection rationale, and empirical validation of the five-state framework.

\subsubsection{Design Requirements and Constraints}
State representation in conversational reinforcement learning must balance expressiveness with learnability. AURA faces two critical constraints that shape state design. First, conversations are brief: our prior conversations exhibit a mean of 4.9 exchanges per conversation (median = 2.5), and 70.8\% contain at least two exchanges. Adaptive systems must therefore learn rapidly, discovering effective strategies within the limited exchanges typical of voluntary survey interactions. Second, AURA employs session-specific learning with no cross-user memory retention: the system initializes each conversation with population-level priors but adapts only to the current user, resetting completely before the next interaction. This design preserves privacy and prevents overfitting to specific individuals but requires state spaces that enable meaningful learning from limited within-session experience.

These constraints necessitate discrete rather than continuous state representations. In a typical conversation with continuous states (where each quality score is unique), the system would encounter each state exactly once, precluding any learning from repeated state visits. Discrete states, by contrast, enable state revisits: if a user's engagement remains in the ``medium'' category across multiple exchanges, the system observes which actions succeed or fail in that state, updating its policy accordingly. With 5 discrete states, AURA is designed for conversations of 10-15 exchanges, enabling users to revisit each state 2-3 times on average and providing sufficient feedback for within-session adaptation. Although prior TigerGPT conversations averaged only 4.9 exchanges, AURA's adaptive design aims to sustain longer, more productive dialogues.

\subsubsection{Hybrid State Representation: Quality Level and Trajectory}

We adopt a hybrid approach that combines absolute engagement level with recent trajectory, yielding five states: \texttt{low\_improving}, \texttt{low\_stable}, \texttt{medium}, \texttt{high\_improving}, and \texttt{high\_stable}. Each state is determined by two factors: the current response's composite quality score ($Q_{\text{composite}}$) and the change in quality relative to the previous exchange ($\Delta Q$):

\begin{equation}
\Delta Q_t = Q_{\text{composite}, t} - Q_{\text{composite}, t-1}
\end{equation}

State assignment follows a hierarchical decision rule. First, quality is categorized into three levels based on thresholds derived from the prior data distribution: low ($Q < 0.3$), medium ($0.3 \leq Q < 0.6$), and high ($Q \geq 0.6$). These thresholds were selected to align with meaningful quality distinctions: the low threshold (0.3) approximates the dataset mean (0.328), capturing below-average responses; the high threshold (0.6) exceeds the 75th percentile (Q75 = 0.520), identifying top-quartile engagement.

For low and high quality levels, trajectory information further refines the state. A quality change exceeding $|\Delta Q| > 0.05$ indicates improving engagement, while smaller changes ($|\Delta Q| \leq 0.05$) reflect stable patterns. The 0.05 threshold was calibrated to detect meaningful shifts without excessive sensitivity to measurement noise: it represents approximately one standard deviation of exchange-to-exchange variability in the prior corpus. Responses in the medium quality range are not subdivided by trajectory, as pilot analysis revealed insufficient distinction between improving and stable medium-engagement patterns given the available data.

The complete state assignment function is:

\begin{equation}
\label{eq:state_assignment}
s_t = \begin{cases}
\texttt{low\_improving} & \text{if } Q_t < 0.3 \text{ and } \Delta Q_t > 0.05 \\
\texttt{low\_stable} & \text{if } Q_t < 0.3 \text{ and } \Delta Q_t \leq 0.05 \\
\texttt{medium} & \text{if } 0.3 \leq Q_t < 0.6 \\
\texttt{high\_improving} & \text{if } Q_t \geq 0.6 \text{ and } \Delta Q_t > 0.05 \\
\texttt{high\_stable} & \text{if } Q_t \geq 0.6 \text{ and } \Delta Q_t \leq 0.05
\end{cases}
\end{equation}

For the first exchange of each conversation, where no previous quality score exists, $\Delta Q$ is defined as zero. This assigns the user to either a ``stable'' state (\texttt{low\_stable} or \texttt{high\_stable}) or to \texttt{medium}, depending solely on initial quality level.


\subsubsection{Rationale for Quality-Trajectory Hybrid}
The hybrid design addresses a fundamental question in adaptive surveying: should the system's next action depend only on how engaged the user currently is, or also on whether engagement is increasing or decreasing? Pure quality-level states (e.g., simply ``low,'' ``medium,'' ``high'') ignore dynamics, treating a user whose engagement is declining identically to one whose engagement is improving—even though these patterns may warrant different questioning strategies. Conversely, pure trajectory-based states (e.g., ``improving,'' ``stable,'' ``declining'') disregard absolute engagement levels, potentially recommending the same action for a slightly improving disengaged user and a slightly improving highly engaged user.

The hybrid approach captures both dimensions. For example, \texttt{\seqsplit{low\_improving}} indicates rising quality from a low baseline, suggesting that recent question types may be resonating with the user. By contrast, \texttt{low\_stable} reflects persistent low engagement, indicating that previous strategies have not improved response quality. Similarly, distinguishing \texttt{high\_improving} from \texttt{high\_stable} enables the system to differentiate between users building momentum and those maintaining consistent engagement—patterns that may respond differently to continued questioning. This two-factor representation provides the reinforcement learning algorithm (Section \ref{sec:rl_framework}) with sufficient information to adapt its strategy based on both current engagement and recent trajectory.

\subsubsection{Empirical Validation and Coverage}
To validate the five-state framework, we categorized all 467 prior responses according to Equation~\ref{eq:state_assignment}. Table~\ref{tab:AURA_state_distribution} presents the resulting distribution. The \texttt{low\_stable} state dominates (46.7\%), reflecting the engagement challenge inherent in voluntary campus-climate surveys: many participants provide minimal responses. The \texttt{medium} state accounts for 27.8\% of responses, representing moderate engagement. High-engagement states (\texttt{high\_improving} and \texttt{high\_stable}) together constitute 16.7\%, confirming that sustained detailed disclosure remains relatively rare in brief survey interactions.

\begin{table}[h]
\centering
\caption{State Distribution in Prior Conversation Data (N = 467 responses).}
\label{tab:AURA_state_distribution}
\begin{tabular}{lrrr}
\toprule
State & Count & Percentage (\%) & Mean Quality Score \\
\midrule
\texttt{low\_stable} & 218 & 46.7 & 0.071 \\
\texttt{medium} & 130 & 27.8 & 0.454 \\
\texttt{high\_improving} & 60 & 12.8 & 0.709 \\
\texttt{low\_improving} & 41 & 8.8 & 0.184 \\
\texttt{high\_stable} & 18 & 3.9 & 0.738 \\
\midrule
Total & 467 & 100.0 & 0.328 \\
\bottomrule
\end{tabular}
\end{table}

Critically, all five states are well-represented in the prior data, ensuring sufficient examples for policy initialization (Section~\ref{sec:rl_framework}). Although \texttt{high\_stable} constitutes only 3.9\% of responses—the smallest category—this proportion still represents 18 observations, adequate for establishing baseline patterns while acknowledging that rare states will benefit most from within-session adaptation as the system encounters individual users.

The state categories exhibit meaningful quality separation. Mean composite scores range from 0.071 (\texttt{low\_stable}) to 0.738 (\texttt{high\_stable}), spanning nearly the full measurement range and confirming that the discretization captures substantive engagement differences rather than arbitrary divisions. The trajectory distinction within low and high ranges further refines these categories: \texttt{low\_improving} (mean = 0.184) falls between \texttt{low\_stable} (0.071) and \texttt{medium} (0.454), consistent with users transitioning from disengagement toward moderate participation. This empirical validation confirms that the five-state representation provides both sufficient coverage of engagement patterns and meaningful differentiation for reinforcement learning.

\subsection{Reinforcement Learning Framework}
\label{sec:rl_framework}
Having established AURA's quality-scoring methodology (Section~\ref{sec:lsde}), action taxonomy (Section~\ref{sec:actions}), and state representation (Section~\ref{sec:states}), we present the reinforcement learning algorithm enabling within-session adaptation. AURA employs a two-level architecture (Figure~\ref{fig:AURA_two_level_learning}): offline learning initializes an expected-value (EV) table from 96 prior conversations; online learning adapts the policy in real time based on each user's responses. This section describes both phases and details question generation.

\begin{table}[htbp]
\centering
\begin{sideways}
\begin{minipage}{\textheight}
\centering
\caption{Initial Expected-Value Estimates from Prior Data (N = 371 exchange pairs)}
\label{tab:AURA_ev_table}
\normalsize
\begin{tabular}{lrrrrr}
\toprule
\textbf{State} & \textbf{Specification} & \textbf{Elaboration} & \textbf{Topic Probe} & \textbf{Validation} & \textbf{Continuation} \\
\midrule
low\_improving & 
    \begin{tabular}[t]{@{}r@{}}0.058\\(n=15, M)\end{tabular} & 
    \begin{tabular}[t]{@{}r@{}}0.047\\(n=9, M)\end{tabular} & 
    \begin{tabular}[t]{@{}r@{}}0.032\\(n=3, L$^\dagger$)\end{tabular} & 
    \begin{tabular}[t]{@{}r@{}}0.000\\(n=0, N$^\ddagger$)\end{tabular} & 
    \begin{tabular}[t]{@{}r@{}}0.000\\(n=0, N$^\ddagger$)\end{tabular} \\
\midrule
low\_stable & 
    \begin{tabular}[t]{@{}r@{}}\textbf{0.288}\\\textbf{(n=112, R)}\end{tabular} & 
    \begin{tabular}[t]{@{}r@{}}\textbf{0.170}\\\textbf{(n=27, R)}\end{tabular} & 
    \begin{tabular}[t]{@{}r@{}}\textbf{0.305}\\\textbf{(n=20, R)}\end{tabular} & 
    \begin{tabular}[t]{@{}r@{}}0.348\\(n=4, L$^\dagger$)\end{tabular} & 
    \begin{tabular}[t]{@{}r@{}}0.476\\(n=1, L$^\dagger$)\end{tabular} \\
\midrule
medium & 
    \begin{tabular}[t]{@{}r@{}}\textbf{0.071}\\\textbf{(n=66, R)}\end{tabular} & 
    \begin{tabular}[t]{@{}r@{}}\textbf{0.073}\\\textbf{(n=28, R)}\end{tabular} & 
    \begin{tabular}[t]{@{}r@{}}0.039\\(n=22, M)\end{tabular} & 
    \begin{tabular}[t]{@{}r@{}}0.000\\(n=0, N$^\ddagger$)\end{tabular} & 
    \begin{tabular}[t]{@{}r@{}}0.000\\(n=0, N$^\ddagger$)\end{tabular} \\
\midrule
high\_improving & 
    \begin{tabular}[t]{@{}r@{}}0.004\\(n=33, M)\end{tabular} & 
    \begin{tabular}[t]{@{}r@{}}0.020\\(n=14, M)\end{tabular} & 
    \begin{tabular}[t]{@{}r@{}}0.000\\(n=4, L$^\dagger$)\end{tabular} & 
    \begin{tabular}[t]{@{}r@{}}0.000\\(n=0, N$^\ddagger$)\end{tabular} & 
    \begin{tabular}[t]{@{}r@{}}0.000\\(n=0, N$^\ddagger$)\end{tabular} \\
\midrule
high\_stable & 
    \begin{tabular}[t]{@{}r@{}}0.040\\(n=9, M)\end{tabular} & 
    \begin{tabular}[t]{@{}r@{}}0.083\\(n=1, L$^\dagger$)\end{tabular} & 
    \begin{tabular}[t]{@{}r@{}}0.028\\(n=3, L$^\dagger$)\end{tabular} & 
    \begin{tabular}[t]{@{}r@{}}0.000\\(n=0, N$^\ddagger$)\end{tabular} & 
    \begin{tabular}[t]{@{}r@{}}0.000\\(n=0, N$^\ddagger$)\end{tabular} \\
\bottomrule
\end{tabular}
\vspace{2mm}

\normalsize
\textit{Note:} Each cell shows EV value with sample size and confidence category in parentheses. Confidence: \textbf{R}eliable ($n \geq 20$), M(oderate) ($5 \leq n < 20$), L(ow)$^\dagger$ ($1 \leq n < 5$), N(one)$^\ddagger$ ($n = 0$, cold-start). EV represents expected quality improvement.
\end{minipage}
\end{sideways}
\end{table}

\subsubsection{Offline Learning: Initial Expected-Value Calculation}
\label{sec:offline}

Before engaging with users, AURA initializes its policy using expected-value estimates derived from 371 prior exchange pairs (Section \ref{sec:data}). Unlike frequency-based approaches that select the most common actions, AURA's EV estimation quantifies each action's effectiveness by combining success probability and improvement magnitude.

\textbf{Expected-Value Formula.} For each state-action pair $(s, a)$:
\begin{equation}
\text{EV}(s, a) = P(\text{improvement} \mid s, a) \times \mathbb{E}[\Delta Q \mid \text{improvement}, s, a]
\end{equation}
where:
\begin{itemize}
\item $P(\text{improvement} \mid s, a)$ is the proportion of instances where action $a$ in state $s$ resulted in quality improvement ($\Delta Q > 0$)
\item $\mathbb{E}[\Delta Q \mid \text{improvement}, s, a]$ is the average quality gain when improvement occurred
\item $\Delta Q = Q_{\text{composite},t+1} - Q_{\text{composite},t}$ is the change in composite quality between consecutive exchanges
\end{itemize}

This formulation captures both the probability and magnitude of quality improvements.

\textbf{Calculation Example.} For the state-action pair (\textit{low\_stable}, \textit{topic probe}):
\begin{itemize}
\item Observations: 20 exchange pairs
\item Improvements: 16 of 20 instances (80\%)
\item Average $\Delta Q$ when improved: $+0.3815$
\item $\text{EV}(\text{low\_stable}, \text{topic\_probe}) = 0.80 \times 0.3815 = 0.305$
\end{itemize}

Table~\ref{tab:AURA_ev_table} presents the complete set of initial EV estimates with sample sizes and confidence indicators. Seventeen of 25 state-action pairs (68\%) have prior data; the remaining 8 pairs (32\%) are initialized to $\text{EV} = 0.0$, representing neutral prior expectation.

\textbf{Coverage and Confidence.} Among pairs with data, five have $n \geq 20$ (reliable estimates), six have $5 \leq n < 20$ (moderate confidence), and six have $n < 5$ (low confidence). The exploration rate ($\epsilon = 0.30$) described in Section \ref{sec:online} ensures that all actions, including those with limited or no prior data, receive trials during individual conversations.

\subsubsection{Online Learning: Within-Session Adaptation}
\label{sec:online}
During each conversation, AURA updates its policy in real time based on 
observed quality changes, enabling rapid personalization. This within-session 
learning combines an $\epsilon$-greedy policy balancing exploration and 
exploitation with an update rule that adjusts EV estimates after each exchange.

\textbf{Reward Signal.} At exchange $t$, the immediate reward is the change in 
composite quality:
\begin{equation}
\label{eq:reward}
r_t = \Delta Q_t = Q_{\text{composite},t} - Q_{\text{composite},t-1}
\end{equation}
For $t=1$, no previous quality exists, so $r_1$ is undefined and no update 
occurs.

\textbf{$\epsilon$-greedy Action Selection.} At exchange $t$, given state $s_t$, 
AURA selects the next action:
\begin{equation}
\label{eq:epsilon_greedy}
a_t = \begin{cases}
\text{random action from } \mathcal{A} & \text{with probability } \epsilon = 0.30 \\
\arg\max_{a \in \mathcal{A}} \text{EV}(s_t, a) & \text{with probability } 1 - \epsilon = 0.70
\end{cases}
\end{equation}
The system explores randomly 30\% of the time, ensuring all actions receive
trials despite prior imbalance (specification 62.3\%; validation 0.9\%, 
continuation 0.4\%). Otherwise, it exploits current knowledge by selecting 
the highest-EV action.

\textbf{EV Update Rule.} After observing reward $r_t$, AURA updates the 
expected value for the state-action pair from the previous exchange:
\begin{equation}
\label{eq:ev_update}
\text{EV}(s_{t-1}, a_{t-1}) \leftarrow \text{EV}(s_{t-1}, a_{t-1}) + \alpha \left[ r_t - \text{EV}(s_{t-1}, a_{t-1}) \right]
\end{equation}
where $\alpha = 0.3$ is the learning rate. This standard temporal-difference 
update \cite{sutton1998reinforcement} occurs at exchange $t$ after observing the 
outcome of action $a_{t-1}$ selected in state $s_{t-1}$ at the previous exchange, 
increasing EV when the observed reward exceeds the prior estimate and 
decreasing it when the action underperforms.

These EV updates are session-specific: each conversation initializes with the prior EV table (Table~\ref{tab:AURA_ev_table}), adapts based on the current user's responses, then discards the modified table at session end, ensuring privacy and individual-focused adaptation.

\textbf{Exchange-by-Exchange Process.} At each exchange $t$, AURA executes the following 
sequence:
\begin{enumerate}
\item Collect and score user response: Compute $Q_t$ via LSDE 
      (Section~\ref{sec:lsde})
\item If $t > 1$: 
      \begin{itemize}
      \item Compute reward: $r_t = Q_t - Q_{t-1}$
      \item Update previous action's value: Apply Equation~\ref{eq:ev_update} 
            to $(s_{t-1}, a_{t-1})$
      \end{itemize}
\item Discretize current state: Map $(Q_t, \Delta Q_t) \rightarrow s_t$ via 
      Section~\ref{sec:states}
\item Select next action: Choose $a_t$ using $\epsilon$-greedy policy 
      (Equation~\ref{eq:epsilon_greedy})
\item Generate question: Produce natural-language prompt 
      (Section~\ref{sec:question_gen})
\end{enumerate}

After 15 exchanges or user termination, the session ends and all within-session 
EV updates are discarded, ensuring the next user begins fresh with priors.


\subsubsection{Question Generation}
\label{sec:question_gen}
The reinforcement learning module selects abstract action types, but users 
interact with natural-language questions. AURA employs GPT-4o-mini to translate 
RL decisions into contextually appropriate prompts.

\textbf{Action-to-Prompt Mapping.} Each action type (Section~\ref{sec:actions}) 
is associated with generation directives that instruct the LLM:
\begin{itemize}
    \item Specification: Request specific examples with contextual details
    \item Elaboration: Request deeper explanation about reasoning or feelings
    \item Topic Probe: Explore related but different aspects of campus life
    \item Validation: Briefly acknowledge contributions (under 20 words)
    \item Continuation: Provide minimal prompt (5-10 words) to sustain flow
\end{itemize}

\textbf{Context Integration.} The prompt includes conversation history 
(previous 2--3 exchanges) and action-specific instructions. For specification 
and elaboration, the LLM deepens the current topic; for topic probe, it 
explores related campus climate dimensions (academic, social, diversity, 
resources, mental health) while maintaining conversational coherence.

\textbf{Generation Parameters.} The LLM uses temperature = 0.7 to generate 
natural variation in phrasing while maintaining contextual appropriateness. 
The system does not employ strict templates; each question is generated 
dynamically based on conversation history and selected action type.

\textbf{Example.} Given action = \textit{topic probe} after discussion about 
Greek life exclusivity, the system generates: \textit{``That's a really 
important perspective. Beyond Greek life, how would you describe the broader 
sense of community and belonging you feel on campus—like in classes, clubs, 
or other social settings?''} The question transitions to related content 
(general social climate) through explicit connection to the previous topic.

By separating strategic decisions (which action) from tactical implementation 
(how to phrase), the system combines RL optimization with LLM fluency.


\subsection{Evaluation Design}
\label{sec:evaluation_design}

\subsubsection{Research Questions \& Hypotheses}
This evaluation validates AURA's within-session learning capability and optimizes the exploration parameter ($\epsilon$) for conversational survey applications. Analysis of the prior TigerGPT dataset (Section~\ref{sec:data}) revealed problematic questioning patterns: over-reliance on specification (62.3\%, Table~\ref{tab:AURA_action-distribution}) and severe under-utilization of validation (0.9\%). These patterns reflect common practices in conversational survey design where systems prioritize information extraction over rapport maintenance. We address two research questions:

\textit{RQ1: Does AURA's RL-based adaptation improve response quality compared to prior TigerGPT behavior?} 

We test whether within-session learning produces significant quality improvements ($p < 0.05$, Cohen's $d > 0.50$) over the non-adaptive prior baseline, measured by quality change from initial to final conversation exchanges.

\textit{RQ2: What exploration rate ($\epsilon$) optimizes within-session learning in brief survey conversations?} 

We compare three $\epsilon$-greedy configurations to identify the exploration-exploitation balance that maximizes quality gains within 15-exchange interactions: conservative fixed ($\epsilon=0.15$), moderate fixed ($\epsilon=0.30$), and decaying ($\epsilon: 0.40 \rightarrow 0.05$).

\subsubsection{Baseline System}
The \textbf{prior distribution baseline} replicates actual TigerGPT behavior from the dataset described in Section~\ref{sec:data}. This semi-adaptive campus climate survey system, deployed October 2024--February 2025, collected the 96 conversations (371 consecutive exchange pairs) used to initialize AURA's expected-value table (Section~\ref{sec:rl_framework}). While TigerGPT employed contextual follow-ups and flexible topic selection, it lacked reinforcement learning and exhibited the action distribution patterns shown in Table~\ref{tab:AURA_action-distribution}: specification 62.3\%, elaboration 23.6\%, topic probe 12.8\%, validation 0.9\%, and continuation 0.4\%.

The baseline implements these empirical action frequencies through weighted random sampling. At each exchange, actions are selected according to the prior probabilities, independent of current engagement state, conversation history, or observed quality patterns. The baseline exhibits \textbf{no state awareness}---all exchanges use identical probabilities regardless of user engagement level---and \textbf{no learning mechanism}---the distribution remains constant throughout and across conversations.

This baseline operationalizes two engagement problems AURA targets: (1) \textbf{specification over-reliance} at 62.3\% risks respondent fatigue through repetitive requests for concrete examples, and (2) \textbf{validation neglect} at 0.9\% undermines rapport-building despite its theoretical importance for sustaining engagement. Improvement over this ecologically valid baseline---representing real survey practice rather than an artificial control---demonstrates that RL adaptation addresses documented methodological challenges in conversational data collection.

\subsubsection{Evaluation Metrics}

\textbf{Primary metric:} Quality Improvement
\begin{equation}
\Delta Q = Q_{\text{final}} - Q_{\text{initial}}
\end{equation}
where $Q_{\text{initial}}$ and $Q_{\text{final}}$ are composite LSDE scores at Exchange~1 and Exchange~15. This metric isolates adaptive contribution by controlling for baseline engagement, directly measuring AURA's core objective of improving response quality through within-session learning.

\textbf{Secondary metrics:} Final quality ($Q_{\text{final}}$), average quality ($\bar{Q}$), and exchange success rate (proportion of exchanges with $\Delta Q > 0$).

\textbf{Action-level analysis:} Distribution of five action types and action appropriateness (proportion of selections with positive expected value in the prior initialization table, Table~\ref{tab:AURA_ev_table}).

\textbf{Temporal analysis:} Early (Exchange~1--5), mid (6--10), and late (11--15) quality trajectories to characterize adaptation dynamics across conversation phases.

\subsubsection{Experimental Design}

We employed a between-subjects factorial design with 4 system conditions (3 RL variants + prior baseline) $\times$ 4 simulated user profiles. Each condition comprised $n = 20$ conversations (5 repetitions per profile), yielding 80 total conversations. All RL configurations shared $\alpha = 0.30$ learning rate, 15-exchange conversations, and within-session learning only (complete EV table reset between users to preserve privacy and ensure individual-focused adaptation).

\textbf{RL Exploration Configurations.} Three epsilon variants tested alternative exploration-exploitation hypotheses:
\begin{itemize}
\item \textbf{Configuration~1 ($\epsilon = 0.15$):} Conservative fixed exploration with steady 15\% random action sampling throughout all exchanges. The hypothesis was that low exploration prevents distraction from learned patterns while allowing occasional discovery of alternatives.

\item \textbf{Configuration~2 ($\epsilon = 0.30$):} Moderate fixed exploration with consistent 30\% random sampling. This higher exploration rate was hypothesized to prevent premature convergence to suboptimal policies, particularly important given sparse prior data (only 17 of 25 state-action pairs have examples, Section~\ref{sec:rl_framework}).

\item \textbf{Configuration~3 ($\epsilon: 0.40 \rightarrow 0.05$):} Decaying exploration schedule with high initial exploration (40\%) decreasing linearly to minimal exploration (5\%) by Exchange~15. At each exchange $t$, $\epsilon_t = 0.40 - (0.35 \times t/15)$. The hypothesis was that early exploration discovers effective actions while late exploitation refines learned strategy.
\end{itemize}

\textbf{Simulated User Profiles.} To enable experimental control, reproducibility, and statistical power while maintaining computational feasibility, we employed AI-simulated users rather than human participants. Simulated users were implemented using GPT-4o-mini (temperature~=~0.8) with profile-specific prompts ensuring consistent persona characteristics. Four profiles were selected to represent diverse disciplinary backgrounds and academic levels commonly found in campus populations:
\begin{itemize}
\item \textbf{Biology Senior:} Science-oriented student with moderate engagement style, focused on academic experiences and research opportunities.

\item \textbf{Psychology Junior:} Social science perspective with high emotional expressiveness and interpersonal focus.

\item \textbf{Computer Science Sophomore:} Technical background with analytical communication style and emphasis on academic rigor.

\item \textbf{English Senior:} Humanities orientation with elaborative responses and attention to campus culture and community.
\end{itemize}

This disciplinary diversity (STEM, social science, humanities) combined with variation in academic seniority (sophomore, junior, senior) ensured that evaluation captured a range of communication styles and engagement patterns typical of university survey populations. Profile characteristics remained constant across all system conditions, ensuring fair comparison and isolating the effect of questioning strategy from individual differences.

The use of simulated users enabled three critical experimental advantages: (1)~precise control, with identical profiles tested across all conditions; (2)~perfect reproducibility, with conversations deterministically reconstructable from logs; and (3)~adequate statistical power ($> 0.80$ for $d \geq 0.65$ at $\alpha = 0.05$, given $n = 20$ per condition). While simulated users cannot capture the full complexity of human behavior, they provide a controlled environment for validating within-session learning mechanisms before deployment with actual survey participants.

\subsubsection{Statistical Analysis.} We employed independent samples $t$-tests comparing each RL configuration to the prior baseline. Three planned comparisons were conducted with uncorrected $p$-values reported ($\alpha = 0.05$, two-tailed). Cohen's $d$ effect sizes were computed for all comparisons, with interpretative guidelines: small ($d = 0.20$--$0.49$), medium ($d = 0.50$--$0.79$), large ($d \geq 0.80$).


The detailed results for AURA are presented in Section ~\ref{sec:AURA_results}.

\chapter{New Intervention AI Methodologies – Early Detection and Assessment for Mental Health}
\label{chap:intervention}
\blfootnote{Portions of this chapter were previously published in \cite{tang2024decoding}, \cite{tang2024advancing}, and \cite{tang2025layered}. © 2024--2025 IEEE. Reprinted, with permission, from J. Tang et al., \textit{2024 IEEE 6th International Conference on Cognitive Machine Intelligence (CogMI)}, Washington, DC, USA, 2024; and \textit{2025 IEEE Conference on Artificial Intelligence (CAI)}, Santa Clara, CA, USA, 2025.}

\section{Introduction to New Early Detection and Assessment Tools}
In addition to improving overall campus satisfaction, another critical dimension of well-being lies in the timely identification and support of individuals facing mental health challenges. By leveraging advanced Natural Language Processing (NLP) and Machine Learning techniques, we can move beyond reactive counseling models to implement more proactive, automated screening and assessment tools. These intervention-focused AI systems aim to detect early indicators of conditions such as depression or anxiety, thereby guiding individuals toward appropriate resources before issues escalate.

Over the course of this research, we developed and evaluated three new AI-driven methods that collectively address diverse aspects of early detection and assessment. The next sections (4.2, 4.3, 4.4) detail the completed work from our previous papers:

\begin{enumerate}
    \item \textbf{Decoding Linguistic Nuances \cite{tang2024decoding}} – Demonstrates how novel methods that analyze Expressive Narrative Stories (ENS) can improve text classification for detecting mental health signals.

    \item \textbf{Psycho Analyst (Custom GPT) \cite{tang2024advancing}} – Introduces a new specialized GPT-4-based model refined with DSM-5 and PHQ-8 guidelines, optimized for pre-screening distress in clinical transcripts.

    \item \textbf{SMMR (Layered Multi-Expert Framework) \cite{tang2025layered}} – Proposes a new multi-step approach that integrates outputs from multiple large language models to handle complex, long-context mental health assessments.
\end{enumerate}


\section{New Linguistic Decoding Methods for Mental Health Clues \cite{tang2024decoding}}
\label{sec:Section 4.2}
Mental health disorders represent a significant public health challenge, impacting over 20\% of adults in the United States each year. Conditions such as depression, bipolar disorder, and anxiety disorders profoundly affect individuals' physical and social well-being \cite{RN679}. This situation is exacerbated by the difficulty in providing effective assistance to those in need, a challenge partly rooted in the stigma surrounding mental health.

The digital age, particularly social media platforms like Twitter, Reddit, and Weibo, has provided novel avenues for mental health research. These platforms reveal insights into individuals' mental states through their natural language expressions, yielding a vast and diverse data source. This study focuses on Expressive Narrative Stories (ENS), a rich, yet under-explored domain within this context. Originating from Pennebaker's concept of Expressive Writing and Narrative Stories \cite{pennebaker1986confronting}, ENS offer profound psychological insights. Documented benefits of expressive writing, a practice involving writing about one's feelings and emotions related to a personal event or interaction, include improved physical health markers like reduced blood pressure \cite{davidson2002expressive} and enhanced immune functioning \cite{pennebaker1988disclosure, rivkin2006effects}, with narrative stories being crucial for processing experiences and constructing coherent self-identities \cite{mcadams2013narrative}.

Despite the richness of ENS and their known benefits, their potential in identifying mental health issues, particularly on social media, is not fully realized. Our study seeks to explore the linguistic nuances of ENS among individuals with and without mental health disorders. We hypothesize that these nuances will offer novel insights into mental health and natural language analysis.

This study will analyze Reddit text posts, contrasting Expressive Narrative Stories (ENS) from individuals with self-declared mental health disorders against those from ostensibly healthy users. Our objectives include: 1) fine-tuning (or training) and evaluating various machine learning models, such as Bidirectional Encoder Representations from Transformers (BERT), MentalBERT, Naive Bayes (NB), Logistic Regression (LR), and Support Vector Machine (SVM). These models are intended to classify text posts based on whether they come from individuals with or without self-declared mental health disorders, helping to identify linguistic features unique to these groups; 2) assessing the significance of topic words in distinguishing depression-related ENS; 3) examining the impact of sentence structure on narrative coherence in ENS; and 4) broadening our analysis to encompass multiple mental health disorders to uncover common linguistic features.

Key contributions of our research to the fields of mental health analysis and natural language processing are as follows:
\begin{itemize}
    \item {\bfseries Focused Analysis on ENS:} By concentrating on ENS within social media, our research fills a significant gap, shedding light on the nuanced ways narrative storytelling intertwines with mental health.
    \item {\bfseries In-depth Linguistic Exploration:} Our comprehensive examination of linguistic features such as topic words and sentence order in ENS establishes a foundation for subsequent studies, underscoring the critical role of narrative context in understanding mental health.
    \item {\bfseries Model Comparative Evaluation:} Our detailed comparison of diverse machine learning models elucidates their strengths and weaknesses in mental health narrative text analysis, guiding future technological developments.
\end{itemize}

\subsection{Data Collection and Preprocessing}

From late 2022 to early 2023, we retrieved Reddit posts spanning several subreddits. Posts from \emph{r/AnxietyDepression} were labeled as target samples (label = 1), while those from other subreddits served as control samples (label = 0). Beyond these core posts, we also gathered data from subreddits devoted to \emph{expressive narrative stories (ENS)}—namely \emph{r/IamA}, \emph{r/relationship}, and \emph{r/AmItheAsshole}—as well as \emph{general narrative stories (GNS)} from subreddits like \emph{r/cooking}, \emph{r/healthyfood}, and \emph{r/eathealthy}. This broader mix of data was intended to strengthen the generalizability of our findings.

In preparing the data, we first converted all text to lowercase. We then filtered out any posts shorter than 10 words or containing redirected URLs (e.g., \emph{http}), since such content often comprised advertisements or otherwise irrelevant material. Certain Unicode artifacts—such as \textbackslash{}u2019 for apostrophes—were replaced with standard punctuation to maintain consistent formatting. We also excluded promotional posts and any ENS posts lacking first-person pronouns, whether they were specific to mental health or not. Finally, we performed a manual review to retain only those entries that aligned with our focus on expressive narratives. \autoref{tab:decoding1} shows an example of raw text after these preprocessing steps.

\begin{table}[!htbp]
\caption{Sample Raw Text}
\label{tab:decoding1}
\begin{tabularx}{\columnwidth}{X}
\hline
\textbf {Sample Post Class \& Content} \\ \hline \hline
\textbf {GNS (Label: 0):} So I had some prime Denver steaks and I cut a couple up and braised them. They were amazing. So much so I went to the grocery store a week later and bought some grocery store choice chuck rib meat and braised that too. It was really good as well. I\textbackslash{}u2019ve only braised meat a handful of times so I don\textbackslash{}u2019t really know any standards. In my mind the prime cuts were better but my question is if it really makes a difference when slow cooking. Assuming I can afford it, am I wasting money braising some prime short ribs or is it really better in your experience? \\ \hline
\textbf {ENS-Common (Label: 0):} We\textbackslash{}u2019ve been in a relationship for almost 2 years now, and for the most part it\textbackslash{}u2019s been healthy \& happy.Unfortunately, before I met him I was in a string of bad relationships with men that cheated on me with their exes and were abusive. My most recent ex compared me to his exes, criticized my body, and more.My bf knows about this history, and I\textbackslash{}u2019ve expressed some discomfort that he still has photos of his ex, still has all his exes numbers in his phone, and also still has the numbers of girls he met from dating apps.I have no reason to distrust him and don\textbackslash{}u2019t want to be controlling. I know I need to work on this for myself. However, it hurts that he doesn\textbackslash{}u2019t seem to care about helping me feel more at ease when those small gestures would help. \\ \hline
\textbf {Depression-ENS (Label: 1):} I don't really know what to say here, just needed a place vent.Ive struggled with anxiety my whole life and depression for most of it. There's way more context to it than this but my family is made of well know scholars, both mom and dad are full professors at their universities, sister(27) is a honors master's student, aunt retired full professor, younger cousin(22) majoring in civil engineering and his brother(24), who's a mechanic engineer, just got in the best master program of our country. All of them went to either one of the two best federal university here in Brazil. I'm a 25yo communications student at a private university who's still one year away from graduating and decided not to do a masters for now. By all accounts my family see me as a deadbeat that has no future what so ever and makes zero effort to hide or constrain their opinion. Left my dads house a few years ago and had to return because life happen; The pandemic stretched my stay here from 1 year to 3 now and my relationship with my dad is at the limit. I don't have any idea on wtf I'm doing with my life and I'm starting to give up on everything.  \\ \hline
\end{tabularx}
\end{table}

Using the resulting corpus, we formed three main datasets to train and validate our models:

\begin{itemize} 
    \item \textbf{GNS-Depression Dataset:} Containing 1650 general narrative posts and 1076 posts from \emph{r/AnxietyDepression}.
    \item \textbf{ENS-Depression Dataset:} Combining 1110 expressive narrative posts (from \emph{r/IamA, r/relationship, r/AmItheAsshole}) with 1076 posts from \emph{r/AnxietyDepression}.
    \item \textbf{Mix-Depression Dataset:} Merging the two sets above, including repeated \emph{r/AnxietyDepression} posts in both. 
\end{itemize}

\autoref{tab:decoding2} presents descriptive statistics for these datasets, including label distributions and word count patterns. To ensure randomization and balanced labeling, each dataset was shuffled and then split into training (70\%) and validation (30\%) subsets, while preserving the same ratio of target labels in each subset.

\begin{table}[!t]
\centering
\begin{threeparttable}
\fontsize{12pt}{16pt}\selectfont
\setlength{\tabcolsep}{18pt}
\caption{Description of Training and Validation Datasets}
\label{tab:decoding2}
\begin{tabular}{l|ccc}
\hline
\textbf{Attribute} & \textbf{GNS-Dep.} & \textbf{ENS-Dep.} & \textbf{Mix-Dep.} \\ \hline
No. of Posts       & 2721           & 2186            & 3836           \\
Avg. Words/Post    & 99             & 145             & 122            \\
Max Word Count     & 1111           & 751             & 1111           \\
Ratio of Label     & 1.53           & 1.03            & 2.57           \\ \hline
\end{tabular}
\smallskip
\begin{tablenotes}
\item Note: Dep. is the abbreviation for depression
\end{tablenotes}
\end{threeparttable}
\end{table}

Additional testing sets were developed by gathering new Reddit posts from the same (or closely related) subreddits. These included newly collected ENS and GNS content, as well as posts from \emph{r/Anxiety}, \emph{r/Bipolar}, and a Kaggle-sourced \emph{SuicideWatch} dataset \cite{Rigoulet2021}. In all, 12 additional testing and generalization sets were built. We applied the same preprocessing steps to these sets, ensuring consistency across all phases of experimentation. The ratio of target to control posts in these testing sets was set to 1:1, allowing for a straightforward evaluation of classifier performance on each distinct group. \autoref{tab:decoding3} summarizes their key characteristics.

\begin{table}[!t]
\centering
\fontsize{12pt}{16pt}\selectfont
\setlength{\tabcolsep}{18pt}
\caption{List of Testing and Generalization Set}
\label{tab:decoding3}
\begin{tabular}{lc}
\hline
\textbf{Dataset Name} & \textbf{Number of Posts} \\ \hline
GNS-Depression   & 3562            \\
ENS-Depression & 3070            \\
Mix-Depression       & 3976            \\
GNS-Anxiety      & 3565            \\
ENS-Anxiety    & 3070            \\
Mix-Anxiety          & 4220            \\
GNS-Suicide      & 3562            \\
ENS-Suicide    & 3070            \\
Mix-Suicide          & 6632            \\
GNS-Bipolar      & 3048            \\
ENS-Bipolar    & 3048            \\
Mix-Bipolar          & 3048            \\ 
\hline
\end{tabular}
\end{table}

\subsection{Decoding Linguistic Nuances Methodology}

The pursuit of high model accuracy in mental health detection often overlooks the importance of using data that genuinely reflects personal emotional expression. Generic datasets—such as news articles or jokes—rarely capture the rich, first-person perspectives needed to identify nuanced mental health signals. In response, our approach prioritizes narratives that contain personal and emotional experiences, as these offer deeper insights into an individual’s mental state. By distinguishing between general narratives and those that explicitly convey personal or emotional content, we aim to uncover linguistic features that are more predictive of mental health conditions. Focusing on these expressive narratives not only addresses gaps in previous research but also bolsters real-world applicability, particularly in early screening situations where users may not explicitly discuss their mental health challenges.

In this study, we undertook a three-phase process to analyze linguistic features associated with Expressive Narrative Stories (ENS) and more conventional narratives. After initially collecting three sets of Reddit data—one featuring ENS related to depression (Depression), another containing general ENS (ENS), and a third comprising generic narratives (GNS)—we proceeded with the following steps:

\begin{enumerate} 
    \item \textbf{Classifier Implementation and Evaluation:} We developed and tested multiple machine learning models, including BERT, MentalBERT, SVM, Naive Bayes, and Logistic Regression, to determine which yielded the best performance on our datasets. The primary objective was to accurately distinguish posts authored by individuals with self-declared mental health disorders from those by users who appear not to have such conditions. This step provided an essential foundation for subsequent, more detailed analyses of language use and mental health indicators within Reddit posts.
    
    \item \textbf{Impact of Topic Words:} We examined how classifiers performed under different manipulations of \textit{topic words}—highly relevant or frequently used terms within the narratives—to understand each model’s sensitivity to specific lexical cues. By assessing changes in classification results when certain words were removed or replaced, we gained insights into how strongly these models relied on key terms rather than overall context.
    
    \item \textbf{Logical Connection Analyses:} We then investigated how modifications to sentence order affected classifier performance. Specifically, we looked at the coherent and meaningful relationships that bind sentences within a narrative and tested how well both BERT and MentalBERT handled variations introduced by within-post and cross-post sentence shuffling. These manipulations simulate the cognitive disruptions that are sometimes observed in individuals experiencing mental distress. Additionally, we assessed how these findings generalized to other mental health-related expressive narratives by observing model behavior under conditions of narrative disorganization. 
\end{enumerate}

\subsubsection{Phase 1: Fine-Tuning and Comparing Models}

We evaluated two transformer-based models—\textit{BERT-base} \cite{devlin2018bert} and \textit{MentalBERT} \cite{ji2021mentalbert}—alongside traditional machine learning approaches (Naïve Bayes, Logistic Regression, and SVM).

\paragraph{Model Setup.}
Both BERT-base and MentalBERT have 12 hidden layers, 12 attention heads, and a hidden size of 768. We employed their built-in tokenizers with WordPiece subword segmentation, retaining stopwords for completeness. Training hyperparameters included: Batch size: 32; Epochs: 5; Learning rate: 2e-5, optimized with Adam; Activation: Sigmoid.
We tested input sequence lengths of 10, 64, 128, and 300 tokens to determine optimal context size.

\paragraph{Traditional Classifiers.}
For benchmarking, we also trained Naïve Bayes, Logistic Regression, and SVM on the same training and test splits, using either TF-IDF (up to 5000 terms) or Unigram features. Comparing these classical methods against the transformer-based architectures highlights each model’s relative strengths in handling the complex, expressive narratives central to our study.

\subsubsection{Phase 2: Topic-Word Manipulations}
Building on the classifier configurations and baseline comparisons established in Phase 1, we next examined how strongly the models depend on explicit vocabulary cues, particularly in scenarios where users may not explicitly mention mental health symptoms. To do this, we analyzed the impact of altering or removing key topic words across various test sets (Depression, ENS, GNS).

Using a WordCloud-based approach, we identified ten highly influential topic words for each training dataset, alongside common variations. We then applied two main manipulations (see \autoref{tab:decoding5}):
\begin{enumerate}
    \item Words Removing: Eliminating all instances of these words in the test sets.
    \item Words Replacing: Substituting those words with a neutral token (e.g., “nothing”).
\end{enumerate}

\begin{table}[!t]
\centering
\fontsize{12pt}{16pt}\selectfont
\caption{Example of Topic Word Manipulations}
\label{tab:decoding5}
\begin{tabular}{p{5cm}|p{8cm}} 
\hline
\textbf{Words Process}                & \textbf{Sentence}                                        \\ \hline
Raw                                   & i have very good relationship with my friend. \\ \hline
Words Removing                        & i have very good with my.                     \\ \hline
Words Replacing to “nothing” & i have very good nothing with my nothing.    \\ \hline
\end{tabular}
\end{table}

By tracking changes in accuracy and F1 scores, we sought to determine whether the models and traditional classifiers—originally configured and compared in Phase 1—primarily rely on specific keywords or broader contextual cues. This step is especially important for real-world applications, where individuals experiencing mental distress may avoid using explicit mental health terminology.

\subsubsection{Phase 3: Logical Connection Analyses}
After examined how explicit word usage affects classification in Phase 2, we next turned our attention to the underlying coherence of expressive narratives—an area equally critical for detecting mental health signals. Research suggests that cognitive disruptions, often associated with conditions like anxiety or depression, can alter an individual’s ability to maintain logical narrative flow, leading to disorganized sentences or thematic jumps \cite{jung2014emotions, videbeck2010psychiatric}.

\begin{figure}[t]
  \centering
  \includegraphics[width=\columnwidth]{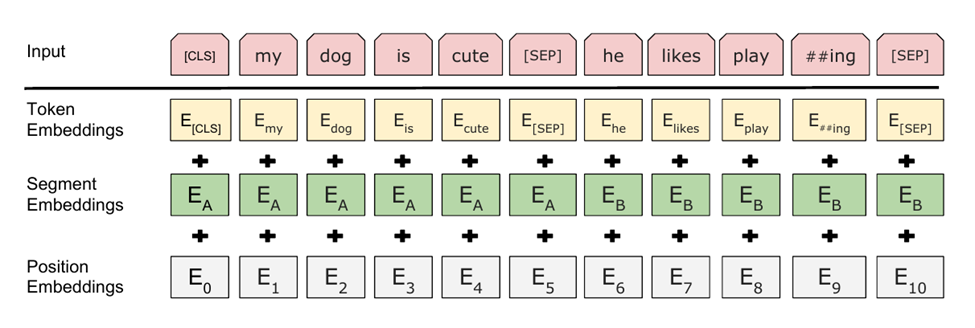} 
  \caption{Illustration of BERT Embeddings \cite{devlin2018bert}}
  \label{fig:Decoding_BERT}
\end{figure}

To evaluate how BERT and MentalBERT handle these coherence challenges, we leveraged their bidirectional attention (\autoref{fig:Decoding_BERT}), which integrates token, segment, and position embeddings. This capacity to read “both ways” is vital for detecting subtle disruptions in narrative flow—disruptions that may mirror real-life conditions such as severe anxiety.

\begin{figure}
  \centering
  \includegraphics[width=\columnwidth]{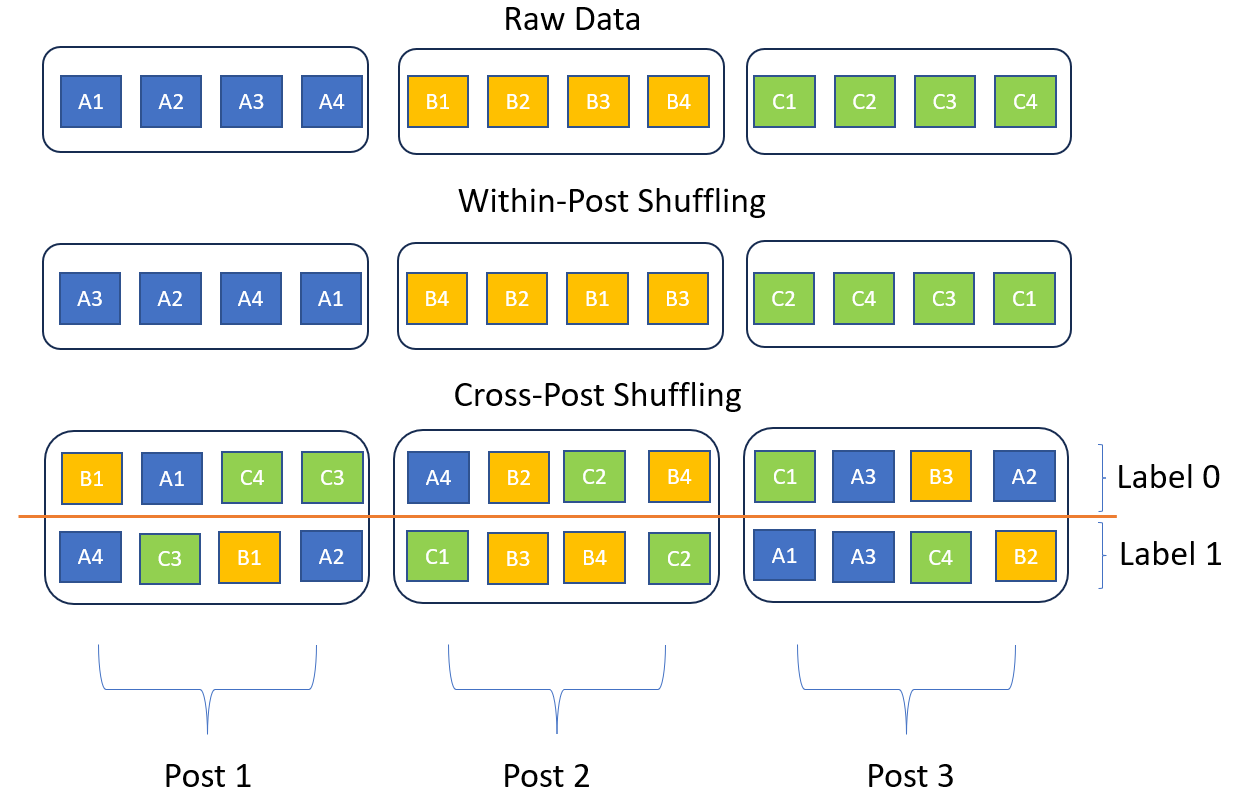} 
  \caption{Illustration of Two Sentence Shuffling Manipulations}
  \label{fig:Decoding_sentence_random}
\end{figure}

We introduced two sentence-shuffling techniques, illustrated in \autoref{fig:Decoding_sentence_random}, across all 12 testing sets:
\begin{itemize}
    \item \textbf{Within-Post Shuffling:} Randomly permutes sentences within a single post, simulating moderate logical breaks.
    \item \textbf{Cross-Post Shuffling:} Mixes sentences from posts sharing the same label, creating severe narrative disorganization.
\end{itemize}

Consistent with our findings from Phase 1, we limited our experiments to BERT(128) and MentalBERT(128) due to their superior performance in capturing nuanced context. We also considered additional mental health categories—Anxiety, Bipolar, and Suicide—alongside the control sets (GNS, ENS, Mix), creating nine additional testing scenarios that allowed for broader generalizability. Full details on these 12 testing sets are provided in \autoref{tab:decoding3}.

In essence, this phase tested how robustly our models, originally optimized in Phase 1 and validated in Phase 2, could handle disruptions in sentence order—a critical step toward real-world applications where users’ narratives may lack coherent flow.

The detailed result of this three-phase method will be presented in Chapter~\ref{chap:evaluation}.


\section{Mental Health Detection Using New Custom GPT Models \cite{tang2024advancing}}
\label{sec:Section 4.3}
While transformer-based classifiers (e.g., BERT, MentalBERT) can excel at classifying mental health text data, they often function as “black boxes,” providing limited interpretability for end users or clinicians. This limitation is especially problematic when early detection tools need to convey not just that an individual may be experiencing distress, but also why the model reached such a conclusion—information that can be critical for mental health professionals’ decision-making.

To address these challenges, we adopt advanced GPT-based approaches that are not reliant on traditional fine-tuning methods. Instead, our method leverages prompt engineering and carefully designed domain adaptations informed by the Diagnostic and Statistical Manual of Mental Disorders, Fifth Edition (DSM-5) \cite{edition2013diagnostic} and the Patient Health Questionnaire-8 (PHQ-8) guidelines \cite{Kroenke2009PHQ8}. By embedding these clinical criteria into GPT’s generative framework, the model can produce natural language “rationales” or short explanations alongside classification outputs. This dual capability helps bridge the gap between robust text classification and real-world applicability, allowing campus counselors, administrative staff, or research teams to better understand how the AI system interprets user narratives.

In what follows, we describe how custom GPT architectures (1) identify mental health indicators from users’ longer, expressive submissions, and (2) provide context or explanatory text that can support more immediate, transparent, and actionable referrals to mental health resources. We will highlight how these GPT-based models have been validated on clinical-style datasets (e.g., DAIC-WOZ) and discuss the potential for integrating their explainable feedback into user well-being frameworks.

This new work contributes to the field in several areas:
\begin{enumerate}
\item \textbf{Innovative Model Development:} We developed 'Psycho Analyst,' a custom GPT model optimized for mental health pre-screening, utilizing OpenAI's ChatGPT-4 service. This model uniquely integrates the DSM-5 and PHQ-8, both of which are globally recognized for their clinical validity. Additionally, it incorporates detailed data descriptions and training data from the DAIC-WOZ database. This comprehensive integration significantly enhances the model’s capacity to accurately interpret nuanced language, enabling it to effectively identify various mental health conditions.
\item \textbf{Dual Task Framework:} 'Psycho Analyst' operates through a sophisticated dual-task framework designed to handle both classification of mental health status and computation of PHQ-8 scores. This dual approach allows for a more comprehensive evaluation of mental health conditions, offering both binary classification and detailed severity assessments.
\item \textbf{Empirical Evaluation and Validation:} The Psycho Analyst model has been rigorously validated using the DAIC-WOZ clinical transcript dataset. Our evaluations confirm the model's effectiveness in real-world clinical environments, reinforcing its practical applicability for mental health diagnostics. This empirical assessment not only underscores the robustness of Psycho Analyst but also highlights its superior performance in comparison to other large language models such as GPT-4o and Mixtral-8*7B.
\item \textbf{Innovative Application in Learning Scenarios:} The model's superior performance in zero-shot and few-shot learning scenarios highlights its capability to adapt to varied data with minimal training. This feature is particularly valuable in mental health settings where the model can quickly adjust to the nuances of different patient interactions without extensive retraining.
\item \textbf{New potential for Public Mental Health:} By successfully integrating DSM-5 and PHQ-8 criteria into the generative AI analysis process, the Psycho Analyst model opens new possibilities for early intervention and tailored mental health care. This integration enables more precise, personalized, and cost-effective assessments, significantly improving the potential for early detection and customized treatment plans in public health settings.
\end{enumerate}

\subsection{Data Description and Preprocessing}
In this paper, we utilized the DAIC-WOZ dataset, a subset of the larger Distress Analysis Interview Corpus (DAIC) \cite{gratch2014distress}, extensively described in \cite{devault2014simsensei}. This dataset is distinguished by its integration of audio and video recordings alongside comprehensive questionnaire responses, obtained from Wizard-of-Oz style interviews. These interviews were facilitated by 'Ellie,' an animated virtual interviewer operated remotely by a human interviewer. The primary focus of the DAIC-WOZ database lies in its clinical interviews, which are specifically designed to facilitate the diagnosis of psychological distress conditions such as anxiety, depression, and PTSD.
\begin{table}[t]
\centering
\caption{Number of Participants in Subsets of the DAIC-WOZ Dataset and the Numbers of Participants with PHQ-8 Scores \textless or \textgreater= 10.}
\label{table:Advanced_table1}
\begin{tabular}{l|ccc}
\hline
\textbf{} & \textbf{Overall} & \textbf{Training } & \textbf{Test} \\
\hline
NoP (No. of Participants) & 187 & 105 & 47 \\

NoP with PHQ-8 Scores \textgreater=10 (A)& 55 & 30 & 14 \\

NoP with PHQ-8 Scores \textless10 (B)& 132 & 75 & 33 \\

Ratio A/B & 0.42 & 0.40 & 0.42 \\
\hline
\end{tabular}
\end{table}

The dataset encompasses 187 interview entries that are divided into 105 records for training, 35 for validation, and 47 for testing.
\autoref{table:Advanced_table1} shows the number of participants with PHQ-8 scores smaller than or equal or larger than 10, respectively, in different subsets of the DAIC-WOZ dataset. Patient Health Questionnaire-8 (PHQ-8) is a standard depression screening metric.
 
\begin{table}[t]
\centering
\caption{Sample Excerpts of Raw Transcript from the Interactions in DAIC-WOZ Dataset.}
\label{tab:Advanced_table2}
\begin{tabularx}{\columnwidth}{@{}lX@{}}
\toprule
Speaker     & Dialogue \\ \midrule
Ellie       & hi i'm ellie thanks for coming in today \\
Ellie       & i was created to talk to people in a safe and secure environment \\
Ellie       & think of me as a friend i don't judge i can't i'm a computer \\
Ellie       & i'm here to learn about people and would love to learn about you \\
Ellie       & i'll ask a few questions to get us started and please feel free to tell me anything your answers are totally confidential \\
Ellie       & how are you doing today \\
Participant & good \\
Ellie       & that's good \\
Ellie       & where are you from originally \\
Participant & atlanta georgia \\
Ellie       & really \\
Ellie       & why'd you move to l\_a \\
Participant & um my parents are from here um \\
Ellie       & how do you like l\_a \\
Participant & i love it \\
Ellie       & what are some things you really like about l\_a \\
Participant & i like the weather \\
Participant & i like the opportunities \\
Participant & um \\
Participant & yes \\ \bottomrule
\end{tabularx}
\end{table}
 
\autoref{tab:Advanced_table2} shows sample excerpts of raw transcript from the interactions between the virtual interviewer--Ellie and the participants from the DAIC-WOZ dataset.

\begin{figure}[t]
  \centering
  \includegraphics[width=0.9\columnwidth]{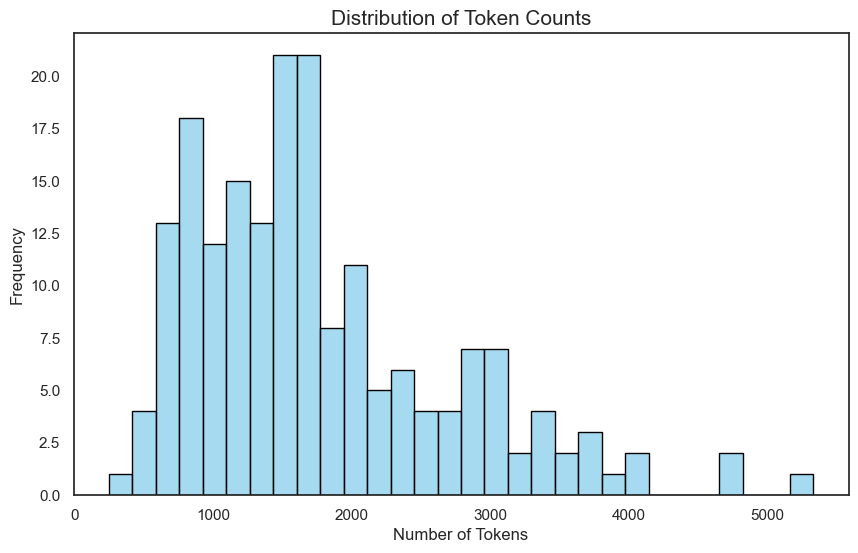} 
  \caption{Histogram of ChatGPT token counts of combined participant dialogues in the DAIC-WOZ dataset. }
  \label{fig:Advanced_Token}
\end{figure}

To prepare the data for the development and evaluation of the Psycho Analyst model, we extracted textual responses from participants within the DAIC-WOZ dataset. These responses were concatenated to form a continuous text sequence for each participant. This method ensures that the context of each question and response is maintained, which is crucial for the model's understanding and analysis. To enhance the preservation of question context and minimize potential misinterpretations, we included markers ("./ ") to separate response within the concatenated sequences. This adjustment aims to provide clearer separations between distinct conversational turns, thereby aiding the model in its contextual comprehension. \autoref{tab:Advanced_table3} shows samples of these concatenated dialogues of 10 participants. Moreover, the distribution of ChatGPT token counts of combined participant dialogues in the DAIC-WOZ dataset is depicted in \autoref{fig:Advanced_Token}.

\begin{table}[t]
\centering
\caption{Sample Raw Participant Dialogue Compilation Segment}
\label{tab:Advanced_table3}
\begin{tabularx}{\columnwidth}{@{}lX@{}}
\toprule
Participant ID & Joint Participant Dialogue Text \\
\midrule
300 & good./ atlanta georgia./ um my parents are from here... \\
301 & thank you./ mmm k./ i'm doing good thank you./ i'm ... \\
302 & i'm fine how about yourself./ i'm from los angeles... \\
303 & okay how 'bout yourself./ here in california. yeah. ... \\
304 & i'm doing good um./ from los angeles california./ ... \\
305 & i'm doing alright./ uh originally i'm from california./ uh... \\
306 & fine./ uh colorado./ mhm./ uh career./ career ... \\
307 & \textless laughter\textgreater./ um moscow./ um my family moved to... \\
308 & los angeles california./ yes./ um the southern... \\
309 & \textless laughter\textgreater./ \textless laughter\textgreater yeah\textless laughter\textgreater./ ... \\
\bottomrule
\end{tabularx}
\end{table}

\subsection{Custom GPT Methodology}

We developed 'Psycho Analyst,' a custom GPT utilizing OpenAI's ChatGPT-4 service, tailored for mental health pre-screening. This model is designed to identify signs of mental health disorders such as anxiety, depression, stress, and bipolar within textual data. It integrates the Diagnostic and Statistical Manual of Mental Disorders, Fifth Edition (DSM-5), and the Patient Health Questionnaire-8 (PHQ-8) as foundational elements. These standards are globally recognized for their clinical validity and provide structured diagnostic criteria, thereby enhancing the model's capability to accurately interpret nuanced language indicative of mental health conditions.

\textbf{Dual Task Framework:} Psycho Analyst operated through structured prompts that guided its analytical processes. Its evaluation framework was based on two distinct tasks: (1) classification and (2) PHQ-8 score computation. Its performance was evaluated using various settings of background knowledge, as discussed later.

\subsubsection{Task 1: Classification}
To accurately reflect the complexities of mental health assessments, we used a nuanced 7-point scale for classifying mental health conditions. This scale ranged from 1, indicating 'not at all likely,' to 7, representing 'extremely likely.' This method allowed for a more detailed representation of the likelihood of a condition, accommodating the often nuanced and uncertain nature of mental health diagnostics. The final classification into binary outcomes was computed based on a threshold value determined through careful evaluation of the likelihood scores.

The single comprehensive prompt guided the model to:
\begin{itemize}
\item \textbf{Assess Mental Health Status:} Determine whether each patient, referenced by their ID, was likely to have mental health issues. This included a detailed analysis of the conditions present, using linguistic indicators found in the text data.
\item \textbf{Quantify Condition Severity:} Calculate the severity on a 7-point scale. This approach provided a more refined assessment compared to a simple binary outcome, better capturing the subtleties and uncertainties inherent in mental health evaluations.
\end{itemize}

This framework enhanced the model’s ability to conduct detailed and context-sensitive analyses, furthering the application of generative AI in mental health evaluations.

\autoref{tab:Advanced_table4} presents examples of the model's outputs, illustrating how Psycho Analyst processed clinical interview transcripts to assess mental health conditions and assign likelihood scores. These examples emphasize the model's capability to move beyond mere yes/no classifications, thus accommodating the often ambiguous and complex nature of psychological symptoms for more clinically relevant and precise evaluations.
\begin{table}[t!]
\centering
\caption{Sample Psycho Analyst GPT Output.}
\label{tab:Advanced_table4}
\begin{tabularx}{\columnwidth}{@{}X@{}}
\toprule
To assess the mental health status of each participant, I will analyze their responses using my knowledge of psychological patterns and criteria associated with mental health issues, specifically focusing on signs of anxiety, depression, stress, and bipolar disorder. It's important to remember that these assessments are not definitive diagnoses but rather interpretations based on the provided text. Let's proceed with each participant.\\
\hline

\textbf{Participant 300:}
Mental Health Assessment: The participant seems to exhibit a balanced outlook on life, with no overt signs of distress in their responses. They mention feeling frustrated occasionally but also engage in positive activities and have supportive family relationships.
Likelihood of Ongoing Mental Health Disorder: \textbf{2/7}. The participant appears to have some minor irritations but overall demonstrates a healthy psychological profile.

\textbf{Participant 306:}
Mental Health Assessment: The participant discusses past challenges, including a legal issue and the implications of peer pressure on their child. They also mention feeling stressed and fatigued but do not exhibit strong signs of depression or severe anxiety.
Likelihood of Ongoing Mental Health Disorder: \textbf{3/7}. While there are signs of stress and past trauma, there's not enough evidence to suggest a severe ongoing mental health disorder.

\textbf{Participant 308:}
Mental Health Assessment: This participant displays several signs of depression, including joblessness, feelings of hopelessness, sleep disturbances, and a history of therapy for mood issues. The recent loss of a family member and homelessness add to the severity of their situation.
Likelihood of Ongoing Mental Health Disorder: \textbf{6/7}. The participant's current life circumstances and psychological state strongly indicate ongoing depression.

\textbf{Participant 309:}
Mental Health Assessment: The participant shows variability in mood and mentions a history of depression. They also speak about anger management issues and recent life changes (wife leaving). This fluctuation in mood and emotional response could indicate bipolar tendencies or severe stress.
Likelihood of Ongoing Mental Health Disorder: \textbf{5/7}. The participant's history and current emotional state suggest a significant possibility of ongoing mental health issues, possibly bipolar disorder or depression.

\textbf{Participant 311:}
Mental Health Assessment: The participant talks about experiences with depression, hospitalization for mental health issues, and auditory hallucinations. They also mention ongoing struggles with sleep and mood. These are strong indicators of a serious mental health condition, likely chronic in nature.
Likelihood of Ongoing Mental Health Disorder: \textbf{7/7}. The participant's history and current symptoms strongly suggest an ongoing and severe mental health disorder. \\
\hline

\textit{These assessments are based on the information provided and should be taken as preliminary insights. It's crucial for individuals showing signs of mental health issues to seek professional help for accurate diagnosis and appropriate treatment.}
\\
\bottomrule
\end{tabularx}
\end{table}

\subsubsection{Task 2: PHQ-8 Score Computation}
\begin{figure*}[h!]
  \centering
  \includegraphics[width=1\linewidth]{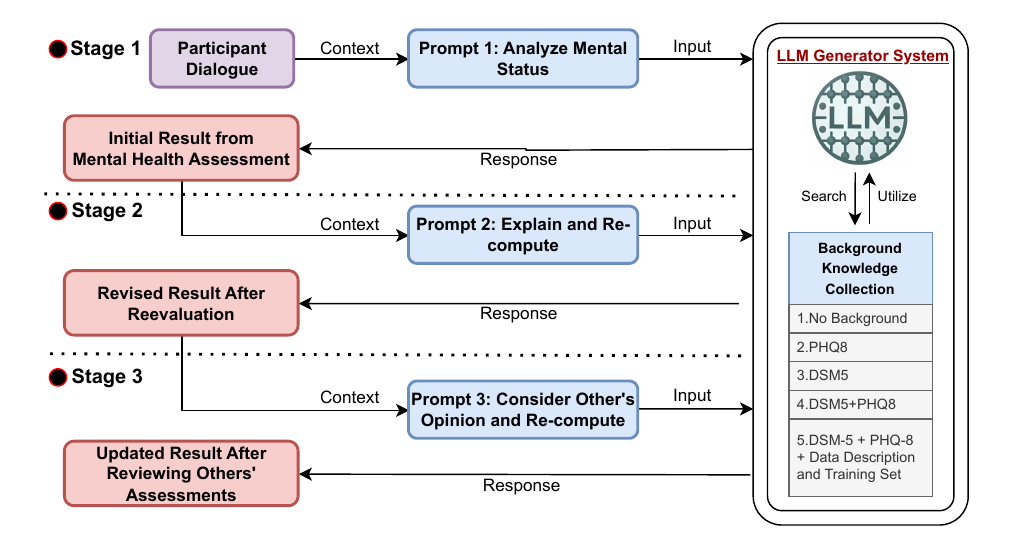}
  \caption{Three-Stage Mental Health Evaluation Framework: Comprehensive Analysis, Re-Evaluation, and Independent Validation of Others' Opinions.}
  \label{fig:Advanced_jtllm_3}
\end{figure*}

The PHQ-8 assessment task, a comprehensive three-step process, evaluates how effectively our model, Psycho Analyst, analyzes mental health symptoms from individual transcripts. A detailed visual representation of this task is shown in Figure \ref{fig:Advanced_jtllm_3}. The PHQ-8 tool measures mental health challenges on a scale from 0 to 24. Scores of 10 or higher might indicate significant mental health concerns, while scores of 20 or higher suggest more severe issues.

\paragraph{Stage 1: Initial Analysis}
In the first stage, Psycho Analyst is tasked with conducting a detailed analysis of a clinical interview transcript to determine the mental health status of an individual. This analysis is comprehensive, utilizing both the content of the transcript and the specific evaluation criteria outlined in the background materials provided. The evaluation focuses on two primary objectives:
\begin{itemize}
\item \textbf{Mental Health Assessment:} Analyze the transcript to determine if the individual has mental health issues, providing a reasoned explanation for the assessment. The analysis should cite relevant portions of the transcript and illustrate how these align with the criteria provided in the background knowledge, including the PHQ-8 items.
\item \textbf{PHQ-8 Score Estimation:} Estimate the PHQ-8 score for the individual based on the transcript's content and the evaluation guidelines from the background materials. The PHQ-8 score is crucial for assessing the severity of mental health symptoms.
\end{itemize}

\paragraph{Stage 2: Detailed Breakdown}
In the second stage, the model is asked to break down how it derives the PHQ-8 score for each symptom. This involves a step-by-step explanation of the scoring process, ensuring transparency and clarity in how the PHQ-8 score is computed based on the individual's responses in the transcript.

\paragraph{Stage 3: Independent Assessment}
The third stage evaluates the model's capacity for independent reasoning. Psycho Analyst is presented with a PHQ-8 score previously assigned to an individual by another evaluator, extracted from the original dataset. The tasks for the model in this stage are as follows:
\begin{itemize}
\item \textbf{Agree or Disagree:} Determine whether it agrees with the assigned PHQ-8 score.
\item \textbf{Explain Reasoning:} Provide a detailed explanation for its agreement or disagreement with the assigned score.
\item \textbf{Reanalyze and Discuss:} Reanalyze the PHQ-8 score, discussing the implications of the score and any potential discrepancies with the initial assessment.
\end{itemize}
\autoref{tab:Advanced_table_3stage} shows an abbreviated simple output of this task.
\begin{longtable}{c|p{0.8\textwidth}}
\caption{Sample 3-Stage Assessment Output}
\label{tab:Advanced_table_3stage}\\
\toprule
\textbf{Stage} & \textbf{Output} \\
\endfirsthead

\toprule
\textbf{Stage} & \textbf{Output} \\
\endhead

\hline
\multicolumn{2}{r}{\textit{(Continued on next page)}}\\
\endfoot

\bottomrule
\endlastfoot
Stage 1 & \textbf{Signs of Symptoms:}\\
 & \textit{Motivation and Interest:} The individual frequently expresses a lack of motivation and interest in previously enjoyed activities, stating: ``I find it hard to relax, I don’t have the same outlets that I did before... I’ve lost my motivation.''\\
 & \textit{Feelings of Hopelessness:} There are multiple indications of feeling down and hopeless: ``I've kind of lost myself... I don’t really have anybody in my life who I think cares enough about me.''\\
 & \textit{Fatigue and Energy:} Persistent fatigue and low energy are noted: ``I get inspired every once in a while but overall I just don’t seem to be making much progress.''\\
 & \textit{Self-Worth:} Thoughts of worthlessness or feeling like a failure are prevalent: ``I don’t really have any pursuits... I don’t have much of a sense of pride anymore.''\\
 & \textit{Concentration:} Difficulty concentrating is evident: ``I’ve lost quite a bit of that... whatever. I used to... help people and now I don't seem to be able to help myself.''\\
 & \textbf{Estimated PHQ-8 Score:} 2+3+1+2+1+2+1+1=13 (indicating moderate depressive symptoms)\\
 & \textbf{Summary:} The individual exhibits substantial symptoms of depression, with significant impairment in social and occupational areas. Based on the transcript and using the DSM-5 and PHQ-8 criteria, the likelihood of an ongoing mental health disorder is high. Professional mental health support is recommended for a comprehensive evaluation and appropriate intervention. \\
\hline
Stage 2 & \textbf{PHQ-8 Scoring Breakdown:}\\
 & \textit{Little interest or pleasure in doing things:} ``I used to do a lot of things lately... I find it hard to relax. I don't have the same outlets that I did before. I don't pursue them as much anymore.'' \textbf{Score:} 2 (More than half the days)\\
 & \textit{Feeling down, depressed, or hopeless:} ``I'm depressed about it... There it is right there.'' and ``I've kind of lost myself... I used to think I had a purpose and such, but I don't really have anybody in my life who I think cares enough about me.'' \textbf{Score:} 3 (Nearly every day)\\
 & \textit{Trouble falling or staying asleep, or sleeping too much:} ``Yeah, when I get a chance to fall asleep, I don't have any problem with it.'' \textbf{Score:} 0 (Not at all)\\
 & \textit{Feeling tired or having little energy:} ``I've lost quite a bit of that motivation.'' and ``I get inspired every once in a while but overall I just don't seem to be making much progress.'' \textbf{Score:} 2 (More than half the days)\\
 & \textit{Feeling bad about yourself, or that you are a failure, or have let yourself or your family down:} ``I don't really have any pursuits... I don't have much of a sense of pride anymore.'' and ``I've lost my motivation... I'm depressed about it, you know.'' \textbf{Score:} 2 (More than half the days)\\
 & \textit{Trouble concentrating on things, such as reading the newspaper or watching television:} ``I don't seem to be able to help myself.'' \textbf{Score:} 1 (Several days)\\
 & \textbf{Total PHQ-8 Score:} 2+3+0+2+0+2+1+0 = 10\\
 & \textbf{Summary:} The individual exhibits symptoms consistent with moderate depression, as indicated by a PHQ-8 score of 10. This suggests a moderate to high probability of an ongoing depressive disorder. Professional evaluation and intervention are recommended.\\
\hline
Stage 3 & \textbf{Re-analysis of PHQ-8 Scoring :}\\
 & Let's reassess the PHQ-8 score in light of the other evaluator's assigned value of 5 and compare it to my initial assessment....\\
 & \textbf{Discussion:} Comparing this to the score of 5 assigned by another evaluator: The transcript reflects significant depressive symptoms, particularly in areas of interest, mood, energy, and self-worth, which seem to align more closely with a higher score than 5. The initial analysis of symptoms and their frequency suggests a higher PHQ-8 score.\\
 & \textbf{Conclusion:} The revised PHQ-8 score of 10 is consistent with the symptoms described in the transcript. This suggests moderate depressive symptoms. Therefore, the likelihood rating of 5 (moderate probability of an ongoing depressive disorder) is appropriate based on this re-evaluation. In conclusion, I \textbf{do not agree} with the assessment of a PHQ-8 score of 5 by another evaluator. The re-evaluated score of 10 aligns better with the individual's reported symptoms. This highlights the importance of thorough and consistent evaluation of the individual's experiences against established criteria.\\
\end{longtable}

\subsubsection{Custom GPT Evaluation}
Using the concatenated DAIC-WOZ dataset, we empirically evaluated Psycho Analyst's performance across different configurations of background knowledge, designed to simulate a variety of clinical knowledge settings. The configurations were as follows:
\begin{enumerate}
    \item A baseline scenario devoid of any background information, serving as a control.
    \item Incorporation of the PHQ-8, a widely-used depression screening tool consisting of eight questions that assess symptoms and severity of depression based on patient self-reporting. This configuration aims to evaluate the model's efficacy in applying standard depression screening metrics.
    \item Incorporation of the DSM-5, the definitive guide for psychiatric diagnoses. This integration includes comprehensive criteria and standardized language for mental disorders, intended to assess the model's alignment with widely recognized psychiatric criteria.
    \item A hybrid configuration with both PHQ-8 and DSM-5 to assess the synergistic effect of these diagnostic tools.
    \item An enhanced configuration that integrates DSM-5, PHQ-8, and a detailed data description from the dataset's curator along with an expanded training set, designed to optimize the model's contextual understanding and diagnostic precision.
\end{enumerate}
Additionally, Psycho Analyst's performance was compared with two baseline large language models: GPT-4o, a newly released model, and Mixtral-8*7B, a highly effective local model. Both models operated without any background knowledge, facilitating a direct assessment of Psycho Analyst's specialized enhancements. Mixtral-8*7B, running on an NVIDIA A100 GPU with 80GB of memory, provided a robust benchmark for computational efficiency and data processing capabilities. This comparison helped to contextualize the advancements of Psycho Analyst and its enhanced capability to interpret complex clinical data effectively. The effectiveness of each configuration was then assessed using the test set from the DAIC-WOZ dataset. The detailed result will be presented in Chapter~\ref{chap:evaluation}.


\section{A New Layered Multi-Expert Framework for Long-Context Mental Health Assessments \cite{tang2025layered}}
\label{sec:Section 4.4}

While the Custom GPT models discussed in Section 4.3 greatly advance the accuracy and scalability of mental health detection, they also face unique challenges in high-stakes mental health assessments. On the one hand, \textbf{mental health data are inherently subjective}: instruments like the PHQ-8 provide structured guidelines, yet the interpretation of symptoms can vary considerably from one individual or conversation to another \cite{ng2019provider,Marsolek20240969PD}. On the other hand, \textbf{long and complex transcripts}—such as multi-turn interviews or extended narrative case studies—can overwhelm even advanced models, resulting in hallucinations, inconsistencies, or overlooked details \cite{huang2023survey,dahl2024large}. Tools like the PHQ-8 only partially mitigate this issue; a single “expert” LLM may still miss subtle cues without additional context or repeated checks \cite{tangshang2024advancing}. Although iterative reassessments and third-party evaluations can improve performance, they often rely on highly structured prompts or assume a known best configuration.

To overcome these issues, we developed a new multi-expert framework that builds upon our GPT-based systems to more effectively handle extended, nuanced mental health evaluations. The next section introduces our new \textit{Stacked Multi-Model Reasoning (SMMR)} approach, wherein multiple model generate preliminary assessments and then collectively refine them to reduce hallucinations and inaccuracies. By incorporating diverse “expert” perspectives and systematically reconciling their outputs, SMMR offers a more robust and reliable long-context evaluation strategy, particularly for multi-turn or multi-session psychiatric interviews.

\subsection{SMMR Methodology}

\subsubsection{Stacked Multi-Model Reasoning (SMMR)}

\begin{figure}[!ht]
    \centering
    \includegraphics[width=0.7\textwidth]{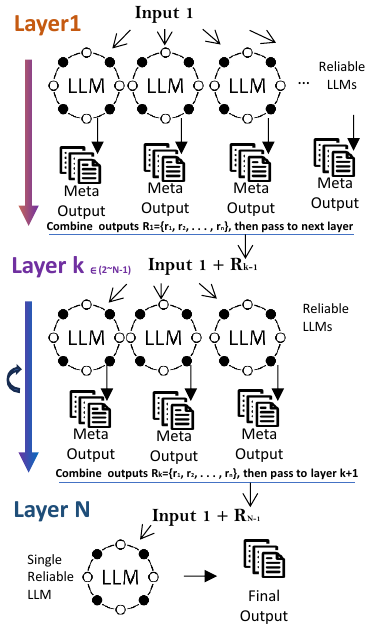} 
    \caption{Overview of the Stacked Multi-Model Reasoning (SMMR) framework. Each layer refines or integrates outputs from the previous layer to form the next layer’s input. No prior ranking of model quality is assumed.}
    \label{fig:SMMR_smmr_framework}
\end{figure}

\textbf{Motivation and Overview:}
Long-form mental health assessments often require careful synthesis of complex, multi-turn data. Even advanced Large Language Models (LLMs) can become prone to hallucinations or inconsistent reasoning in these extended contexts. To address this problem, we propose \textit{Stacked Multi-Model Reasoning (SMMR)}, a framework that treats each LLM (or smaller specialized model) as an independent ``expert,'' without \textit{a priori} knowledge of which model might be ``best.'' By layering multiple models---each offering a second opinion---SMMR leverages collective insights to mitigate the weaknesses of any single model. \autoref{fig:SMMR_smmr_framework} illustrates the conceptual architecture of SMMR, while Algorithm~\ref{alg:SMMR_SMMR} provides pseudocode for the overall process.

\begin{algorithm}[ht]
\caption{Stacked Multi-Model Reasoning (SMMR)}
\label{alg:SMMR_SMMR}
\SetAlgoLined
\KwIn{Long-context data $\mathcal{X}$; set of $L_1$-type Single-Step Models $\{M_1^1,\dots,M_1^n\}$; set of $L_k$-type Long-Context Models $\{M_k^1,\dots,M_k^m\}$ for each layer $k$; number of layers $N$.}
\KwOut{Final aggregated output $R_{\mathrm{final}}$.}

\textbf{Layer 1: Initial Opinions} \\
\ForEach{model $M_1^i \in \{\text{Single-Step Models}\}$}{
    $r_1^i \gets M_1^i(\mathcal{X})$\;
}
Aggregate outputs: $R_{1} \gets \{r_1^1, r_1^2,\dots,r_1^n\}$\;

\textbf{Layers 2 to $N-1$: Iterative Refinement} \\
\For{$k = 2$ to $N - 1$}{
  \ForEach{model $M_k^j \in \{\text{Long-Context Models at layer }k\}$}{
    $r_k^j \gets M_k^j(R_{k-1})$\; 
  }
  Aggregate outputs: $R_k \gets \{r_k^1, r_k^2,\dots,r_k^m\}$\;
}

\textbf{Layer N: Final Consolidation} \\
Use a single, reliable Long-Context Model $M_N^*$: \\
$R_{\mathrm{final}} \gets M_N^*(R_{N-1})$\;

\Return $R_{\mathrm{final}}$ 
\end{algorithm}

\paragraph{Layer 1: Multiple Independent Experts.} 
In this initial step, SMMR applies multiple \emph{single-step} models, each independently processing the input $\mathcal{X}$. These models could be smaller LLMs or specialized classifiers that excel at short-context tasks. Since we do not assume any model to be inherently superior, all outputs are aggregated on equal footing, forming $R_1$ for further refinement.

\paragraph{Layers 2 to $N-1$: Iterative Refinement with Long-Context Models.}
Subsequent layers introduce \emph{long-context} LLMs capable of handling extended or detailed inputs. Each model in layer $k$ takes the aggregated output $R_{k-1}$ from the previous layer and refines it, generating new outputs $r_k^j$. These outputs are then aggregated into $R_k$. The iterative process allows each layer to provide a ``second opinion,'' reconciling discrepancies and minimizing hallucinations or inconsistencies that could arise from any single model.

\paragraph{Layer N: Final Consolidation by a Reliable Model.}
In the last layer, a single, long-context \textit{Reliable Model} $M_N^*$ synthesizes the refined outputs from $R_{N-1}$ into the ultimate result $R_{\mathrm{final}}$. This top-tier model is selected for its stable performance on nuanced, long-context data—making it especially well-suited for high-stakes mental health evaluations.

\paragraph{Dynamic Stopping Based on Performance Optimization.}
To ensure optimal performance and computational efficiency, the SMMR framework incorporates a dynamic stopping mechanism. After each layer processes and refines the aggregated outputs from the preceding layer, the framework evaluates the current performance using predefined metrics (e.g., Accuracy, F1-score, MAE). If the performance metrics improve compared to the previous layer by a significant threshold $\delta$, SMMR proceeds to the next layer for further refinement. This iterative process continues until adding another layer does not result in performance gains beyond the threshold. The final output used for evaluation is thus the result from the layer that achieved the highest performance metrics, optimizing both accuracy and resource utilization.

\medskip

By dividing the reasoning process into distinct layers and integrating diverse model outputs at each stage, SMMR effectively mitigates the risk of hallucinations and maintains stronger consistency for complex mental health tasks.

\subsection{Datasets and Task Setup}
\label{subsec:datasets_and_tasks}

In order to demonstrate the effectiveness of our Stacked Multi-Model Reasoning (SMMR) framework, we evaluate on two complementary mental health datasets: (1) an externally sourced dataset (DAIC-WOZ) \cite{gratch2014distress}, and (2) a curated collection of narrative case studies. Our primary task involves predicting mental health risk and severity, operationalized through PHQ-8 scores or binary labels, based on extended transcripts or descriptive case data. By adopting a layered approach that draws on multiple ``expert'' models, SMMR aims to reduce hallucinations, enhance diagnostic fidelity, and improve the reliability of long-context LLM-based assessments.

\subsubsection{Datasets} 
\textbf{The DAIC-WOZ Database:} A private dataset consists of 187 labeled interviews designed to assess psychological distress such as anxiety, depression, and PTSD. Each interview includes a PHQ-8 score and its corresponding binary label. Following the standard protocol in \cite{gratch2014distress}, we split the dataset into training and testing subsets. To create a more realistic, long-context input for our models, each conversation was concatenated into a single data stream by aligning segments according to the speaker’s starting time. The consolidated dataset includes references to the speaker, the content of each segment, and punctuation marks consisting of a period followed by a slash (./) to denote the end of each speaking turn. We evaluate model outputs using:
\begin{itemize}[leftmargin=*]
    \item \textbf{PHQ-8 Score Estimation (0--24)}: We compare the predicted PHQ-8 score to the ground truth, measuring accuracy with metrics such as Mean Absolute Error (MAE) and Root Mean Squared Error (RMSE).
    \item \textbf{Binary Classification (PHQ-8 $\geq 10$)}: The model’s output is thresholded at 10 to indicate the presence or absence of clinically significant depressive symptoms.
\end{itemize}

\textbf{Case Study Dataset:} To complement the interview-style data from the DAIC-WOZ database, we collected 48 narrative case studies featuring professionals' concluded psychiatric diagnoses. These cases, sourced from academic texts and clinical literature, include detailed demographic, behavioral, and contextual information. We manually extracted binary conclusions and types of disorders from the diagnoses provided by clinical professionals to verify the presence or absence of mental health concerns. Of these case studies, six are formatted as conversational transcripts, while the remaining 42 are presented in a descriptive format. An example is shown in \autoref{tab:SMMR_case_study_example}.

Because these cases do not provide PHQ-8 labels, we adopt an alternative evaluation scheme:
\begin{itemize}[leftmargin=*]
    \item \textbf{Mental Concern (0, 1, 2)}: Determines whether the case indicates no mental health issue (0), presence of a mental health issue (1), or if the conclusion is indeterminate from the data (2).
    \item \textbf{Disorder-Type Identification}: Extracts the specific mental health disorder(s) (if any) from the text. We measure accuracy by comparing the model-identified disorders to a reference list of ground-truth labels, considering minor variations in naming as valid matches.
\end{itemize}

\begin{table}[htbp]
\centering
\caption{Case Study Example}
\label{tab:SMMR_case_study_example}
\begin{tabular}{@{}p{\columnwidth}@{}}
\hline
\textbf{Case:} A 37-year-old white male infantryman stationed in Iraq arrived at a field hospital complaining that his superior officer placed poisonous ants in his helmet. His face is covered with excoriations from persistent scratching. On further examination, he is stuporous and has mildly slurred speech, tremor, and mint odor to his breath. Later his troop leader mentioned that his Humvee was littered with empty bottles of mouthwash and that the man has been reprimanded for falling asleep at his post. After a night of rest, he discussed his excessive use of mouthwash in place of alcohol, which is the only available form of alcohol in Iraq. \\ \hline
\textbf{Conclusion:} The individual in the provided case study exhibits symptoms consistent with delusions of persecution (believing his superior officer placed poisonous ants in his helmet), excoriations from persistent scratching indicative of possible hallucinations or delusions, stupor, slurred speech, tremors, and the smell of mint on his breath. Additionally, his excessive use of mouthwash as a substitute for alcohol suggests a coping mechanism or self-medication behavior. While the text does not explicitly provide a diagnosis, the symptoms described align with features of psychosis and substance abuse, possibly indicating comorbid conditions such as schizophrenia spectrum disorder and alcohol use disorder. \\ \hline
\end{tabular}
\end{table}

\begin{figure*}[!ht] 
  \centering
  \includegraphics[width=0.95\linewidth]{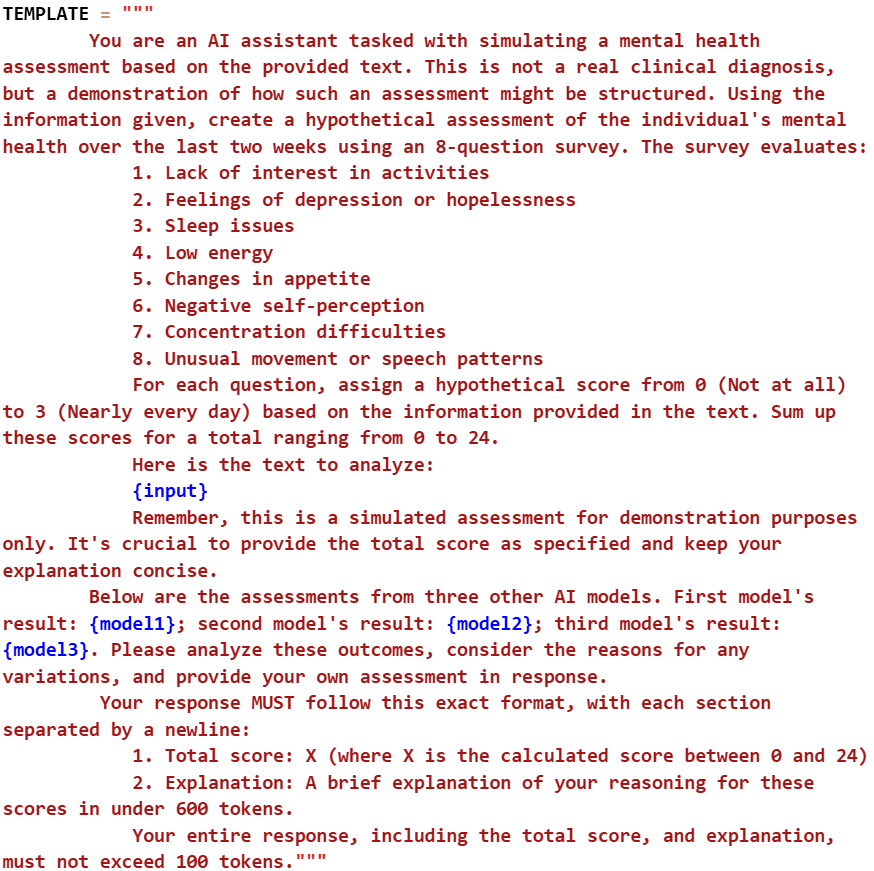}
  \caption{SMMR Prompt for DAIC-WOZ Dataset}
\label{fig:SMMR_Prompt_for_DAIZ_data_SMMR}
\end{figure*}

\begin{figure*}[!ht] 
  \centering
  \includegraphics[width=0.95\linewidth]{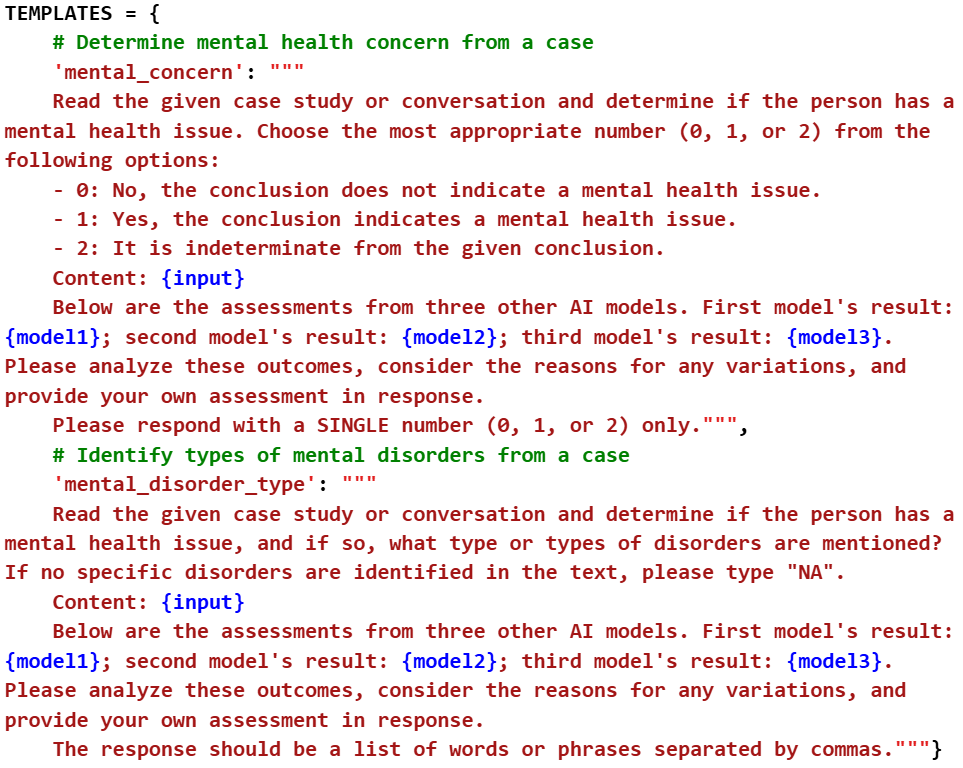}
  \caption{SMMR Prompt for Case Study Dataset}
\label{fig:SMMR_Figure_Case_Study_SMMR}
\end{figure*}

\subsection{Model Selection} 
We initially conducted a pilot test with the smaller, local Mistral model to explore on-premise feasibility. However, the majority of outputs were invalid or incomplete, likely due to context window constraints and the domain-specific nature of the data. Consequently, we omitted local models and go with GPT models but note that future work could revisit them if they become more robust.

\subsection{Evaluation Method for SMMR} 
For each dataset instance—whether an interview transcript from DAIC-WOZ or a narrative case study—SMMR processes the \emph{entire} text in a layered manner (the prompts shown in \autoref{fig:SMMR_Prompt_for_DAIZ_data_SMMR} and \autoref{fig:SMMR_Figure_Case_Study_SMMR}, respectively):
\begin{enumerate}[leftmargin=*]
    \item \textit{Layer 1} applies multiple smaller or specialized models in parallel to gather initial assessments.  
    \item \textit{Layers 2 through $N-1$} employ advanced long-context LLMs to refine, reconcile, and aggregate these preliminary outputs.  
    \item \textit{Layer $N$} employs a single, reliable long-context model to finalize the assessment and generate the ultimate decision, whether it be a binary classification or a PHQ-8 score.
    \item \textit{Dynamic Stopping}: The SMMR framework automatically terminates additional layering when further layers do not yield performance improvements beyond a predefined threshold.
\end{enumerate}

This pipeline effectively creates multiple “checkpoints” for error correction and multi-expert verification, ultimately aiming to improve the reliability of mental health evaluations in lengthy and complex conversations. The detailed result will be presented in Chapter~\ref{chap:evaluation}.


\chapter{Evaluation and Synthesis for Preventive and Intervention Tools}
\label{chap:evaluation}
\section{Evaluation of Preventive Tools}
\blfootnote{Portions of this chapter were previously published in \cite{tang2025tigergpt}, \cite{tang2024decoding}, \cite{tang2024advancing}, and \cite{tang2025layered}. © 2024--2025 IEEE. Reprinted, with permission, from J. Tang et al., \textit{2024 IEEE 6th International Conference on Cognitive Machine Intelligence (CogMI)}, Washington, DC, USA, 2024; \textit{2025 IEEE International Conference on Future Machine Learning and Data Science (FMLDS)}, Los Angeles, CA, USA, 2025; and \textit{2025 IEEE Conference on Artificial Intelligence (CAI)}, Santa Clara, CA, USA, 2025.}

In this section, we will present the evaluation and results for the new TigerGPT chatbot and the new AURA framework presented in Chapter \ref{chap:preventive}.
\subsection{TigerGPT Evaluation}
To evaluate the performance of TigerGPT, we recruited undergraduate students in a psychology class from the University of Missouri to test it. We collected data from two primary sources: (1) a traditional questionnaire-based feedback survey conducted immediately following each TigerGPT survey, and (2) written evaluation reports submitted by student evaluators.

\begin{figure}[!t]
    \centering
        \includegraphics[width=1.0\linewidth]{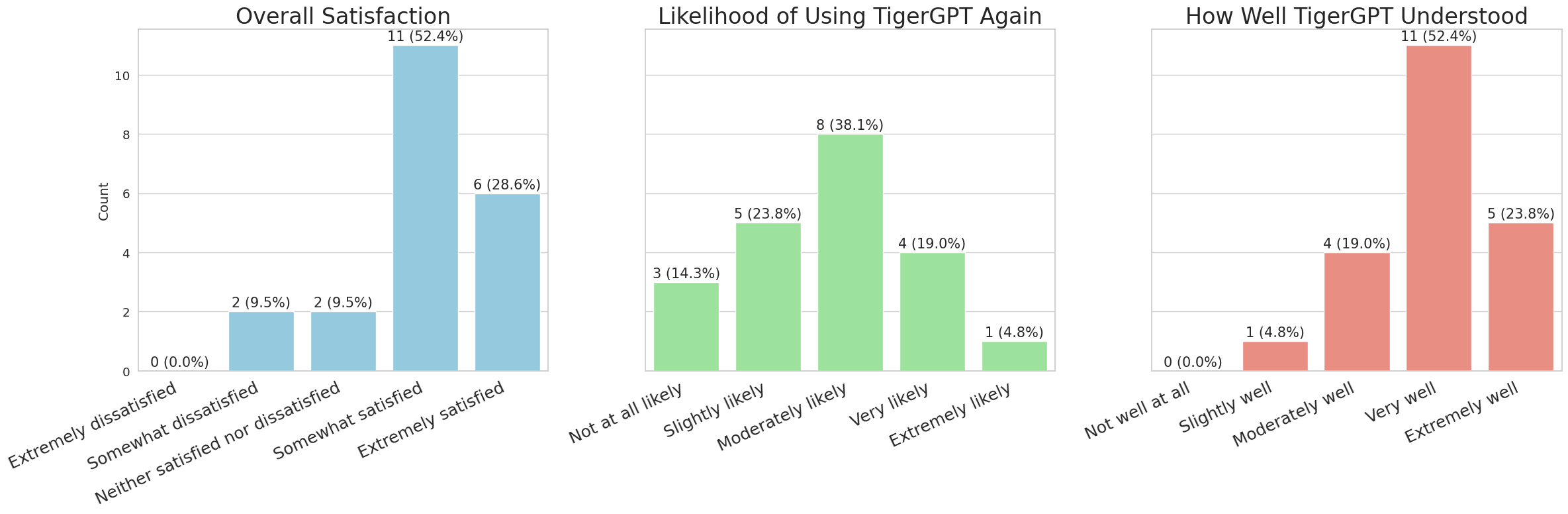}
    \caption{{Evaluation results of user opinions towards TigerGPT.} These three bar charts show participants’ overall satisfaction, likelihood of using TigerGPT again, and how well the system understood their queries.}
    \label{fig:opinion}
    \footnotesize
\end{figure}

\begin{figure}[!t]
    \centering
        \includegraphics[width=0.5\linewidth]{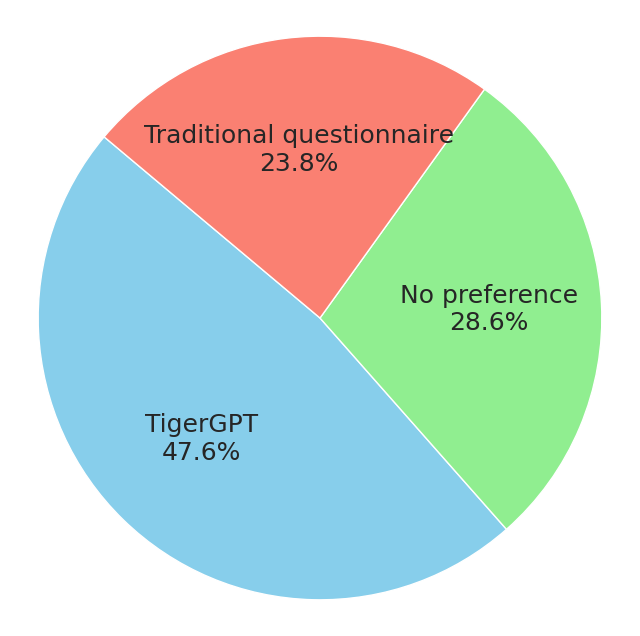}
    \caption{{Evaluation results of participant preference.} This pie chart shows how participants prefer TigerGPT, a traditional questionnaire, or neither.}
    \label{fig:preference}
\end{figure}

\subsubsection{{User Feedback Survey Results}}
A total of 21 participants completed the feedback survey, summarized in \autoref{fig:opinion}. Approximately 81\% reported being at least “somewhat satisfied,” with 28.6\% indicating they were “extremely satisfied,” and only 9.5\% expressing dissatisfaction. Nearly 62\% noted they were “moderately likely” or more inclined to use the system again, while a smaller proportion showed lower willingness. Regarding perceived comprehension, around 76\% felt their input was understood “very well” or “extremely well,” suggesting that most respondents found the tool’s responses suitably tailored.

As shown in \autoref{fig:preference}, about half of the participants preferred TigerGPT, roughly a quarter preferred traditional questionnaires, and the remainder had no strong preference. Overall, these findings indicate a generally positive reception, tempered by some uncertainty about whether this approach offers decisive advantages over traditional survey methods.

\begin{figure}[!t]
    \centering
        \includegraphics[width=0.9\linewidth]{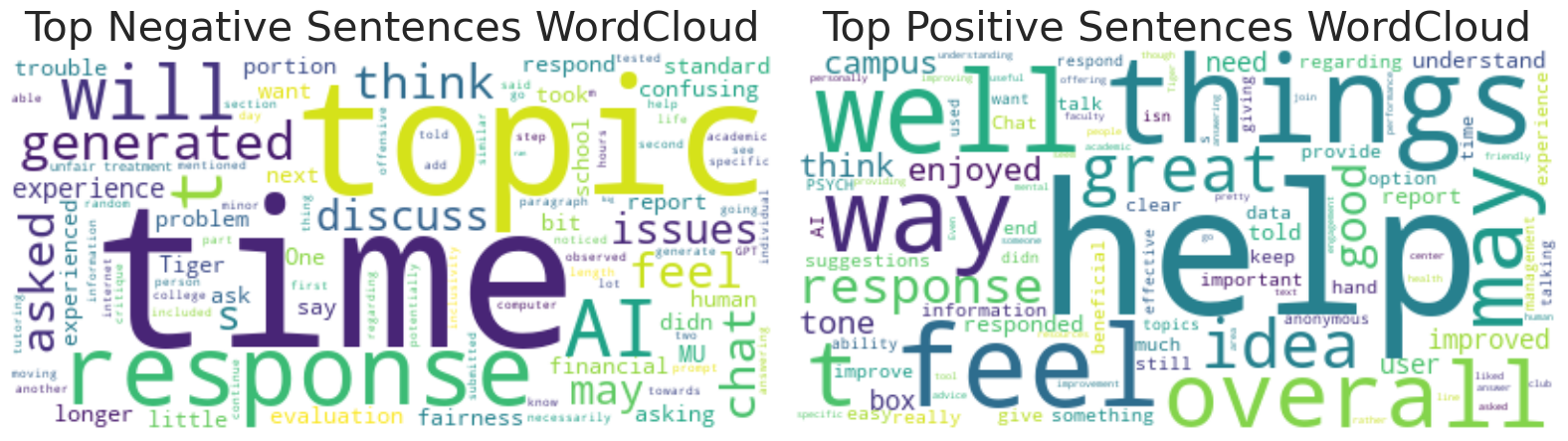}
    \caption{{Word cloud visualization of user feedback.} The left cloud highlights frequently mentioned words from negative feedback (e.g., time, topic, response), while the right cloud shows terms from positive feedback (e.g., help, overall, things). These visuals provide a quick overview of users’ main concerns and appreciation points.}
    \label{fig:Tiger_wordcloud}
    \footnotesize
\end{figure}

\begin{figure}[!t]
    \centering
        \includegraphics[width=1.0\linewidth]{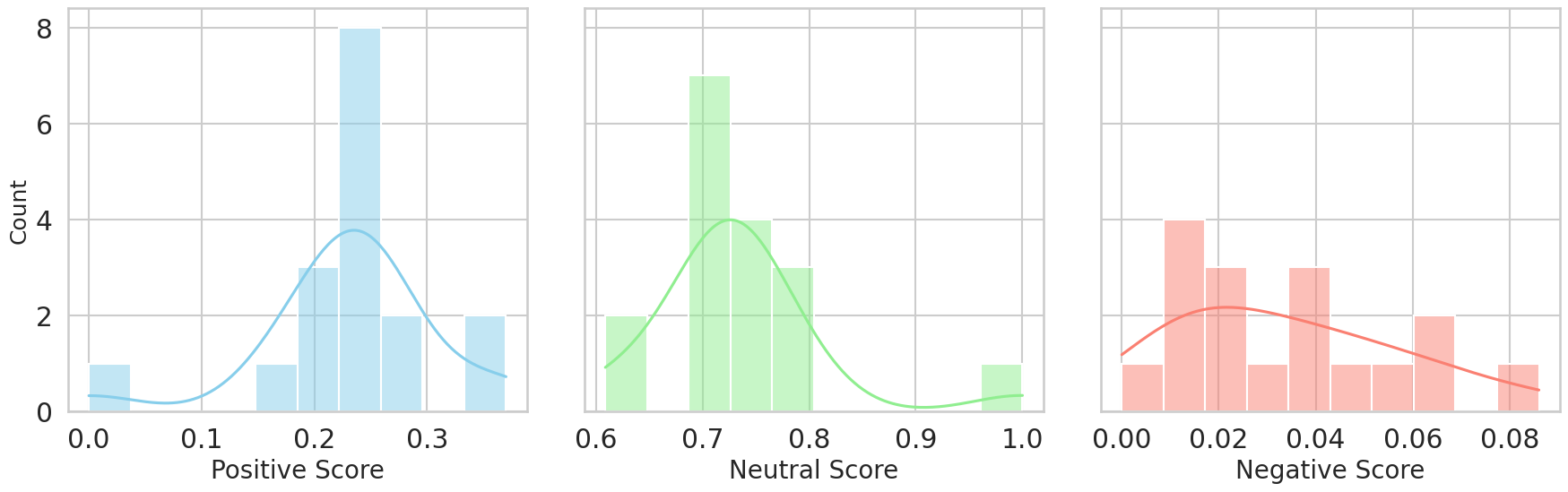}
    \caption{{Distributions of positive, neutral, and negative sentiment scores from participants.} This figure displays three histograms illustrating how frequently users’ feedback falls into varying degrees of positive, neutral, and negative sentiment. Each histogram shows the number of responses (y-axis) at different sentiment score ranges (x-axis), with a smoothing curve overlaid to highlight the general distribution trend.}
    \label{fig:Tiger_sentiment_distribution}
    \footnotesize
\end{figure}

\subsubsection{Evaluation Report Results}

A total of 17 students submitted written reports discussing both promising aspects and areas of concern. \autoref{fig:Tiger_wordcloud} shows word clouds derived from their most positive and negative statements. In the positive word cloud, terms such as \emph{help}, \emph{overall}, and \emph{things} appear frequently, reflecting an appreciation for TigerGPT’s assistance and a generally favorable user experience. Meanwhile, in the negative word cloud, words like \emph{time}, \emph{topic}, and \emph{response} stand out, suggesting frustration with occasional slow responses or limited topic coverage. This contrast illustrates the dual nature of user sentiment: while many enjoyed TigerGPT’s interactivity and friendly tone, some felt it could better handle broader or more flexible discussion prompts.

\autoref{fig:Tiger_sentiment_distribution} further illustrates how users’ written feedback distributes across \emph{positive}, \emph{neutral}, and \emph{negative} sentiment scores, based on VADER (Valence Aware Dictionary and sEntiment Reasoner). Each histogram shows the number of responses (y-axis) in different score intervals (x-axis), with a smoothing curve to highlight overall trends. Notably, positive scores cluster around 0.2–0.3, neutral scores center near 0.7, and negative scores remain low (generally below 0.08). VADER calculates these values by analyzing lexical cues and punctuation, then produces an overall polarity (the \emph{compound} score) alongside distinct positive, neutral, and negative metrics.

\autoref{tab:sentiment_stats} provides VADER descriptive statistics: the mean compound score stands at 0.93, confirming the predominantly positive tone seen in the word clouds. Although some comments addressed issues like slow response times or narrow topic options—reflected in minor negative scores—the overall distribution indicates that students generally found TigerGPT both helpful and engaging. Taken together, these results suggest that while most users had a favorable experience, improvements to performance and topic breadth could further enhance satisfaction.

\begin{table}[!t]
\centering
\begin{threeparttable}
\fontsize{12pt}{20pt}\selectfont
\setlength{\tabcolsep}{18pt}
\caption{VADER Sentiment Statistics of TigerGPT Evaluation Results.} 
\label{tab:sentiment_stats}
\begin{tabular}{lcccc}
\hline
 & {Compound} & {Positive} & {Neutral} & {Negative} \\
\hline
{Mean} & 0.93 & 0.23 & 0.74 & 0.03 \\
{SD}   & 0.24 & 0.08 & 0.08 & 0.02 \\
{Min}  & 0.00 & 0.00 & 0.61 & 0.00 \\
{Max}  & 1.00 & 0.37 & 1.00 & 0.09 \\
\hline
\multicolumn{5}{l}{{Number of reports:} 17}\\
\multicolumn{5}{l}{{Average Compound Score:} 0.93}\\
\hline
\end{tabular}
\smallskip
\begin{tablenotes}
\small
    \item Note. Descriptive statistics (mean, standard deviation, minimum, and maximum) for compound, positive, neutral, and negative sentiment scores across 17 user reports. The average compound score of 0.93 suggests an overall positive sentiment trend.
\end{tablenotes}
\end{threeparttable}
\end{table}

Building on these sentiment findings, many students appreciated the system’s {flexible, user-friendly design}. They highlighted features like its {cross-platform interface} supporting smartphones, tablets, and computers, and {“pick one for me” prompts}, which helped them explore new topics without confusion. The inclusion of {bolded key questions} was particularly beneficial for those with attention difficulties; one student mentioned it helped them stay focused during the conversation. Students also felt {safe and anonymous} when giving feedback, promoting more direct and detailed responses and making the experience {more engaging than a typical survey}.

Regarding content, participants praised the {wide range of relevant topics}, encompassing academic life, financial aid, work–life balance, and campus inclusivity. They also enjoyed the chatbot’s {human-like elements}; for instance, it addressed them by name and major, used an {enthusiastic tone} (often with emojis), and offered {friendly affirmations} that created a warm, conversational atmosphere. Subtle {university branding} (e.g., university icons) reinforced the campus feel without overwhelming the user. Lastly, several students saw {potential for deeper academic integration}, suggesting the chatbot could link to the university’s scheduling system or relevant resource pages, reducing the need for in-person visits and offering a convenient way to access campus support.

While participants generally appreciated the system, they also identified areas for potential improvement. For example, technical stability enhancements could address occasional slow or inconsistent response times, as well as the need to re-enter responses after a page reload. Diversifying language choices might reduce repetitive phrases (e.g., “I see” or “That’s amazing”) and enable more in-depth follow-ups. Some users also proposed broadening the range of topics, covering a larger and more diverse population, and allowing a more flexible conversation flow so predefined prompts would not limit discussions of pressing issues. Finally, several students wished for more detailed feedback from the system, suggesting that deeper insights would better capture their opinions and enrich the overall experience.

\subsection{Sample Summarization Report for TigerGPT V2}

As a demonstration of the automated reporting feature introduced in TigerGPT V2, below is a sample output automatically generated by “Grok 3.” To create this report, we compiled all user logs—108 entries collected between 10/24/24 and 02/07/25—into a single JSON file. Grok 3 then parsed these logs to identify conversation topics, categorize user roles, and infer key insights from the dialogues. This process requires minimal manual oversight, allowing campus administrators or survey coordinators to rapidly gauge recurring issues and shifting priorities in real time.

\subsubsection{AI-Generated Summary of Findings from 108 Conversational Surveys with TigerGPT}
\begin{mdframed}[
  linecolor=black,
  linewidth=1pt,
  backgroundcolor=gray!10,
  roundcorner=5pt,
  leftmargin=0pt,
  rightmargin=0pt,
  skipabove=\medskipamount,
  skipbelow=\medskipamount
]

\noindent
\textbf{Data Source and Overview}
\begin{itemize}
  \item \textbf{Dataset Size:} 108 conversational surveys, spanning 10/24/24 to 02/07/25.
  \item \textbf{Structure:} JSON-based logs from TigerGPT V2.
  \begin{itemize}
    \item \emph{Metadata Entries:} Short JSON records documenting user roles, statuses, or selected topics.
    \item \emph{Conversation Turns:} Longer JSON records containing fields like \texttt{user\_input}, \texttt{chatbot\_response}, and \texttt{conversation\_round}.
  \end{itemize}
\end{itemize}

\noindent
\textbf{Topic Distribution}
\begin{itemize}
  \item Academic/Academic Issues: \(\sim\)35 conversations (\(\sim\)32\%)
  \item Work-Life Balance: \(\sim\)28 conversations (\(\sim\)26\%)
  \item Financial Issues: \(\sim\)17 conversations (\(\sim\)16\%)
  \item Inclusivity and Fairness: \(\sim\)7 conversations (\(\sim\)7\%)
  \item Research Opportunities \& Contributions: \(\sim\)7 conversations (\(\sim\)7\%)
  \item Academic Advising \& Support: \(\sim\)7 conversations (\(\sim\)7\%)
  \item Other (Admissions, Compensation, Lab Resources): \(\sim\)7 conversations (\(\sim\)7\%)
\end{itemize}
\noindent
\emph{Key Observation:} Academic-related questions dominate, followed by work-life balance and financial concerns.

\bigskip

\noindent
\textbf{Role Distribution}
\begin{itemize}
  \item Students: \(\sim\)94 conversations (\(\sim\)87\%)
    \begin{itemize}
      \item Undergraduate: 2
      \item Graduate (Master’s/Doctoral): 25
      \item International: 14 (overlap possible with master’s or doctoral)
    \end{itemize}
  \item Faculty: \(\sim\)10 conversations (\(\sim\)9\%)
  \item Staff: \(\sim\)3 conversations (\(\sim\)3\%)
  \item Parents/Future Students: \(\sim\)1--2 total (\(\sim\)1--2\%)
\end{itemize}
\noindent
\emph{Key Observation:} Students, especially graduate students, formed the largest user group, indicating TigerGPT V2’s primary reach.

\bigskip

\noindent
\textbf{Detailed Topic Insights}

\begin{itemize}
  \item \textbf{Academic/Academic Issues (\(\sim\)35 Conversations)}
  \begin{itemize}
    \item \emph{Satisfaction:} Users praised meaningful coursework and supportive faculty: ``learning a lot, strong interdisciplinary research.''
    \item \emph{Challenges:} Some mentioned non-academic frustrations influencing their academic satisfaction (e.g., ``I hate Greek life,'' ``need more technologies in the curriculum'').
    \item \emph{Support \& Access:} Positive views on faculty mentorship, though certain students struggle to schedule time with advisors.
  \end{itemize}

  \item \textbf{Work-Life Balance (\(\sim\)28 Conversations)}
  \begin{itemize}
    \item \emph{Routines \& Stress:} Students and faculty share coping tactics. Some mention ``just reading and sleeping'' to unwind, while others lack formal stress-management resources.
    \item \emph{Doctoral Pressures:} Doctoral candidates highlight intense schedules and a need for better mental-health or time-management support.
  \end{itemize}

  \item \textbf{Financial Issues (\(\sim\)17 Conversations)}
  \begin{itemize}
    \item \emph{Aid \& Funding:} Users discuss reliance on loans, TA or RA positions, with difficulties finding sustainable financial support.
    \item \emph{Impact on Academics:} Ongoing financial stress can undermine academic focus and well-being.
  \end{itemize}

  \item \textbf{Inclusivity \& Fairness (\(\sim\)7 Conversations)}
  \begin{itemize}
    \item \emph{Equity Concerns:} Brief references to perceived discrimination or controversial policies.
    \item \emph{Need for Broader Data:} TigerGPT consistently asks follow-ups about unfair treatment, but many replies are vague, indicating a need for deeper data or user willingness.
  \end{itemize}

  \item \textbf{Research Opportunities \& Contributions (\(\sim\)7 Conversations)}
  \begin{itemize}
    \item \emph{Funding \& Collaboration:} Users cite limited funding or resource awareness. Some highlight successes like ``TigerAware collaboration,'' but access is uneven.
  \end{itemize}

  \item \textbf{Academic Advising \& Support (\(\sim\)7 Conversations)}
  \begin{itemize}
    \item \emph{Advisor Availability:} Mixed experiences---some rave about regular one-on-one meetings, others struggle to find any advisor.
    \item \emph{Resource Awareness:} Many express uncertainty about existing campus support programs.
  \end{itemize}
\end{itemize}

\noindent
\emph{Overall Finding:} While academic and financial topics dominate, smaller but noteworthy concerns persist around inclusivity, research, and advising.
\end{mdframed}


\subsection{AURA Results}
\label{sec:AURA_results}

\subsubsection{Overall Performance Comparison}
\label{sec:overall_result}

\begin{sidewaystable}[htbp]
\centering
\begin{threeparttable}
\caption{Overall Performance Comparison: RL Configurations vs. Prior Baseline.}
\label{tab:AURA_overall_results}

\begin{tabular}{lcccccc}
\toprule
\textbf{Condition} & \textbf{$\Delta Q$} & \textbf{vs. Baseline} & \textbf{$p$-value} & \textbf{Cohen's $d$} & \textbf{$\bar{Q}$} & \textbf{$Q_{\text{final}}$} \\
\hline
Prior Baseline & $-0.006 \pm 0.118$ & --- & --- & --- & $0.513 \pm 0.127$ & $0.465 \pm 0.137$ \\
Config 1 ($\epsilon=0.15$) & $+0.087 \pm 0.190$ & $+0.093$ & $0.078$ & $+0.573$ & $0.535 \pm 0.124$ & $0.538 \pm 0.132$ \\
Config 2 ($\epsilon=0.30$) & $+0.070 \pm 0.106$ & $+0.076$ & $0.044^*$ & $+0.660$ & $0.553 \pm 0.102$ & $0.582 \pm 0.099$ \\
Config 3 ($\epsilon: 0.40 \rightarrow 0.05$) & $+0.034 \pm 0.135$ & $+0.040$ & $0.345$ & $+0.302$ & $0.525 \pm 0.112$ & $0.517 \pm 0.110$ \\
\bottomrule
\end{tabular}

\begin{tablenotes}[flushleft]
\normalsize
\item \textit{Note.} $n=20$ conversations per condition. $\Delta Q$ = quality improvement from Exchange~1 to Exchange~15; $\bar{Q}$ = average quality across all conversation exchanges; $Q_{\text{final}}$ = quality at Exchange~15. 
Statistical tests (columns 3 to 5) compare $\Delta Q$ between each RL configuration and the prior baseline using independent-samples $t$-tests. 
$^*p < 0.05$. Conventional benchmarks: medium effect $d=0.50$–$0.79$, large effect $d \geq 0.80$.
\end{tablenotes}

\end{threeparttable}
\end{sidewaystable}

Table~\ref{tab:AURA_overall_results} presents primary results comparing three RL configurations to the prior baseline. Configuration~2 ($\epsilon = 0.30$) significantly outperformed the baseline on quality improvement ($t(38) = 2.088$, $p = 0.044$, $d = 0.660$), demonstrating that moderate fixed exploration enables effective within-session adaptation. Configuration~2 also achieved substantially higher final quality scores ($Q_{\text{final}} = 0.582$ vs. baseline $0.465$, a difference of $+0.117$). Configuration~1 ($\epsilon = 0.15$) showed marginal improvement ($p = 0.078$, $d = 0.573$), while Configuration~3 (decaying epsilon) failed to reach significance ($p = 0.345$, $d = 0.302$).

The results provide clear answers to our research questions. For \textbf{RQ1}, RL-based adaptation yields significant quality improvements over non-adaptive baseline behavior when properly configured. The best-performing system (Configuration~2) achieved a quality improvement advantage of $+0.076$ relative to baseline (effect size $d = 0.660$, medium effect), and final quality exceeded baseline by $+0.117$. These effects demonstrate that within-session learning successfully addresses the engagement problems documented in the prior data.

For \textbf{RQ2}, moderate fixed exploration ($\epsilon = 0.30$) proved optimal for brief survey conversations. Configuration~1's conservative exploration ($\epsilon = 0.15$) approached but did not achieve significance, suggesting insufficient sampling leads to premature convergence (examined further in Section~\ref{sec:phase_analysis}). Configuration~3's decaying schedule ($\epsilon: 0.40 \rightarrow 0.05$) produced the weakest performance, indicating that high early exploration introduces excessive variability that disrupts learning before the system can benefit from increased exploitation.

Notably, the prior baseline exhibited minimal quality change ($\Delta Q = -0.006$), indicating that the non-adaptive fixed-probability strategy produced no measurable improvement in engagement over 15 exchanges. In contrast, all three RL configurations achieved positive quality improvement, with Configuration~2 achieving a final quality advantage of $+0.117$ over baseline. This demonstrates that adaptive learning not only improves over initial engagement levels but also maintains higher engagement throughout the conversation compared to static questioning strategies.

\subsubsection{Temporal Adaptation Dynamics}
\label{sec:phase_analysis}

Section~\ref{sec:overall_result} established that Configuration~2 ($\epsilon = 0.30$) significantly outperformed the prior baseline on overall quality improvement ($p = 0.044$, $d = 0.660$), while Configuration~1 showed only marginal effects and Configuration~3 failed to reach significance. To understand the mechanisms underlying these results---specifically, how within-session learning unfolds and why moderate exploration proves superior---we analyzed quality improvement in three temporal phases: early (Exchanges~1--5), mid (Exchanges~6--10), and late (Exchanges~11--15). This decomposition reveals critical insights into the progression of RL adaptation and identifies the failure modes that distinguish successful from unsuccessful exploration strategies. Table~\ref{tab:AURA_phase_results} presents phase-specific comparisons for all conditions.

\begin{sidewaystable}[htbp]
\centering
\begin{threeparttable}
\caption{Phase-Specific Quality Improvement by Condition.}
\label{tab:AURA_phase_results}

\begin{tabular}{lccccc}
\toprule
\textbf{Phase} & \textbf{Prior} & \textbf{Config 1} & \textbf{Config 2} & \textbf{Config 3} & \textbf{Significant} \\
 & \textbf{Baseline} & ($\epsilon=0.15$) & ($\epsilon=0.30$) & ($\epsilon: 0.40{\rightarrow}0.05$) & \textbf{Differences} \\
\hline
Early (1--5) & $+0.058 \pm 0.169$ & $+0.069 \pm 0.174$ & $+0.053 \pm 0.101$ & $+0.066 \pm 0.142$ & n.s. \\
Mid (6--10) & $+0.021 \pm 0.128$ & $+0.004 \pm 0.193$ & $-0.016 \pm 0.117$ & $+0.041 \pm 0.157$ & n.s. \\
Late (11--15) & $+0.010 \pm 0.150$ & $-0.116 \pm 0.143$ & $+0.043 \pm 0.131$ & $-0.016 \pm 0.151$ & $p=0.025^*$ \\
\bottomrule
\end{tabular}

\begin{tablenotes}[flushleft]
\normalsize
\item \textit{Note.} Values represent mean quality change within each conversation phase, computed as the last-exchange quality minus the first-exchange quality within that phase, averaged across all conversations ($n=20$ per condition). Statistical tests compare each RL configuration with the prior baseline using independent-samples $t$-tests. Only Configuration~1 in the late phase achieved statistical significance. n.s. = not significant at $\alpha=0.05$.
\item Significance detail: $^*p<0.05$ for Config~1 vs. Prior in the late phase ($t(38)=-2.357$, $d=-0.862$), indicating significant quality degradation.
\end{tablenotes}

\end{threeparttable}
\end{sidewaystable}

\textbf{Early phase} (Exchanges~1--5) showed no significant differences in within-phase quality improvement across conditions. All systems, including the prior baseline, achieved modest positive quality changes ($\Delta Q \approx +0.06$), suggesting that initial engagement benefits from novelty effects and general conversational scaffolding regardless of adaptive strategy. At this stage, RL systems have accumulated insufficient exchange data to differentiate effective policies through learning, relying primarily on population-level priors from the prior initialization (Section~\ref{sec:offline}). However, it is important to note that Configuration~2 achieved significantly higher absolute quality levels during this early phase (mean quality: 0.542 vs. baseline 0.493, $p = 0.006$, $d = 0.392$), indicating that moderate exploration enabled more effective initial question selection even before substantial within-session learning occurred. This early advantage provided a foundation for cumulative quality gains throughout the conversation.

\textbf{Mid phase} (Exchanges~6--10) exhibited high variability with no statistically significant performance differentiation across conditions. Quality improvements were near zero across all systems, with Configuration~2 showing slight within-phase degradation ($\Delta Q = -0.016$), Configuration~1 near zero ($\Delta Q = +0.004$), and Configuration~3 showing modest gains ($\Delta Q = +0.041$). This phase represents a critical transition period where learning mechanisms begin discovering engagement patterns but have not yet converged to stable policies. The apparent decline in Configuration~2 reflects normal fluctuation in quality trajectories rather than systematic failure---crucially, Configuration~2 maintained the highest absolute quality levels throughout this phase (mean quality: 0.559 vs. baseline 0.519, $p = 0.013$, $d = 0.378$), demonstrating sustained effectiveness despite temporary within-phase variance. Notably, Configuration~1 displayed the highest variability ($\pm 0.193$), suggesting unstable policy updates, while Configuration~2 maintained the most consistent performance ($\pm 0.117$), indicating more reliable learning dynamics.

\textbf{Late phase} (Exchanges~11--15) revealed critical differences in adaptation sustainability and exposed a fundamental failure mode in conservative exploration strategies. Configuration~1 ($\epsilon = 0.15$) exhibited significant \emph{negative} quality change ($\Delta Q = -0.116$) compared to prior baseline ($t(38) = -2.357$, $p = 0.025$, $d = -0.862$), indicating that conservative exploration led to premature convergence to a suboptimal policy that actively degraded engagement in later exchanges. This late-phase collapse represents a critical failure of insufficient exploration: the system locked onto learned patterns too quickly, preventing discovery of more effective strategies as conversation context evolved. Paradoxically, despite this degradation, Configuration~1 achieved the highest absolute quality in this phase (0.562) due to strong cumulative gains from earlier exchanges---highlighting that the degradation represents a \emph{failure to sustain adaptation} rather than overall poor performance. The system performed well early but could not maintain its trajectory.

In contrast, Configuration~2 ($\epsilon = 0.30$) maintained positive quality improvement ($\Delta Q = +0.043$) throughout the late phase, outperforming both baseline and Configuration~1 in trajectory sustainability. While this late-phase improvement did not achieve statistical significance ($p = 0.544$), it demonstrates consistent positive adaptation without the collapse observed in Configuration~1. This finding explains why Configuration~2 achieved overall significance in Section~\ref{sec:overall_result}: sustained late-phase performance and stable adaptation throughout the conversation drove the cumulative quality gains that produced the significant overall effect ($\Delta Q = +0.076$ relative to baseline, $p = 0.044$). Configuration~3 ($\epsilon: 0.40 \rightarrow 0.05$) showed slight negative change ($\Delta Q = -0.016$), suggesting that excessive early exploration disrupted coherent policy formation, and subsequent low exploration prevented recovery.

The phase-specific analysis demonstrates that within-session learning in brief conversational contexts requires sufficient exploration to avoid premature policy convergence while maintaining enough exploitation to capitalize on discovered patterns. Configuration~1's conservative 15\% exploration rate proved inadequate for the data-sparse environment (17 of 25 state-action pairs with prior examples, Table~\ref{tab:AURA_ev_table} in Section~\ref{sec:offline}), causing the system to lock onto suboptimal strategies that failed in later exchanges. With only 2--3 exploration trials across 15 exchanges, the system had insufficient sampling diversity to discover effective adaptations for evolving conversation states. Configuration~2's 30\% exploration maintained enough sampling diversity (4--5 exploration trials) to discover effective adaptations while still exploiting learned patterns, exhibiting the most consistent performance across all phases (standard deviations: 0.101--0.131 compared to Configuration~1's 0.143--0.193). This balanced approach enabled the system to avoid both premature convergence (Configuration~1's failure mode) and excessive randomness (Configuration~3's disruption from 40\% initial exploration).

Configuration~3's decaying schedule began with excessive exploration (40\%), introducing noise that disrupted early learning and prevented coherent policy development. By the time exploration decreased to facilitate exploitation, the system had accumulated noisy value estimates that hindered effective decision-making. This finding challenges the common assumption in RL that exploration should decay over time---in brief, high-stakes interactions where every exchange matters, moderate fixed exploration outperforms adaptive schedules. The fixed moderate rate allows the system to continuously validate learned patterns while remaining open to discovering better strategies as conversation context shifts, a critical capability for adaptive survey dialogues where user engagement states evolve unpredictably within single sessions.

These temporal dynamics reveal a fundamental design principle for within-session RL in conversational systems: moderate fixed exploration ($\epsilon = 0.30$) outperforms both conservative ($\epsilon = 0.15$) and decay-based schedules ($\epsilon: 0.40 \rightarrow 0.05$) in brief interactions. The failure modes differ meaningfully: insufficient exploration causes premature convergence and late-phase collapse, while excessive early exploration disrupts coherent policy formation. For survey applications and other brief goal-oriented dialogues, maintaining steady moderate exploration throughout the interaction proves more effective than traditional exploration-decay strategies optimized for long-horizon reinforcement learning tasks. These findings provide clear mechanistic support for the conclusion in Section~\ref{sec:overall_result} that Configuration~2 optimally addresses RQ2, establishing $\epsilon = 0.30$ as the preferred exploration rate for within-session conversational adaptation.

\subsubsection{Action Distribution Analysis}
\label{sec:action_analysis}

A key question for adaptive conversational systems is whether learning produces behaviorally meaningful changes in questioning strategies. Table~\ref{tab:AURA_action_dist} presents action distributions across all four conditions, revealing the specific behavioral adaptations that underlie the quality improvements demonstrated in Sections~\ref{sec:AURA_results}.

\begin{sidewaystable}[htbp]
\centering
\begin{threeparttable}
\caption{Action Type Distribution across Conditions.}
\label{tab:AURA_action_dist}

\begin{tabular}{lccccc}
\toprule
\textbf{Action Type} & \textbf{Prior} & \textbf{Config 1} & \textbf{Config 2} & \textbf{Config 3} & \textbf{Change (C2)} \\
 & \textbf{Baseline} & ($\epsilon=0.15$) & ($\epsilon=0.30$) & ($\epsilon: 0.40{\rightarrow}0.05$) & \\
\midrule
Specification & 58.5\% & 25.4\% & 22.2\% & 23.6\% & $-36.3$pp \\
Elaboration   & 27.4\% & 53.3\% & 48.1\% & 51.4\% & $+20.7$pp \\
Topic Probe   & 12.9\% & 8.2\%  & 13.0\% & 14.9\% & $+0.1$pp  \\
Validation    & 0.8\%  & 2.9\%  & 7.9\%  & 3.8\%  & $+7.1$pp  \\
Continuation  & 0.4\%  & 10.2\% & 8.8\%  & 6.2\%  & $+8.4$pp  \\
\bottomrule
\end{tabular}

\begin{tablenotes}[flushleft]
\normalsize
\item \textit{Note.} pp = percentage point. Values represent percentage of total actions selected during 15-exchange conversations ($n=20$ conversations per condition). Prior baseline percentages closely match empirical frequencies from original TigerGPT data (Table~\ref{tab:AURA_action-distribution}: specification 62.3\%, elaboration 23.6\%, topic probe 12.8\%, validation 0.9\%, continuation 0.4\%), validating baseline implementation within expected sampling variability.
\end{tablenotes}
\end{threeparttable}
\end{sidewaystable}

The prior baseline accurately reproduced TigerGPT's empirical distribution (specification 58.5\% vs. 62.3\% in Table~\ref{tab:AURA_action-distribution}, within sampling variability), validating the baseline implementation. All three RL configurations produced dramatic shifts in action selection, indicating that the system learned qualitatively different questioning strategies rather than minor parameter adjustments:

\textbf{Specification reduction.} The most striking behavioral change was a 62\% reduction in specification usage (58.5\% $\rightarrow$ 22.2\% for Configuration~2), demonstrating that the RL system learned to avoid over-reliance on requesting concrete examples---the primary engagement problem identified in prior data (Section \ref{sec:histor_action_dist}). The consistent reduction across all three RL configurations (57--62\% relative decrease) suggests this shift reflects discovered effectiveness patterns rather than random exploration artifacts. This dramatic rebalancing addresses the prior specification over-reliance (62.3\% in Table~\ref{tab:AURA_action-distribution}) that risked respondent fatigue through repetitive questioning.

\textbf{Validation increase.} Configuration~2 increased validation usage nearly 10-fold (0.8\% $\rightarrow$ 7.9\%), directly addressing the prior under-utilization problem documented in Section~\ref{sec:histor_action_dist}. This finding is particularly notable because validation had minimal prior examples ($n = 4$ total instances, Table~\ref{tab:AURA_action-distribution}), creating a cold-start problem where initial EV estimates were highly uncertain (EV = 0.348 with only 4 observations in the low\_stable state, Table~\ref{tab:AURA_ev_table}). The system's discovery that validation improves engagement---despite sparse prior data and requiring exploration-driven sampling---demonstrates effective learning from limited examples. Configuration~2's higher validation rate (7.9\% vs. 2.9--3.8\% for other RL systems) likely contributes to its superior quality performance, as validation provides rapport-building acknowledgment without demanding additional user effort.

\textbf{Elaboration shift.} All RL configurations substantially increased elaboration usage (27.4\% $\rightarrow$ 48--53\%), suggesting the system discovered that asking users to expand on previous responses yields higher quality than repeatedly requesting new specific examples. This shift aligns with conversational scaffolding principles where building on established topics maintains coherence and reduces cognitive load. However, the magnitude of this shift (approximately doubling elaboration frequency) may indicate over-correction, where the system over-exploits elaboration at the expense of other potentially effective actions. The similar elaboration rates across all three RL configurations (48--53\%) suggest this pattern emerges from prior EV estimates (Table~\ref{tab:AURA_ev_table}) rather than within-session learning, as all systems initialized with the same priors favoring elaboration in multiple states.

\textbf{Continuation adoption.} RL systems increased continuation usage from near-zero (0.4\%) to moderate levels (6--10\%), indicating the system learned to employ minimal follow-up prompts when users demonstrate sustained engagement. This represents discovery of a questioning strategy with virtually no prior precedent ($n = 2$ prior instances, Table~\ref{tab:AURA_action-distribution}), requiring pure exploration-driven learning. Configuration~1 showed the highest continuation usage (10.2\%), possibly compensating for its insufficient exploration in other action dimensions---the system may have defaulted to continuation when uncertain about more strategic alternatives. Configuration~2's more moderate continuation rate (8.8\%) suggests better-calibrated action selection.

\textbf{Topic probe stability.} Configuration~2 maintained topic probe frequency near prior levels (13.0\% vs. 12.9\% baseline), while Configuration~1 decreased topic probe usage (8.2\%). This stability in Configuration~2 demonstrates balanced exploration across action types rather than over-specialization. Topic probe serves a distinct function---introducing new conversation dimensions---that complements rather than competes with elaboration and specification. Configuration~2's preservation of topic probe usage while dramatically shifting other actions suggests more effective policy learning.

Configuration~2's action distribution most effectively balanced these shifts: it achieved the highest validation usage (7.9\%), lowest specification usage (22.2\%), maintained topic probe frequency near prior levels (13.0\% vs. 12.9\%), and showed moderate continuation and elaboration rates. This balanced portfolio of actions likely contributes to Configuration~2's superior overall performance (Section~\ref{sec:overall_result}), avoiding both the specification trap of the prior baseline and potential over-specialization on elaboration observed in Configuration~1 (53.3\%). The distribution demonstrates that effective within-session learning requires not just discovering individual effective actions, but learning when to deploy each action appropriately---a capability that Configuration~2's 30\% exploration rate enabled through sufficient sampling diversity across conversation contexts.

\subsubsection{Summary and Interpretation}
\label{sec:result_summary}

The evaluation results provide strong evidence that within-session reinforcement learning improves conversational survey quality compared to non-adaptive baseline behavior. Configuration~2 ($\epsilon = 0.30$) achieved statistically significant improvements in both overall quality change ($p = 0.044$, $d = 0.660$) and final engagement levels ($p = 0.005$, $d = 0.949$), validating AURA's core adaptive learning mechanism.

Three key findings emerge from the analysis:

\textbf{1. Exploration rate critically affects adaptation sustainability.} While all RL configurations learned to shift action distributions away from problematic prior patterns, only moderate exploration ($\epsilon = 0.30$) maintained performance gains across all conversation phases. Conservative exploration ($\epsilon = 0.15$) led to late-phase degradation through premature convergence, while decaying exploration schedules introduced excessive early variability that prevented effective learning.

\textbf{2. Behavioral changes reflect discovered effectiveness patterns.} The dramatic reduction in specification usage (63\%), substantial increase in validation (10-fold), and adoption of previously rare continuation actions demonstrate that the system discovered questioning strategies more effective than prior practice. These shifts cannot be attributed to random exploration given their consistency across RL configurations and their alignment with communication theory principles (validation supports rapport, excessive specification causes fatigue).

\textbf{3. Within-session learning succeeds despite sparse prior data.} AURA's policy initialization relied on only 96 prior conversations with substantial action imbalance (17 of 25 state-action pairs had examples). Despite these limitations, the system achieved significant quality improvements within 15-exchange interactions, suggesting that moderate exploration combined with rapid online learning ($\alpha = 0.30$) enables effective adaptation from limited priors.

These results establish Configuration~2 ($\epsilon = 0.30$, $\alpha = 0.30$) as the recommended parameter setting for deploying AURA in conversational survey applications. The system addresses documented engagement problems from prior practice while maintaining computational feasibility and privacy preservation through session-specific learning.

\section{Synthesis of Preventive Tools}

In Chapter~\ref{chap:preventive}, we presented two conversational survey tools for campus satisfaction: \textbf{TigerGPT} (see \autoref{sec:TigerGPTV1}) and \textbf{AURA} (see \autoref{sec:AURA}). Here we recap the key findings and synthesize their roles in the preventive layer.

\subsection{Recap of Key Findings}

\paragraph{TigerGPT.}
The pilot demonstrated that a domain-aligned, context-aware chatbot elicits nuanced campus-climate reflections and maintains a supportive tone with perceived personalization. Participants generally rated the experience positively and described it as more engaging and natural than form-based surveys, although template-driven follow-ups sometimes became repetitive after brief or low-emotion responses.

\paragraph{AURA.}
AURA achieved sustained within-session adaptation when using moderate exploration: $\epsilon=0.30$ preserved performance gains across early, mid, and late phases, whereas conservative exploration ($\epsilon=0.15$) converged too quickly and degraded late-phase quality, and decayed schedules ($\epsilon: 0.40{\rightarrow}0.05$) introduced early variability that impeded learning. Learned behavior shifted in theory-consistent ways, including a 63\% reduction in specification prompts, a ten-fold increase in validation, and increased use of continuation actions. Despite sparse priors (96 conversations, many state–action pairs unobserved), rapid online updates ($\alpha=0.30$) produced significant quality gains within sessions. Configuration~2 ($\epsilon=0.30$, $\alpha=0.30$) is recommended for deployment.

\subsection{Synthesis}
TigerGPT establishes a safe, reliable baseline for conversational sensing with strong user acceptance but limited adaptivity. AURA adds within-session learning that adjusts question types in real time based on observed LSDE quality, reducing repetition and increasing response quality. Together they form a layered preventive capability: TigerGPT for stable, domain-grounded elicitation and AURA for within-session optimization of response quality.

\subsection{Why this Matters for Campus Well-Being}
These tools lower barriers to sharing and surface actionable concerns earlier than periodic forms. TigerGPT provides consistent pulse-taking at scale across students, faculty, and staff, while AURA elicits specific, self-disclosing details that improve signal quality for triage. The resulting narratives help campus service units (student affairs, counseling services, human resources, and DEI) prioritize outreach and tailor resources, enabling faster, data-informed adjustments to services and communication in support of campus well-being.

\subsection{Limitations and Takeaway}
AURA’s exploration can be locally suboptimal in early exchanges and LSDE is a proxy that may reflect dataset biases. Even so, the evidence indicates complementary value: TigerGPT supplies safe and consistent elicitation, and AURA supplies adaptive control that lifts response quality in real time. See Chapter~\ref{chap:discussion} for detailed limitations.

\section{Evaluation of Intervention Tools}
\label{sec:intervention_result}
In this section, we present the results of the three new intervention tools—Decoding Linguistic Nuances, Psycho Analyst, and SMMR—each targeting mental health evaluations at a distinct level of complexity, from rapid detection to multi-expert, long-context analysis.

\subsection{Decoding Linguistic Nuances Results}

\subsubsection{Phase 1 Fine-Tuning and Comparing Models Results}

\begin{table*}[!t]
\centering
\begin{threeparttable}
\fontsize{11.5pt}{16pt}\selectfont
\caption{Comparative Accuracy of Various Models Across Different Depression Testing Sets}
\label{table:decoding4}
\begin{tabular}{lccc}
    \toprule
    Model & \multicolumn{3}{c}{Accuracy} \\ 
    \cmidrule(lr){2-4}
          & GNS-Dep. Test Set & ENS-Dep. Test Set & Mix-Dep. Test Set \\
    \midrule
    BERT(10) & 0.86 & 0.83 & 0.82 \\
    BERT(64) & \textbf{0.99} & 0.94 & 0.95 \\
    BERT(128) & \textbf{0.99} & \textbf{0.96} & \textbf{0.96} \\
    BERT(300) & \textbf{0.99} & \textbf{0.96} & 0.94 \\
    MentalBERT(10) & 0.86 & 0.80 & 0.81 \\
    MentalBERT(64) & 0.98 & 0.92 & 0.94 \\
    MentalBERT(128) & \textbf{0.99} & \textbf{0.96} & \textbf{0.96} \\
    MentalBERT(300) & \textbf{0.99} & 0.95 & 0.95 \\
    NB(TDIDF: 5000) & \textbf{0.99} & 0.94 & 0.90 \\
    SVM(TDIDF: 5000) & \textbf{0.99} & 0.95 & 0.93 \\
    LR(TDIDF: 5000) & 0.97 & 0.94 & 0.90 \\
    NB(unigram) & \textbf{0.99} & 0.93 & 0.93 \\
    SVM(unigram) & 0.97 & 0.93 & 0.92 \\
    \bottomrule
   \end{tabular}
    \begin{tablenotes}
\item * Dep. = depression
\item * Bold numbers indicate the highest accuracy for each test set.
\item * Parameters in parentheses are model configurations: 'BERT' and 'MentalBERT' numbers specify input sequence lengths. 'TDIDF: 5000' and 'Unigram' denote the type of text processing used.
\end{tablenotes}
\end{threeparttable}
\end{table*}

\autoref{table:decoding4} presents how BERT and MentalBERT perform at different token lengths, alongside simpler methods like SVM and Naïve Bayes, for classifying Depression-related Expressive Narrative Stories (ENS) versus General Narrative Stories (GNS).

Overall, both transformer-based models achieved high accuracy and F1 scores. Performance on ENS-Depression and Mix-Depression datasets peaked at 128 tokens (0.96 accuracy/F1), while GNS-Depression reached 0.99 accuracy/F1 at 64 tokens. Increasing context beyond these lengths did not yield further gains.

Interestingly, classical classifiers rivaled deep learning on GNS-Depression but were less effective on ENS tasks. This underscores the value of transformer architectures for capturing the nuanced language found in expressive narratives. Taken together, these results suggest that BERT(128) and MentalBERT(128) offer an optimal balance between context coverage and classification accuracy in mental health text analysis.
\begin{sidewaystable*}[htbp]  
\centering
\begin{threeparttable}
\fontsize{7.5pt}{12pt}\selectfont
\caption{Performance Change and T-test Result of Word Manipulations across Different Models and Different Testing Sets.}
\label{table:decoding6}
\setlength\tabcolsep{1pt} 
\begin{tabular}{llcccccccccccc}
\toprule
\multicolumn{1}{l}{\multirow{2}{*}{Model}} & \multicolumn{1}{l}{\multirow{2}{*}{Word Manipulation}} & \multicolumn{4}{c}{GNS-Depression Testing Set} & \multicolumn{4}{c}{ENS-Depression Testing Set} & \multicolumn{4}{c}{Mix-Depression Testing Set} \\ 
\cmidrule(lr){3-6} \cmidrule(lr){7-10} \cmidrule(lr){11-14}
& & Accuracy(\%) & AccDiff(\%) & t-statistics & P-value & Accuracy(\%) & AccDiff(\%) & t-statistics & P-value & Accuracy(\%) & AccDiff(\%) & t-statistics & P-value \\
\midrule
\multirow{3}{*}{BERT(128)} & Raw & 99.3 & - & - & - & 95.9 & - & - & - & 95.9 & - & - & - \\
& Word Removing & 98.8 & -0.5 & -0.79 & 0.43 & 95.3 & -0.6 & -0.20 & 0.84 & 94.6 & -1.3 & 2.14 & \textbf{0.03*} \\
& Word Replacing & 98.4 & -0.9 & -2.11 & \textbf{0.04*} & 95.3 & -0.6 & -0.55 & 0.58 & 94.4 & -1.5 & 1.43 & 0.67 \\
\midrule
\multirow{3}{*}{MentalBERT(128)} & Raw & 99.4 & - & - & - & 96.0 & - & - & - & 96.2 & - & - & - \\
& Word Removing & 98.9 & -0.5 & -1.40 & 0.16 & 95.4 & -0.6 & 3.14 & \textbf{\textless{}0.01**} & 95.1 & -1.1 & 3.75 & \textbf{\textless{}0.01**} \\
& Word Replacing & 98.1 & -1.3 & -3.99 & \textbf{\textless{}0.01**} & 95.3 & -0.7 & 1.15 & 0.25 & 94.3 & -1.9 & 5.38 & \textbf{\textless{}0.01**} \\
\midrule
\multirow{3}{*}{NB(TDIDF: 5000)} & Raw & 98.6 & - & - & - & 93.8 & - & - & - & 90.2 & - & - & - \\
& Word Removing & 98.2 & -0.4 & -0.22 & 0.83 & 93.3 & -0.5 & 1.22 & 0.22 & 88.3 & -1.9 & 8.14 & \textbf{\textless{}0.01**} \\
& Word Replacing & 98.0 & -0.6 & -4.33 & \textbf{\textless{}0.01**} & 93.0 & -0.8 & -2.47 & \textbf{\textless{}0.01**} & 88.1 & -2.1 & 7.80 & \textbf{\textless{}0.01**} \\
\midrule
\multirow{3}{*}{SVM(TDIDF: 5000)} & Raw & 98.5 & - & - & - & 94.7 & - & - & - & 93.4 & - & - & - \\
& Word Removing & 97.6 & -0.9 & 0.34 & 0.73 & 94.0 & -0.7 & 2.14 & \textbf{0.03*} & 91.2 & -2.2 & 6.30 & \textbf{\textless{}0.01**} \\
& Word Replacing & 97.0 & -1.5 & -4.10 & \textbf{\textless{}0.01**} & 93.7 & -1.0 & -0.87 & 0.39 & 91.6 & -1.8 & 3.09 & \textbf{\textless{}0.01**} \\
\midrule
\multirow{3}{*}{LR(TDIDF: 5000)} & Raw & 96.5 & - & - & - & 93.7 & - & - & - & 90.3 & - & - & - \\
& Word Removing & 95.0 & -1.5 & 4.55 & \textbf{\textless{}0.01**} & 93.9 & 0.2 & 2.12 & \textbf{0.03*} & 87.0 & -3.3 & 11.60 & \textbf{\textless{}0.01**} \\
& Word Replacing & 94.5 & -2.0 & 3.27 & \textbf{\textless{}0.01**} & 92.6 & -1.1 & -0.15 & 0.88 & 86.5 & -0.5 & 10.43 & \textbf{\textless{}0.01**} \\
\midrule
\multirow{3}{*}{NB(unigram)} & Raw & 99.3 & - & - & - & 93.3 & - & - & - & 93.1 & - & - & - \\
& Word Removing & 98.7 & -0.6 & -1.15 & 0.25 & 91.4 & -1.9 & 2.54 & \textbf{\textless{}0.01**} & 92.3 & -0.8 & 3.00 & \textbf{\textless{}0.01**} \\
& Word Replacing & 98.1 & -1.2 & -3.62 & \textbf{\textless{}0.01**} & 90.3 & -3.0 & 0.61 & 0.54 & 91.0 & -2.1 & 1.84 & \textbf{0.07*} \\
\midrule
\multirow{3}{*}{SVM(unigram)} & Raw & 97.2 & - & - & - & 93.1 & - & - & - & 91.9 & - & - & - \\
& Word Removing & 99.5 & 2.3 & 2.31 & \textbf{0.02*} & 91.6 & -1.5 & 2.29 & \textbf{0.02*} & 88.8 & -3.1 & 8.94 & \textbf{\textless{}0.01**} \\
& Word Replacing & 93.7 & -3.5 & 3.41 & \textbf{\textless{}0.01**} & 90.1 & -3.0 & -1.49 & 0.14 & 86.6 & -5.3 & 9.02 & \textbf{\textless{}0.01**} \\
\bottomrule
\end{tabular}
\begin{tablenotes}
\small
\item Note: AccDiff indicates the difference of accuracy between raw data and word manipulation. Bold numbers indicate the significant differences between raw data and the manipulated data. *P-value\textless{}=0.05, **P-value\textless{}0.01.
\end{tablenotes}
\end{threeparttable}
\end{sidewaystable*}

\subsubsection{Phase 2 Topic-Word Manipulations Results}

\autoref{table:decoding6} shows how each model responds to changes in topic words across three testing sets. Overall, traditional shallow learners and MentalBERT were highly sensitive to word removal or replacement (P\textless{}0.05), with notable declines in accuracy and F1 scores. For Mix-Depression, most models (including MentalBERT) were affected by both manipulations, except Naïve Bayes (Unigram) under word replacement. In ENS-Depression, word removal had the greatest impact, while GNS-Depression performance was primarily influenced by word replacement, notably affecting Logistic Regression (TF-IDF:5000) and SVM (Unigram).

Despite being tailored for psychiatric contexts, MentalBERT still depended on explicit terms to maintain high accuracy, suggesting limited utility when mental health keywords are absent. By contrast, BERT(128) proved more robust in the ENS-Depression set, hinting that it captures subtler linguistic cues beyond surface-level keywords. Consequently, BERT(128) may be better suited for real-world applications where individuals rarely use explicit clinical language.
\begin{table*}[htbp]
\centering
\fontsize{8.5pt}{12pt}\selectfont
\begin{threeparttable}
\caption{Performance Change and T-test Result of Sentence Manipulations across Different Models and Different Testing Sets.}
\label{table:decoding7}
\begin{tabular}{@{}l|l|l|cccc@{}}
\toprule
Model & Testing Set & Approach & Accuracy(\%) & AccDiff(\%) & T Score & P-value \\ 
\midrule
\multirow{9}{*}{BERT(128)-GNS} & \multirow{3}{*}{GNS-Depression} & Raw & 99.3 & - & - & - \\
 &  & CPS. & 100.0 & 0.7 & -0.46 & 0.65 \\
 &  & WPS. & 99.4 & 0.1 & 0.75 & 0.45 \\ 
\cmidrule(l){2-7} 
 & \multirow{3}{*}{GNS-Anxiety} & Raw & 99.4 & - & - & - \\
 &  & CPS. & 100.0 & 0.6 & -0.47 & 0.64 \\
 &  & WPS. & 99.1 & -0.3 & 0.18 & 0.86 \\ 
\cmidrule(l){2-7} 
 & \multirow{3}{*}{GNS-Suicide} & Raw & 99.4 & - & - & - \\
 &  & CPS. & 99.9 & 0.5 & -0.43 & 0.67 \\
 &  & WPS. & 99.4 & 0.0 & 0.04 & 0.97 \\ 
\cmidrule(l){2-7} 
 & \multirow{3}{*}{GNS-Bipolar} & Raw & 99.5 & - & - & - \\
 &  & CPS. & 100.0 & 0.5 & -0.58 & 0.56 \\
 &  & WPS. & 99.5 & 0.0 & 0.12 & 0.90 \\ 
\midrule
\multirow{9}{*}{BERT(128)-ENS} & \multirow{3}{*}{ENS-Depression} & Raw & 95.9 & - & - & - \\
 &  & CPS. & 94.8 & -1.1 & -1.79 & \textbf{\textless{}0.01**} \\
 &  & WPS. & 94.4 & -1.5 & -3.00 & \textbf{0.05*} \\ 
\cmidrule(l){2-7} 
 & \multirow{3}{*}{ENS-Anxiety} & Raw & 95.7 & - & - & - \\
 &  & CPS. & 95.2 & -0.5 & -2.01 & \textbf{\textless{}0.01**} \\
 &  & WPS. & 94.6 & -1.1 & -3.33 & \textbf{0.04*} \\ 
\cmidrule(l){2-7} 
 & \multirow{3}{*}{ENS-Suicide} & Raw & 96.2 & - & - & - \\
 &  & CPS. & 94.8 & -1.4 & -2.19 & \textbf{\textless{}0.01**} \\
 &  & WPS. & 95.1 & -1.1 & -2.68 & \textbf{0.03*} \\ 
\cmidrule(l){2-7} 
 & \multirow{3}{*}{ENS-Bipolar} & Raw & 96.0 & - & - & - \\
 &  & CPS. & 95.3 & -0.7 & -2.08 & \textbf{\textless{}0.01**} \\
 &  & WPS. & 94.1 & -1.9 & -2.93 & \textbf{0.04*} \\ 
\midrule

\multirow{12}{*}{MentalBERT(128)-ENS} & \multirow{3}{*}{ENS-Depression} & Raw & 96.0 & - & - & - \\
 &  & CPS. & 96.7 & 0.7 & -1.29 & 0.20 \\
 &  & WPS. & 94.6 & -1.4 & -2.92 & \textbf{\textless{}0.01**} \\ 
\cmidrule(l){2-7} 
 & \multirow{3}{*}{ENS-Anxiety} & Raw & 95.9 & - & - & - \\
 &  & CPS. & 97.2 & 1.3 & -1.73 & \textbf{0.08*} \\
 &  & WPS. & 95.4 & -0.5 & -5.79 & \textbf{\textless{}0.01**} \\ 
\cmidrule(l){2-7} 
 & \multirow{3}{*}{ENS-Suicide} & Raw & 96.0 & - & - & - \\
 &  & CPS. & 96.7 & 0.7 & -0.91 & 0.36 \\
 &  & WPS. & 95.0 & -1.0 & -4.28 & \textbf{\textless{}0.01**} \\ 
\cmidrule(l){2-7} 
 & \multirow{3}{*}{ENS-Bipolar} & Raw & 96.7 & - & - & - \\
 &  & CPS. & 97.2 & 0.5 & -1.57 & 0.12 \\
 &  & WPS. & 95.6 & -1.1 & -4.87 & \textbf{\textless{}0.01**} \\ 
\bottomrule
\end{tabular}
\smallskip
\begin{tablenotes}
            \small
            \item Note: CPS. = Cross-Post Shuffling; WPS. = Within-Post Shuffling
            \item AccDiff indicates the difference of accuracy between raw data and sentence manipulation. Bold numbers indicate the significant differences between raw data and the manipulated data. Note: *P-value\textless{}=0.05, **P-value\textless{}0.01
\end{tablenotes}
\end{threeparttable}
\end{table*}

\subsubsection{Phase 3 Logical Connection Analyses Results}

\autoref{table:decoding7}, shows classification outcomes under both Within-Post and Cross-Post sentence shuffling, along with paired t-tests comparing shuffled data to the original. Notably, shuffling had little effect on GNS-related tests, but significantly impacted ENS-related sets: Within-Post Shuffling altered performance at p\textless{}0.05, while Cross-Post Shuffling caused even stronger differences at p\textless{}0.01. The Mix set, which combines various ENS, was also significantly affected by Cross-Post but less so by Within-Post Shuffling.

Interestingly, BERT(128)-GNS showed no meaningful changes, and Cross-Post Shuffling even produced a slight accuracy uptick. Both BERT(128)-ENS and BERT(128)-Mix were more sensitive to shuffling, with Within-Post changes leading to larger performance drops. This underscores how critical a coherent sentence flow is for identifying mental health cues in ENS.

Overall, these results mirror insights from Phase 2: BERT(128) is notably resilient to disruptions in word usage or sentence structure, suggesting it captures underlying linguistic patterns beyond explicit keywords. In contrast, MentalBERT remains more sensitive to explicit word cues. Thus, while models like MentalBERT may excel in more controlled settings, BERT(128) appears more robust for real-world mental health analysis, where narrative disorganization or minimal explicit terminology is common.

\subsection{Custom GPT Analysis and Results}
\label{sec:GPT_result}

This section evaluates the performance of the models for both tasks, beginning with an examination of execution times across different AI models.

Execution times varied significantly among the models tested. OpenAI's models, including GPT-4 and GPT-4o, processed Task 1 in approximately 10-20 seconds and completed the three-stage Task 2 in about 1.5 to 2 minutes. In contrast, Mixtral-8*7B, despite utilizing an NVIDIA A100 GPU with 80GB of memory, took an average of 12 minutes to produce a single output for both Task 1 and each stage of Task 2. To maintain the integrity of the test conditions, any single output from Mixtral-8*7B that exceeded three hours were excluded from the analysis.

\subsubsection{Task 1: Classification Result}
In the evaluation of Psycho Analyst's performance, the 7-point likelihood scores provided by Psycho Analyst were binarized to classify subjects as either having or not having mental health issues. To determine the optimal threshold that would reflect the most accurate classification by the model, we tried threshold values from 3 to 7 (cut off at \textgreater{}= the threshold value) and model accuracy for the four backgound knowledge configurations across different score threshold values are shown in Fig. 3. The results indicate that score threshold 5 consistently yielded the highest model accuracy. Therefore, we used score 5 as the cut-off point to binarized the 7-point likelihood scores provided by Psycho Analyst in subsequent evaluations of Psycho Analyst's performance.
\begin{figure}[h]
  \centering
  \includegraphics[width=\columnwidth]{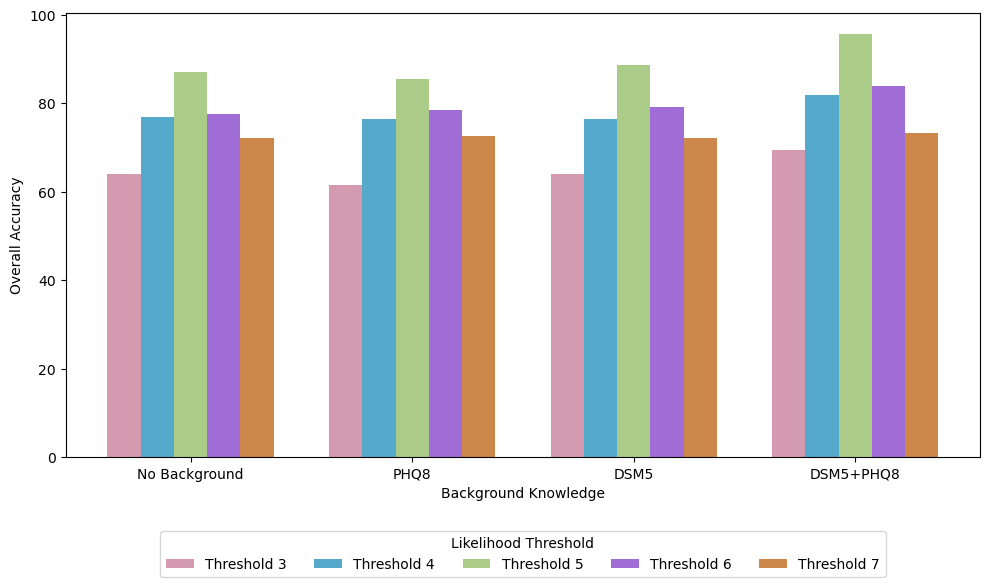} 
  \caption{Psycho Analyst prediction accuracy on the overall DAIC-WOZ dataset for the four different background knowledge configurations and likelihood score threshold values from 3 to 7.}
  \label{fig:Advanced_likelihood}
\end{figure}

\begin{table*}[t]
\centering
\fontsize{10pt}{16pt}\selectfont
\caption{Performance Comparison of Psycho Analyst on the Overall DAIC-WOZ Dataset with Various Background Knowledge Configurations}
\label{tab:Advanced_table5}
\begin{tabular}{l|ccccccc}
\toprule
Background Knowledge & F1 & Macro-F1 & Accuracy & Recall & Precision & ROC-AUC \\
\midrule
GPT4 + No Background & 0.769 & 0.840 & 0.872 & 0.727 & 0.816 & 0.853 \\

GPT4 + PHQ-8 & 0.738 & 0.819 & 0.856 & 0.691 & 0.792 & 0.851 \\

GPT4 + DSM-5 & 0.792 & 0.858 & 0.888 & 0.727 & 0.870 & 0.882 \\

GPT4 + DSM-5 + PHQ-8 & \textbf{0.929} & \textbf{0.949} & \textbf{0.957} & \textbf{0.945} & \textbf{0.912} & \textbf{0.968} \\

\bottomrule
\end{tabular}
\end{table*}

\begin{sidewaystable*}[htbp]
\centering
\fontsize{11pt}{16pt}\selectfont
\begin{threeparttable}
\caption{Performance comparison of Psycho Analyst on the DAIC-WOZ Test Set with Various Configurations and Models}
\label{tab:Advanced_table6}
\begin{tabular}{l|cccccc}
\toprule
Models and Background Knowledge &    F1 &  Macro-F1 &  Accuracy &  Recall &  Precision &  ROC-AUC \\
\midrule
GPT4 + No Background & 0.759 & 0.825 & 0.851 & 0.786 & 0.733 & 0.827 \\
GPT4 + PHQ-8 & 0.690 & 0.776 & 0.808 & 0.714 & 0.667 & 0.804 \\ 
GPT4 + DSM-5 & 0.786 & 0.847 & 0.872 & 0.786 & 0.786 & 0.860 \\ 
GPT4 + DSM-5 + PHQ-8 & \textbf{0.857} & \textbf{0.898} & \textbf{0.915} & \textbf{0.857} & \textbf{0.857} & \textbf{0.905} \\
GPT4 + DSM-5 + PHQ-8 + Data Description and Training Set & \textbf{0.929} & \textbf{0.949} & \textbf{0.957} & \textbf{0.929} & \textbf{0.929} & \textbf{0.952} \\ \hline
GPT-4o + No Background & 0.667 & 0.699 & 0.702 & 0.1.00 & 0.500 & 0.838 \\
Mixtral-8*7B  + No Background & 0.649 & 0.680 & 0.683 & 0.923 & 0.500 & 0.780 \\
\bottomrule
\end{tabular}
\begin{tablenotes}
\item Note: The results for Mixtral-8*7B were computed after excluding 6 missing values.
\end{tablenotes}
\end{threeparttable}
\end{sidewaystable*}

\autoref{tab:Advanced_table5} compares Psycho Analyst's performance across four configurations of background knowledge in the complete DAIC-WOZ dataset, assessing metrics such as F1 score, Macro-F1 score, Accuracy, Recall, Precision, and ROC-AUC score.

With no background knowledge, the model achieves an F1 score of 0.769 and an Accuracy of 0.872. Adding PHQ-8 slightly lowers all performance metrics, while incorporating DSM-5 alone enhances them, yielding an F1 score of 0.792 and an Accuracy of 0.888. The combination of DSM-5 and PHQ-8 results in the highest improvements, achieving an F1 score of 0.929, Accuracy of 0.957, Recall of 0.945, Precision of 0.912, and ROC-AUC of 0.968, highlighting the synergistic effect of these tools in boosting diagnostic accuracy and reliability.

Further insights are provided in \autoref{tab:Advanced_table6}, which compares the performance of the Psycho Analyst model on the DAIC-WOZ Test Set across various configurations. This includes an enhanced configuration that integrates DSM-5, PHQ-8, a detailed data description from the dataset’s curator, and an expanded training set. Performance is also compared with two baseline models.

The GPT-4 model without background knowledge shows moderate performance, with an F1 score of 0.759 and an Accuracy of 0.851. Adding PHQ-8 decreases performance to an F1 score of 0.690 and an Accuracy of 0.808, while including DSM-5 enhances outcomes to an F1 score of 0.786 and an Accuracy of 0.872. The combined use of DSM-5 and PHQ-8 significantly boosts performance, similar to previous results, achieving an F1 score of 0.857 and Accuracy of 0.915. Additional enhancements from incorporating detailed data descriptions and an expanded training set further improve metrics, elevating the F1 score to 0.929 and Accuracy to 0.957.

Comparative analysis with GPT-4o and Mixtral-8*7B, both lacking background knowledge, highlights Psycho Analyst's superior performance. GPT-4o reaches an F1 score of 0.667 and accuracy of 0.702, while Mixtral-8*7B shows slightly lower scores, emphasizing Psycho Analyst’s advantages even without enhanced configurations.

These results indicate that the incorporation of diverse background knowledge, along with relevant data descriptions and training data, significantly enhances Psycho Analyst GPT's diagnostic capabilities. The improved performance demonstrates its potential as a robust tool for mental health pre-screening.

\subsubsection{Task 2: PHQ-8 Score Computation Result}

\begin{figure*}[h!]
  \centering
  \includegraphics[width=\linewidth]{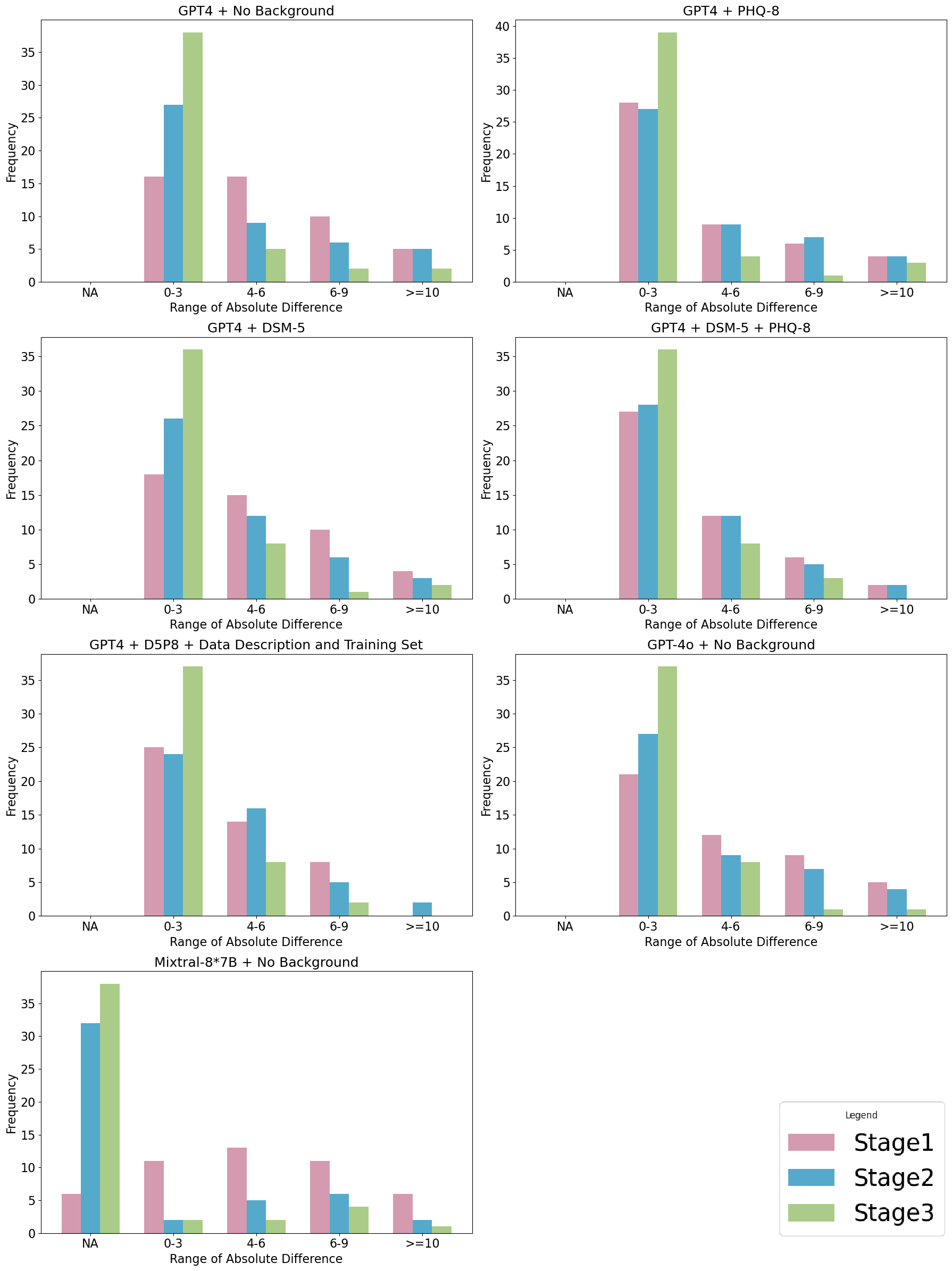}
  \caption{Frequency Distribution of Absolute Differences Across Three Stages of Evaluation for Various Models on the DAIC-WOZ Test Set.}
  \label{fig:Advanced_output}
\end{figure*}

\begin{sidewaystable*}[htbp]
\centering
\fontsize{10pt}{16pt}\selectfont
\begin{threeparttable}
\caption{Comparative Performance of 'Psycho Analyst' in Three-Stage PHQ-8 Analyses with Various Configurations Against Baseline Models GPT-4o and Mixtral-8*7B on the DAIC-WOZ Test Set}
\label{tab:Advanced_MAE}
\begin{tabular}{lccccccccc}
\toprule
 & \multicolumn{3}{c}{Stage1} & \multicolumn{3}{c}{Stage2} & \multicolumn{3}{c}{Stage3} \\
\cmidrule(lr){2-4} \cmidrule(lr){5-7} \cmidrule(lr){8-10}
Model and BackgroundKnowledge & MAE & RMSE & R-squared & MAE & RMSE & R-squared & MAE & RMSE & R-squared \\
\midrule
GPT4 + No Background & 4.42 & 5.59 & 0.24 & 3.53 & 5.00 & 0.39 & 1.57 & 2.72 & 0.82 \\
GPT4 + PHQ-8 & 3.28 & 4.39 & 0.53 & 3.32 & 4.49 & 0.51 & 1.68 & 2.95 & 0.79 \\
GPT4 + DSM-5 & 3.98 & 5.22 & 0.33 & 3.00 & 4.19 & 0.57 & 1.57 & 2.78 & 0.81 \\
GPT4 + DSM-5 + PHQ-8 & 3.04 & 4.16 & 0.58 & \textbf{2.91} & 3.98 & 0.61 & 1.64 & 2.59 & 0.84 \\
GPT4 + D5P8 + Data Description and Training Set  & \textbf{2.89} & \textbf{3.69} & \textbf{0.67} & 2.96 & \textbf{3.81} & \textbf{0.64} & 1.53 & 2.26 & 0.88 \\
GPT-4o + No Background & 4.04 & 5.40 & 0.29 & 3.45 & 4.78 & 0.44 & 1.29 & 2.44 & 0.85 \\
Mixtral 8*7B & 4.88 & 6.11 & 0.14 & 5.93 & 7.35 & -0.50 & 5.44 & 6.94 & 0.00  \\
\bottomrule
\end{tabular}
\begin{tablenotes}
\item * D5P8 is DSM-5 + PHQ-8.
\item * A negative R-squared value indicates that the model performs worse than a model that simply predicts the mean of the dependent variable for all observations, suggesting poor model fit or specification.
\item * For Mixtral 8*7B, all metrics were calculated after removing missing values.
\end{tablenotes}
\end{threeparttable}

\end{sidewaystable*}

\autoref{fig:Advanced_output} illustrates the distribution of absolute differences for the three-stage PHQ-8 score computation. The histograms depict the frequency of absolute differences between the true PHQ-8 scores and those estimated by various configurations of the GPT-4 model, the GPT-4o model, and the Mixtral-8*7B model. Each bar represents the frequency of a specific range of differences, demonstrating the accuracy of each configuration in accurately estimating mental health statuses.

When GPT-4 is enhanced with the PHQ-8 tool, both the initial assessment in Stage 1 and the detailed breakdown in Stage 2 exhibit similar performance, with minimal variation in the distribution of absolute differences. This consistency suggests that the inclusion of PHQ-8 provides a stable and reliable basis for both the initial analysis and the subsequent detailed scoring breakdown.

For other configurations of GPT models excluding PHQ-8, a noticeable improvement in performance during the detailed breakdown (Stage 2) is observed compared to the initial analysis (Stage 1). This indicates that a reassessment and detailed exploration of the initial model's assessment can significantly refine the accuracy of the models' evaluations, correcting any deficiencies noted in the initial analysis.

Moreover, in Stage 3, where other evaluators' opinions were provided as a reference, the GPT models still demonstrated a significant degree of autonomy. Despite the influence of external assessments, these models maintained a substantial level of self-decision, indicating their robustness in integrating and yet independently assessing complex clinical information.

The performance of Mixtral-8*7B is markedly different from the other models. A significant number of assessments were marked as 'NA' (not available), indicating that the model failed to complete the 3-stage task in many instances. Only a few cases reached all the way through Stage 3, suggesting substantial issues with data handling or model capability. This extensive occurrence of missing values undermines the value of Mixtral-87B's computations in this task, rendering it less valuable.

\autoref{tab:Advanced_MAE} presents detailed metrics of these assessments, including Mean Absolute Error (MAE), Root Mean Square Error (RMSE), and R-squared values for each model configuration. The results indicate that the Psycho Analyst model equipped with both DSM-5 and PHQ-8 significantly outperforms other configurations. This performance is further enhanced when additional data descriptions and training data are included. This table allows for a comprehensive comparison of performance across all stages of the PHQ-8 score computation task.


\begin{table*}[!t]
\centering
\begin{threeparttable}
\caption{Performance Enhancement for SMMR Across DAIC-WOZ Subsets}
\label{tab:SMMR_smmr}
\fontsize{8.5}{14}\selectfont 
\setlength{\tabcolsep}{1.5pt} 
\begin{tabular}{l|l|lccccccccc}
\toprule
\textbf{Dataset} & \textbf{Model} & \textbf{Method} & \textbf{Acc.} & \textbf{F1} & \textbf{MF1.} & \textbf{MPrec.} & \textbf{MRec.} & \textbf{ROC AUC} & \textbf{MAE} & \textbf{RMSE} \\
\midrule
\multirow{4}{*}{Training} & \multirow{2}{*}{GPT-3.5-turbo} & Baseline & 0.51 & 0.49 & 0.51 & 0.60 & 0.61 & 0.61 & 5.23 & 6.17 \\
                          &                                & SMMR Enhanced & \textbf{0.69} & \textbf{0.59} & \textbf{0.67} & \textbf{0.68} & \textbf{0.72} & \textbf{0.72} & \textbf{4.19} & \textbf{5.18} \\
                          & \multirow{2}{*}{GPT-4-turbo}   & Baseline & 0.76 & 0.64 & 0.73 & 0.72 & 0.75 & 0.75 & 3.32 & 4.07 \\
                          &                                & SMMR Enhanced & \textbf{0.79} & \textbf{0.69} & \textbf{0.76} & \textbf{0.75} & \textbf{0.79} & \textbf{0.79} & 3.33 & 4.24 \\
\midrule
\multirow{4}{*}{Testing}  & \multirow{2}{*}{GPT-3.5-turbo} & Baseline & 0.55 & 0.57 & 0.55 & 0.70 & 0.68 & 0.68 & 6.04 & 6.81 \\
                          &                                & SMMR Enhanced & \textbf{0.76} & \textbf{0.70} & \textbf{0.75} & \textbf{0.76} & \textbf{0.81} & \textbf{0.81} & \textbf{4.22} & \textbf{5.54} \\
                          & \multirow{2}{*}{GPT-4-turbo}   & Baseline & 0.77 & 0.67 & 0.74 & 0.74 & 0.77 & 0.77 & 3.21 & 4.16 \\
                          &                                & SMMR Enhanced & \textbf{0.80} & \textbf{0.74} & \textbf{0.79} & \textbf{0.79} & \textbf{0.84} & \textbf{0.84} & 3.76 & 4.78 \\
\midrule
\multirow{4}{*}{Validation}& \multirow{2}{*}{GPT-3.5-turbo} & Baseline & 0.60 & 0.63 & 0.60 & 0.73 & 0.70 & 0.70 & 4.86 & 5.54 \\
                           &                                & SMMR Enhanced & \textbf{0.80} & \textbf{0.76} & \textbf{0.79} & \textbf{0.80} & \textbf{0.83} & \textbf{0.83} & \textbf{3.29} & \textbf{3.96} \\
                           & \multirow{2}{*}{GPT-4-turbo}   & Baseline & 0.83 & 0.75 & 0.81 & 0.81 & 0.81 & 0.81 & 2.54 & 3.08 \\
                           &                                & SMMR Enhanced & \textbf{0.83} & \textbf{0.79} & \textbf{0.82} & \textbf{0.82} & \textbf{0.85} & \textbf{0.85} & \textbf{2.43} & 3.37 \\
\bottomrule
\end{tabular}
\begin{tablenotes}
\small
\item Note: Acc, MF1, MPrec, and MRec stand for Accuracy, Macro F1, Macro Precision, and Macro Recall, respectively.
\end{tablenotes}
\end{threeparttable}
\end{table*}
\begin{table}[!t]
  \centering
  \fontsize{12pt}{16pt}\selectfont
  \begin{threeparttable}
    \caption{Mental Health Detection on Case Study Dataset}
    \label{tab:SMMR_case_study_table}
    \begin{tabular}{lccccc}
      \hline
      \textbf{Method} & \textbf{Valid (\%)} & \textbf{Acc.} & \textbf{F1} & \textbf{Ave.} & \textbf{SD} \\
      \hline
      GPT-4o Baseline      & 92   & 0.95 & 0.98 & 6.97 & 3.20 \\
      GPT-4o+SMMR  & \textbf{100}  & 0.93 & 0.97 & 6.85 & \textbf{2.90} \\
      GPT-3.5 Baseline      & 98   & 0.91 & 0.95 & 6.66 & 3.17 \\
      GPT-3.5+SMMR & \textbf{100}  & \textbf{0.93} & \textbf{0.97} & \textbf{7.03} & \textbf{2.57} \\
      GPT-4 Baseline        & 100  & 0.92 & 0.96 & 7.02 & 2.95 \\
      GPT-4+SMMR   & 100  & 0.91 & 0.95 & \textbf{7.40} & \textbf{2.67} \\
      \hline
    \end{tabular}
    \begin{tablenotes}
      \item Note: This table shows the results for both binary mental health detection (accuracy and F1 scores) and the correctness of disorder type identification (Ave, SD).
    \end{tablenotes}
  \end{threeparttable}
\end{table}

\subsection{SMMR Results}
\label{sec:SMMR_result}

We evaluated the SMMR framework on both DAIC-WOZ and our curated case studies, comparing its performance against single-model baselines. \autoref{tab:SMMR_smmr} and \autoref{tab:SMMR_case_study_table} present the key metrics, reflecting a consistent advantage for SMMR-enhanced approaches.

\subsubsection{DAIC-WOZ Performance}
\autoref{tab:SMMR_smmr} highlights the performance of GPT-3.5-turbo and GPT-4-turbo across training, testing, and validation splits. Alongside classification measures such as Accuracy (Acc.), F1, Macro F1 (MF1.), Macro Precision (MPrec.), Macro Recall (MRec.), and ROC AUC (for PHQ-8 $\geq 10$ classification), we also track Mean Absolute Error (MAE) and Root Mean Squared Error (RMSE) for PHQ-8 scoring. Overall, the SMMR-enhanced models outperform baselines on every split, with particularly notable improvements in binary classification accuracy and F1. For example, GPT-3.5-turbo sees its testing accuracy jump from 0.55 to 0.76, while MAE decreases significantly (6.04 to 4.22), suggesting better handling of nuanced long-context data.

Importantly, SMMR also boosts consistency across different subsets. The validation accuracy for GPT-3.5-turbo, for instance, jumps from 0.60 to 0.80, showing that leveraging multiple “experts” and iterative refinement reduces hallucinations and more accurately captures PHQ-8 severity signals.

\subsubsection{Case Study Dataset Performance}
\autoref{tab:SMMR_case_study_table} summarizes results on our 48 narrative case studies, where we evaluate binary detection of mental health concerns and the correctness of disorder-type identification. Although the baseline metrics here are relatively high for GPT-4 variants (exceeding 0.90 accuracy), SMMR demonstrates small but meaningful gains and, in some configurations, yields a 100\% valid output rate. Notably, GPT-3.5 sees its accuracy rise from 0.91 to 0.93 under SMMR, while its F1 measure improves from 0.95 to 0.97. Beyond classification, SMMR also stabilizes disorder-type identification; for instance, the average correctness (Ave.) for GPT-4 climbs from 7.02 to 7.40, indicating richer extraction of relevant diagnostic details.

In summary, these findings confirm that layering multiple models curbs inconsistencies and incomplete responses, leading to more robust and thorough assessments in both interview transcripts and case narratives.


\section{Synthesis of Intervention Tools}
In Chapter~\ref{chap:intervention}, we presented three AI-driven methods for early intervention for mental health:
\begin{enumerate}
    \item Decoding Linguistic Nuances – focuses on subtle signals in shorter texts (see \autoref{sec:Section 4.2} \cite{tang2024decoding}).
    \item Psycho Analyst – a custom GPT model guided by clinical standards (DSM-5, PHQ-8) (see \autoref{sec:Section 4.3} \cite{tang2024advancing}).
    \item SMMR – a multi-expert framework designed for longer, more complex transcripts (see \autoref{sec:Section 4.4} \cite{tang2025layered}).
\end{enumerate}
Below, we look at how these tools perform overall and why they matter for campus well-being.

\subsection{Recap of Key Findings}
A central theme emerging from early detection tools is that all three intervention tools can detect early signs of depression or distress more accurately than simple baseline methods. The Decoding approach helps capture subtle signals in shorter posts by measuring how even small changes in wording or topic words affect classification results. Meanwhile, Psycho Analyst demonstrated strong performance on clinically oriented datasets like DAIC-WOZ, especially when equipped with standard clinical guidelines (DSM-5, PHQ-8) and thorough domain prompts. Lastly, the SMMR framework consistently improved reliability in long transcripts by combining multiple “expert” models and then refining their outputs layer by layer.

Beyond raw accuracy or F1 scores, these tools showed two broader benefits. First, they can adapt to different data scales—from short social media updates to extended interviews—making them versatile enough for the wide range of inputs that campus mental health services might see. Second, the multi-stage frameworks (particularly Psycho Analyst’s three-step PHQ-8 scoring and SMMR’s layered approach) provide opportunities to revisit and refine initial outputs. This iterative process not only improves precision but also gives counselors or administrators a clearer look at how final decisions are reached.

\subsection{Coverage of Different Types of Data}
One key difference among these methods is the kind of text they target. Decoding Linguistic Nuances works well with shorter or medium-length social media posts (for example, those on Reddit), capturing how slight changes in wording or context can affect predictions. Psycho Analyst uses GPT-based processing for more formal transcripts, such as the DAIC-WOZ interviews, and ties its findings to standard clinical measures like the PHQ-8. SMMR merges results from multiple models—some of which can handle large amounts of text—so it is especially useful when dealing with longer interviews or narrative case studies. By combining these three, we cover short posts, in-depth interviews, and everything in between.

\subsection{Emphasis on Subtle Language Clues}
Although they vary in style, all three tools rely on the idea that mental health signals can be subtle. Decoding Linguistic Nuances shows that even removing certain keywords or rearranging sentences can change how a model perceives depression-related content. Psycho Analyst focuses on a more structured approach, mapping user statements directly onto clinical guidelines (DSM-5 and PHQ-8). SMMR addresses the risk of confusion or “hallucination” by gathering “second opinions” from multiple experts before making a final call. These shared insights underscore that just matching keywords is often not enough: truly effective screening must consider the context and flow of a person’s words.

\subsection{Complementary Strengths}
From an evaluation standpoint, each tool has its own strengths. Decoding is relatively lightweight and effective for quick classification on shorter text, even when key terms might be missing. Psycho Analyst provides strong clinical insight on DAIC-WOZ data, producing accurate estimates of mental health severity. Meanwhile, SMMR excels at handling longer, more complex transcripts by combining multiple outputs from different experts to reduce errors. Put together, they offer a well-rounded suite of detection tools that can handle everything from short social media posts to extended counseling dialogues.

\subsection{Practical Considerations and Takeaways for Real-World Use}
Though the quantitative results in \autoref{sec:intervention_result} are promising, practical deployment on campus would require balancing automation with human oversight. For instance, Decoding Linguistic Nuances is efficient for scanning large volumes of brief student posts but does not in itself provide a “why” or interpretive explanation. Psycho Analyst offers more interpretability (it can produce mini-rationales for its assessments), yet it also depends on well-structured prompts and domain knowledge to remain accurate. SMMR, while robust against data gaps or disorganized transcripts, is more computationally intensive and may require careful scheduling if used for real-time triage.

In a real campus setting, different tools can be assigned to different steps. For example, short social media posts or initial brief text interactions could be analyzed by either Decoding methods or a first-stage GPT check (like the one used in Psycho Analyst) to spot early signs of distress. If these early checks suggest more severe concerns, SMMR could take a deeper look at extended interviews or multi-turn discussions, combining evidence from multiple sources. This tiered approach would allow universities to catch risk factors early, then dedicate more resources—like in-person counseling—for cases requiring a deeper evaluation. Ultimately, combining these three methods paves the way for a proactive and comprehensive mental health support system on campus.

Additionally, using these tools on a large scale means we must have clear rules to protect people’s privacy and handle their personal data responsibly. While AI systems can spot warning signs, campus leaders still need to decide how to follow up—whether by offering telehealth sessions (using online tools to provide healthcare or mental health services), connecting people with peer support, or recommending a more detailed evaluation by a mental health professional. As discussed in Chapter~\ref{chap:evaluation}, AI-based approaches can greatly increase how and when students, staff, or faculty get help. However, it’s important to remember that AI is not human and doesn’t replace professional care. These tools should always be part of a larger support system that includes trained experts and ethical data management practices.

\chapter{A New Proposed Framework for Integrating AI Tools into Campus Well-Being Services}
\label{chap:object3}

This dissertation has explored AI-driven solutions to improve campus well-being by addressing two core objectives. Objective 1 developed preventive methodologies, resulting in \textbf{TigerGPT} and, subsequently, \textbf{AURA}—an RL-based framework that adaptively selects follow-up question types within a conversation session to improve user's response quality across heterogeneous users. Objective 2 focused on intervention methodologies, creating AI tools for early detection and assessment of mental health issues—including \textbf{linguistic decoding}, \textbf{custom GPT models}, and \textbf{a layered multi-expert framework}. While these objectives individually tackle different facets of well-being, their combined potential has yet to be fully realized without a unified approach.

This chapter proposes a framework for Objective 3 that integrates these tools into a cohesive system, aiming to proactively identify concerns and deliver timely, personalized support for students, faculty, and staff. Additionally, it addresses ethical considerations—such as privacy and the limitations of AI in replacing human care—to ensure responsible implementation.

\section{Proposed Framework}
The framework is organized as a multi-layered system that links preventive and intervention strategies, harnessing AI-driven technologies to form a holistic approach to campus well-being.

\subsection{Preventive Layer: TigerGPT + AURA}
TigerGPT anchors the preventive layer by engaging users in dynamic conversations to gather detailed feedback on their campus experiences. Unlike static surveys, its dialogic format interprets nuanced responses, identifying early signs of dissatisfaction or distress. AURA augments this layer by introducing \emph{within-session} adaptation: after each user exchange, it computes a response-quality signal (LSDE: Length, Specificity, Self-disclosure, and Emotion) and updates expected gains for candidate follow-up types, choosing the next action with an $\epsilon$-greedy policy. In practice, TigerGPT provides stable, domain-grounded prompting and safety, while AURA reduces repetition and improve response quality by adjusting question types in real time. Pilot testing showed TigerGPT achieving a 75\% usability rating and an 81\% satisfaction score, underscoring its effectiveness as a frontline tool for monitoring campus sentiment.

At a high level, the preventive layer surfaces two complementary signals: (1) campus-climate themes distilled from the conversation, and (2) a compact summary of response quality across the session (primarily from AURA). Either signal can prompt a handoff to the intervention layer when warranted.

\subsection{Intervention Layer: Mental Health Assessment Tools}
The intervention layer leverages Objective 2’s tools to analyze text-based data—such as user dialogues, narrative transcripts, and social media posts—to detect and evaluate mental health concerns. These tools can integrate with mental health assessment system like SouLLMate \cite{guo2024soullmate} to enhance assessment capabilities. Key components include:
\begin{itemize}
    \item Linguistic Decoding: Identifies subtle linguistic cues (e.g., shifts in tone or phrasing) linked to mental distress, such as anxiety or depression.
    \item Custom GPT Models: Provides preliminary assessments using clinical benchmarks like the PHQ-8 or DSM-5, offering insights into distress severity.
    \item Layered Multi-Expert Framework (SMMR): Processes complex, long-context data by integrating multiple AI models, ensuring robust evaluations.
\end{itemize}

By utilizing these tools, the intervention layer enables the detection and evaluation of mental health concerns across various contexts, making it a versatile approach for supporting campus populations.

\section{Broader Benefits for Campus Well-being}
Integrating preventive and intervention-focused AI tools under a single framework yields tangible campus-wide benefits. TigerGPT’s adaptive surveys improve early detection of emerging problems, and AURA’s within-session optimization draws out specific, self-disclosing details that strengthen signal quality for triage. Advanced assessment ensures that individuals showing heightened risk receive targeted scrutiny. This structure encourages collaboration among student affairs, counseling services, and other campus offices, and enables data-informed decision making—allowing administrators to spot trends, allocate resources efficiently, and measure the impact of policy changes. Ultimately, embedding both prevention and intervention within the same overall system creates a more responsive, supportive campus environment where potential well-being issues can be addressed early and effectively.

\chapter{Discussion and Future Work}
\label{chap:discussion}
\section{Discussion}
In this dissertation, we introduced TigerGPT and AURA for improving campus satisfaction (Objective 1) and three AI-driven methods—Decoding Linguistic Nuances, Psycho Analyst, and SMMR—to detect mental health concerns (Objective 2). These efforts work together under a larger plan (Objective 3) to create an integrated, AI-powered framework for supporting well-being on campus. This section discusses their overall impact, limitations, and ethical considerations.

\subsection{Impact on Campus Well-Being}
\paragraph{Preventive Approaches.}
The TigerGPT chatbot represents a major shift from traditional, static surveys. By using adaptive dialogue and context-aware follow-up questions, TigerGPT helps universities gather richer, more specific feedback on students’ everyday experiences. In the pilot study, it achieved 81\% satisfaction and a 75\% usability rating, suggesting that conversational surveys can capture important insights that might be overlooked in ordinary questionnaires. Administrators can use these real-time findings to address emerging issues, such as resource shortages or social climate concerns, before they escalate. Complementing TigerGPT, AURA adds within-session adaptation: after each exchange it uses an LSDE quality signal to select follow-up question types, which reduces repetition and increases specificity and self-disclosure. In evaluation, a moderate exploration setting ($\epsilon=0.30$) sustained gains across early, mid, and late phases. Together, TigerGPT provides stable, domain-grounded elicitation, and AURA improves the quality and actionability of information for early sensing and targeted outreach.

\paragraph{Intervention Tools for Early Detection.} The other three models focus on identifying students who might need mental health support. Decoding Linguistic Nuances explores subtle language markers in expressive narratives, Psycho Analyst uses GPT-4-based analysis tailored with DSM-5 and PHQ-8 guidance, and SMMR applies multiple “experts” to reduce errors in long, complex transcripts. These tools can flag at-risk students or staff sooner, enabling timely referrals to counseling services or peer support groups.

\paragraph{Holistic Prevention + Intervention.}
By combining the preventive capabilities of TigerGPT and AURA with the mental health assessment tools (Decoding Linguistic Nuances, Psycho Analyst, and SMMR), universities can identify campus-wide pain points while also supporting individuals who show early signs of distress. This aligns with public-health practice: proactive action is typically more cost-effective and more inclusive than reacting after crises emerge. By strengthening everyday satisfaction and addressing mental-health needs early, institutions build a more resilient and caring environment for students, faculty, and staff.

\subsubsection{Limitations} 
Though these AI-driven tools show promise, several challenges remain. First, they rely heavily on the quality and diversity of training data. If the datasets do not sufficiently represent certain campus populations—like international students or minority groups—the tools may misinterpret their specific language patterns, inadvertently overlook emerging problems, or produce biased outcomes. For instance, while Decoding Linguistic Nuances performs well on Reddit-based texts, it could struggle with unique campus jargon if such expressions are absent from its training set. In addition, AURA’s priors are learned from previous TigerGPT logs; if those logs underrepresent some groups or states, early-session behavior may reflect that imbalance until sufficient new data accumulate.

Second, there is an ongoing risk of misclassification and proxy error. False positives in mental health assessments can unnecessarily alarm individuals, while false negatives can delay vital support. Psycho Analyst, although it achieves high accuracy in many cases, still experiences some performance variation. TigerGPT has not yet been tested at scale, making it unclear how well it will adapt to the linguistic diversity and real-world demands of a large campus community. AURA further introduces within-session exploration that can occasionally select locally suboptimal prompts early in a conversation, and its LSDE signal is a proxy for “response quality” that may encode stylistic biases. Ensuring robust and fair performance across varied user groups will require expanded training sets, subgroup audits, and regular monitoring.

Third, computational overhead remains a practical constraint. SMMR’s multi-expert structure is effective for handling long or complex data, but the multiple model calls can require substantial CPU or GPU resources, which might not be readily available in all campus environments. While TigerGPT and AURA are designed to operate with modest latency, production deployments still need capacity planning, caching, and fallback modes to maintain responsiveness during peak load. Balancing accuracy with efficiency is therefore essential for institutions hoping to deploy these systems widely, especially when budgets or technical infrastructures are limited.

\subsubsection{Ethical Considerations}

Integrating AI tools into campus well-being programs carries important ethical obligations, beginning with privacy and data security. TigerGPT, AURA, and the mental health assessment models handle sensitive personal information, including detailed experiences and potential mental health indicators, which demands rigorous safeguards. Strong encryption, data minimization, role-based access, and transparent consent are essential to maintain trust, especially given prior concerns about chatbot privacy. Participants should clearly understand what data are collected, how long they are retained, and with whom they may be shared, and they should have a clear opt-in choice. Where feasible, AURA should operate with session-local learning that avoids cross-session identity linkage.

Bias and fairness also require sustained attention. AI models can perpetuate inequities if training data are incomplete or skewed, which can lead to uneven performance for particular groups. If screening tools systematically under-identify or over-identify certain populations, access to care becomes unequal. Routine subgroup monitoring, dataset expansion to improve coverage, and fairness-aware training can reduce these risks. For AURA specifically, the LSDE signal is a proxy for response quality, so audits should examine subgroup differences in LSDE scoring and in the controller’s action choices and escalation rates.

A third consideration is balancing efficiency with human care. These technologies may streamline aspects of well-being support, but they must not replace human expertise, empathy, or judgment, particularly for vulnerable individuals. Over-reliance on automated decisions can erode the human connection students and staff seek. Tools like TigerGPT and AURA can surface concerns and summarize engagement quality, and Psycho Analyst can provide rubric-guided assessments, but human professionals should interpret outputs and decide on actions. Clear escalation criteria, de-escalation options, and documented handoff procedures help preserve a compassionate, person-centered process.

Finally, transparency and oversight are key. Users and administrators should understand how recommendations are generated and what limitations apply, particularly in mental health evaluations. SMMR’s multi-expert structure can aid interpretability by indicating how different models contribute to a final judgment. For AURA, documenting controller settings (for example, exploration rate and update rules), action families, and update history supports institutional oversight. Formal human-in-the-loop checkpoints, explainability reports, and periodic audits of privacy, fairness, and performance strengthen accountability and trust.

\subsection{Future Directions}

Bringing TigerGPT, AURA, and the mental health tools together under a single campus framework (Objective 3) opens new possibilities for prevention, detection, and targeted intervention. Ideally, regular conversational sensing from TigerGPT augmented by AURA’s within-session adaptation would provide two complementary signals for routing: campus-climate themes and compact summaries of response quality. These signals can feed advanced assessment models to prompt deeper investigations and more robust support when warranted. Future work should refine this integration layer, including clear handoff criteria, privacy-preserving data flows, and human-in-the-loop review. Methodologically, priorities include multi-objective tuning for AURA (balancing depth, specificity, and burden), off-policy evaluation from logs before deployment, fairness audits across subgroups, and cross-campus transfer studies to test generalizability.

In short, this dissertation shows that AI-based approaches can enhance campus satisfaction surveys and mental health assessments beyond what traditional methods offer. By bridging everyday feedback through TigerGPT and AURA with screening and analysis via Decoding Linguistic Nuances, Psycho Analyst, and SMMR, universities can identify issues earlier and support students, faculty, and staff before problems intensify, fostering a healthier and more responsive campus environment.


\clearpage
\phantomsection
\addcontentsline{toc}{chapter}{BIBLIOGRAPHY}
\bibliographystyle{IEEEtran}
\bibliography{ref}

@INPROCEEDINGS{tangshang2024advancing,
  author={Tang, Jinwen and Shang, Yi},
  booktitle={2024 IEEE 6th International Conference on Cognitive Machine Intelligence (CogMI)}, 
  title={Advancing Mental Health Pre-Screening: A New Custom GPT for Psychological Distress Assessment}, 
  year={2024},
  volume={},
  number={},
  pages={162-171},
  doi={10.1109/CogMI62246.2024.00030}}

@article{gulliver2010perceived,
  title={Perceived barriers and facilitators to mental health help-seeking in young people: a systematic review},
  author={Gulliver, Amelia and Griffiths, Kathleen M and Christensen, Helen},
  journal={BMC psychiatry},
  volume={10},
  number={1},
  pages={1--9},
  year={2010},
  publisher={BioMed Central}
}

@article{pescosolido1999people,
  title={How do people come to use mental health services? Current knowledge and changing perspectives.},
  author={Pescosolido, Bernice A and Boyer, Carol A},
  year={1999},
  publisher={Cambridge University Press}
}

@inproceedings{devault2014simsensei,
  title={SimSensei Kiosk: A virtual human interviewer for healthcare decision support},
  author={DeVault, David and Artstein, Ron and Benn, Grace and Dey, Teresa and Fast, Ed and Gainer, Alesia and Georgila, Kallirroi and Gratch, Jon and Hartholt, Arno and Lhommet, Margaux and others},
  booktitle={Proceedings of the 2014 international conference on Autonomous agents and multi-agent systems},
  pages={1061--1068},
  year={2014}
}

@article{regier1993facto,
  title={The de facto US mental and addictive disorders service system: Epidemiologic Catchment Area prospective 1-year prevalence rates of disorders and services},
  author={Regier, Darrel A and Narrow, William E and Rae, Donald S and Manderscheid, Ronald W and Locke, Ben Z and Goodwin, Frederick K},
  journal={Archives of general psychiatry},
  volume={50},
  number={2},
  pages={85--94},
  year={1993},
  publisher={American Medical Association}
}

@article{zhang2022natural,
  title={Natural language processing applied to mental illness detection: a narrative review},
  author={Zhang, Tianlin and Schoene, Annika M and Ji, Shaoxiong and Ananiadou, Sophia},
  journal={NPJ digital medicine},
  volume={5},
  number={1},
  pages={46},
  year={2022},
  publisher={Nature Publishing Group UK London}
}

@article{graham2019artificial,
  title={Artificial intelligence for mental health and mental illnesses: an overview},
  author={Graham, Sarah and Depp, Colin and Lee, Ellen E and Nebeker, Camille and Tu, Xin and Kim, Ho-Cheol and Jeste, Dilip V},
  journal={Current psychiatry reports},
  volume={21},
  pages={1--18},
  year={2019},
  publisher={Springer}
}

@article{shatte2019machine,
  title={Machine learning in mental health: a scoping review of methods and applications},
  author={Shatte, Adrian BR and Hutchinson, Delyse M and Teague, Samantha J},
  journal={Psychological medicine},
  volume={49},
  number={9},
  pages={1426--1448},
  year={2019},
  publisher={Cambridge University Press}
}

@article{tran2017predicting,
  title={Predicting mental conditions based on “history of present illness” in psychiatric notes with deep neural networks},
  author={Tran, Tung and Kavuluru, Ramakanth},
  journal={Journal of biomedical informatics},
  volume={75},
  pages={S138--S148},
  year={2017},
  publisher={Elsevier}
}

@phdthesis{belser2023comparison,
  title={Comparison of Natural Language Processing Models for Depression Detection in Chatbot Dialogues},
  author={Belser, Christian Alexander},
  year={2023},
  school={Massachusetts Institute of Technology}
}

@article{milintsevich2023towards,
  title={Towards automatic text-based estimation of depression through symptom prediction},
  author={Milintsevich, Kirill and Sirts, Kairit and Dias, Ga{\"e}l},
  journal={Brain Informatics},
  volume={10},
  number={1},
  pages={1--14},
  year={2023},
  publisher={SpringerOpen}
}

@article{yadav2023novel,
  title={A novel automated depression detection technique using text transcript},
  author={Yadav, Uma and Sharma, Ashish K},
  journal={International Journal of Imaging Systems and Technology},
  volume={33},
  number={1},
  pages={108--122},
  year={2023},
  publisher={Wiley Online Library}
}

@article{xu2024mental,
  title={Mental-llm: Leveraging large language models for mental health prediction via online text data},
  author={Xu, Xuhai and Yao, Bingsheng and Dong, Yuanzhe and Gabriel, Saadia and Yu, Hong and Hendler, James and Ghassemi, Marzyeh and Dey, Anind K and Wang, Dakuo},
  journal={Proceedings of the ACM on Interactive, Mobile, Wearable and Ubiquitous Technologies},
  volume={8},
  number={1},
  pages={1--32},
  year={2024},
  publisher={ACM New York, NY, USA}
}

@article{ng2019provider,
  title={Provider perspectives on integrating sensor-captured patient-generated data in mental health care},
  author={Ng, Ada and Kornfield, Rachel and Schueller, Stephen M and Zalta, Alyson K and Brennan, Michael and Reddy, Madhu},
  journal={Proceedings of the ACM on human-computer interaction},
  volume={3},
  number={CSCW},
  pages={1--25},
  year={2019},
  publisher={ACM New York, NY, USA}
}

@article{Marsolek20240969PD,
  title={0969 Predicting Differences Between Objective and Subjective Sleep Parameters with Mental Health Questionnaires},
  author={Mason Marsolek and Sarah Emert and Jessica R. Dietch and Emme Tucker and Daniel Taylor},
  journal={SLEEP},
  year={2024},
  url={https://api.semanticscholar.org/CorpusID:269288613}
}

@article{huang2023survey,
  title={A survey on hallucination in large language models: Principles, taxonomy, challenges, and open questions},
  author={Huang, Lei and Yu, Weijiang and Ma, Weitao and Zhong, Weihong and Feng, Zhangyin and Wang, Haotian and Chen, Qianglong and Peng, Weihua and Feng, Xiaocheng and Qin, Bing and others},
  journal={ACM Transactions on Information Systems},
  year={2023},
  publisher={ACM New York, NY}
}

@article{dahl2024large,
  title={Large legal fictions: Profiling legal hallucinations in large language models},
  author={Dahl, Matthew and Magesh, Varun and Suzgun, Mirac and Ho, Daniel E},
  journal={Journal of Legal Analysis},
  volume={16},
  number={1},
  pages={64--93},
  year={2024},
  publisher={Oxford University Press UK}
}

@article{tausczik2010psychological,
  title={The psychological meaning of words: LIWC and computerized text analysis methods},
  author={Tausczik, Yla R and Pennebaker, James W},
  journal={Journal of language and social psychology},
  volume={29},
  number={1},
  pages={24--54},
  year={2010},
  publisher={Sage Publications Sage CA: Los Angeles, CA}
}

@article{ohse2024zero,
  title={Zero-Shot Strike: Testing the generalisation capabilities of out-of-the-box LLM models for depression detection},
  author={Ohse, Julia and Had{\v{z}}i{\'c}, Bakir and Mohammed, Parvez and Peperkorn, Nicolina and Danner, Michael and Yorita, Akihiro and Kubota, Naoyuki and R{\"a}tsch, Matthias and Shiban, Youssef},
  journal={Computer Speech \& Language},
  volume={88},
  pages={101663},
  year={2024},
  publisher={Elsevier}
}

@article{agrawal2024illuminate,
  title={Illuminate: A novel approach for depression detection with explainable analysis and proactive therapy using prompt engineering},
  author={Agrawal, Aryan},
  journal={arXiv preprint arXiv:2402.05127},
  year={2024}
}

@article{openai2023gpt,
  title={Gpt-4 technical report. arxiv 2303.08774},
  author={OpenAI, R},
  journal={View in Article},
  volume={2},
  pages={13},
  year={2023}
}

@article{andrews2001shortfall,
  title={Shortfall in mental health service utilisation},
  author={Andrews, Gavin and Issakidis, Cathy and Carter, Greg},
  journal={The British Journal of Psychiatry},
  volume={179},
  number={5},
  pages={417--425},
  year={2001},
  publisher={Cambridge University Press}
}

@article{cavazos2016content,
  title={A content analysis of depression-related tweets},
  author={Cavazos-Rehg, Patricia A and Krauss, Melissa J and Sowles, Shaina and Connolly, Sarah and Rosas, Carlos and Bharadwaj, Meghana and Bierut, Laura J},
  journal={Computers in human behavior},
  volume={54},
  pages={351--357},
  year={2016},
  publisher={Elsevier}
}

@inproceedings{gratch2014distress,
  title={The distress analysis interview corpus of human and computer interviews.},
  author={Gratch, Jonathan and Artstein, Ron and Lucas, Gale M and Stratou, Giota and Scherer, Stefan and Nazarian, Angela and Wood, Rachel and Boberg, Jill and DeVault, David and Marsella, Stacy and others},
  booktitle={LREC},
  pages={3123--3128},
  year={2014},
  organization={Reykjavik}
}

@article{hart2001redeveloping,
  title={Redeveloping DICTION: theoretical considerations},
  author={Hart, Roderick P},
  journal={Progress in communication sciences},
  pages={43--60},
  year={2001},
  publisher={ABLEX PUBLISHING CORPORATION}
}

@article{le2021machine,
  title={Machine learning and natural language processing in mental health: systematic review},
  author={Le Glaz, Aziliz and Haralambous, Yannis and Kim-Dufor, Deok-Hee and Lenca, Philippe and Billot, Romain and Ryan, Taylor C and Marsh, Jonathan and Devylder, Jordan and Walter, Michel and Berrouiguet, Sofian and others},
  journal={Journal of Medical Internet Research},
  volume={23},
  number={5},
  pages={e15708},
  year={2021},
  publisher={JMIR Publications Toronto, Canada}
}

@misc{RN679,
   author = {National Institute of Mental Health},
   title = {Mental Illness Statistics},
   publisher = {National Institute of Mental Health},
   volume = {2022},
   number = {8/23},
   url = {https://www.nimh.nih.gov/health/statistics/mental-illness},
   year = {2021},
   type = {Web Page}
}

@article{pennebaker2003psychological,
  title={Psychological aspects of natural language use: Our words, our selves},
  author={Pennebaker, James W and Mehl, Matthias R and Niederhoffer, Kate G},
  journal={Annual review of psychology},
  volume={54},
  number={1},
  pages={547--577},
  year={2003},
  publisher={Annual Reviews 4139 El Camino Way, PO Box 10139, Palo Alto, CA 94303-0139, USA}
}

@inproceedings{ringeval2019avec,
  title={AVEC 2019 workshop and challenge: state-of-mind, detecting depression with AI, and cross-cultural affect recognition},
  author={Ringeval, Fabien and Schuller, Bj{\"o}rn and Valstar, Michel and Cummins, Nicholas and Cowie, Roddy and Tavabi, Leili and Schmitt, Maximilian and Alisamir, Sina and Amiriparian, Shahin and Messner, Eva-Maria and others},
  booktitle={Proceedings of the 9th International on Audio/visual Emotion Challenge and Workshop},
  pages={3--12},
  year={2019}
}

@article{mcadams2013narrative,
  title={Narrative identity},
  author={McAdams, Dan P and McLean, Kate C},
  journal={Current directions in psychological science},
  volume={22},
  number={3},
  pages={233--238},
  year={2013},
  publisher={Sage Publications Sage CA: Los Angeles, CA}
}

@inproceedings{wang2013depression,
  title={A depression detection model based on sentiment analysis in micro-blog social network},
  author={Wang, Xinyu and Zhang, Chunhong and Ji, Yang and Sun, Li and Wu, Leijia and Bao, Zhana},
  booktitle={Trends and Applications in Knowledge Discovery and Data Mining: PAKDD 2013 International Workshops: DMApps, DANTH, QIMIE, BDM, CDA, CloudSD, Gold Coast, QLD, Australia, April 14-17, 2013, Revised Selected Papers 17},
  pages={201--213},
  year={2013},
  organization={Springer}
}

@inproceedings{yazdavar2017semi,
  title={Semi-supervised approach to monitoring clinical depressive symptoms in social media},
  author={Yazdavar, Amir Hossein and Al-Olimat, Hussein S and Ebrahimi, Monireh and Bajaj, Goonmeet and Banerjee, Tanvi and Thirunarayan, Krishnaprasad and Pathak, Jyotishman and Sheth, Amit},
  booktitle={Proceedings of the 2017 IEEE/ACM international conference on advances in social networks analysis and mining 2017},
  pages={1191--1198},
  year={2017}
}

@article{yates2017depression,
  title={Depression and self-harm risk assessment in online forums},
  author={Yates, Andrew and Cohan, Arman and Goharian, Nazli},
  journal={arXiv preprint arXiv:1709.01848},
  year={2017}
}

@article{inamdar2023machine,
  title={Machine Learning Driven Mental Stress Detection on Reddit Posts Using Natural Language Processing},
  author={Inamdar, Shaunak and Chapekar, Rishikesh and Gite, Shilpa and Pradhan, Biswajeet},
  journal={Human-Centric Intelligent Systems},
  volume={3},
  number={2},
  pages={80--91},
  year={2023},
  publisher={Springer}
}

@misc{Rigoulet2021,
author = {Rigoulet, Xavier},
title = {New Dataset: Reddit r/depression and r/SuicideWatch for NLP},
year = {2021},
url = {http://data.worldbank.org/indicator/SP.DYN.LE00.FE.IN},
}

@article{devlin2018bert,
  title={Bert: Pre-training of deep bidirectional transformers for language understanding},
  author={Devlin, Jacob and Chang, Ming-Wei and Lee, Kenton and Toutanova, Kristina},
  journal={arXiv preprint arXiv:1810.04805},
  year={2018}
}

@article{pennebaker1986confronting,
  title={Confronting a traumatic event: toward an understanding of inhibition and disease.},
  author={Pennebaker, James W and Beall, Sandra K},
  journal={Journal of abnormal psychology},
  volume={95},
  number={3},
  pages={274},
  year={1986},
  publisher={American Psychological Association}
}

@article{davidson2002expressive,
  title={Expressive writing and blood pressure.},
  author={Davidson, Karina and Schwartz, Amy R and Sheffield, David and McCord, Ronald S and Lepore, Stephen J and Gerin, William},
  year={2002},
  publisher={American Psychological Association}
}

@article{pennebaker1988disclosure,
  title={Disclosure of traumas and immune function: health implications for psychotherapy.},
  author={Pennebaker, James W and Kiecolt-Glaser, Janice K and Glaser, Ronald},
  journal={Journal of consulting and clinical psychology},
  volume={56},
  number={2},
  pages={239},
  year={1988},
  publisher={American Psychological Association}
}

@article{rivkin2006effects,
  title={The effects of expressive writing on adjustment to HIV},
  author={Rivkin, Inna D and Gustafson, Julie and Weingarten, Ilene and Chin, Dorothy},
  journal={AIDS and Behavior},
  volume={10},
  pages={13--26},
  year={2006},
  publisher={Springer}
}

@article{jung2014emotions,
  title={How emotions affect logical reasoning: evidence from experiments with mood-manipulated participants, spider phobics, and people with exam anxiety},
  author={Jung, Nadine and Wranke, Christina and Hamburger, Kai and Knauff, Markus},
  journal={Frontiers in psychology},
  volume={5},
  pages={570},
  year={2014},
  publisher={Frontiers Media SA}
}

@article{ji2021mentalbert,
  title={Mentalbert: Publicly available pretrained language models for mental healthcare},
  author={Ji, Shaoxiong and Zhang, Tianlin and Ansari, Luna and Fu, Jie and Tiwari, Prayag and Cambria, Erik},
  journal={arXiv preprint arXiv:2110.15621},
  year={2021}
}

@book{videbeck2010psychiatric,
  title={Psychiatric-mental health nursing},
  author={Videbeck, Sheila L},
  year={2010},
  publisher={Lippincott Williams \& Wilkins}
}

@article{liu2024teaching,
  title={Teaching CS50 with AI},
  author={Liu, Rongxin and Zenke, Carter and Liu, Charlie and Holmes, Andrew and Thornton, Patrick and Malan, David J},
  journal={Portland, OR, US: ACM},
  year={2024}
}

@article{vogel2008assessment,
  title={Assessment of Campus Climate to Enhance Student Success.},
  author={Vogel, Susan A and Holt, Janet K and Sligar, Steven and Leake, Elizabeth},
  journal={Journal of Postsecondary Education and Disability},
  volume={21},
  number={1},
  pages={15--31},
  year={2008},
  publisher={ERIC}
}

@book{dillman2014internet,
  title={Internet, phone, mail, and mixed-mode surveys: The tailored design method},
  author={Dillman, Don A and Smyth, Jolene D and Christian, Leah Melani},
  year={2014},
  publisher={John Wiley \& Sons}
}

@article{galesic2009effects,
  title={Effects of questionnaire length on participation and indicators of response quality in a web survey},
  author={Galesic, Mirta and Bosnjak, Michael},
  journal={Public opinion quarterly},
  volume={73},
  number={2},
  pages={349--360},
  year={2009},
  publisher={Oxford University Press}
}

@article{sahlqvist2011effect,
  title={Effect of questionnaire length, personalisation and reminder type on response rate to a complex postal survey: randomised controlled trial},
  author={Sahlqvist, Shannon and Song, Yena and Bull, Fiona and Adams, Emma and Preston, John and Ogilvie, David and iConnect Consortium},
  journal={BMC medical research methodology},
  volume={11},
  pages={1--8},
  year={2011},
  publisher={Springer}
}

@article{porter2005mail,
  title={E-mail subject lines and their effect on web survey viewing and response},
  author={Porter, Stephen R and Whitcomb, Michael E},
  journal={Social Science Computer Review},
  volume={23},
  number={3},
  pages={380--387},
  year={2005},
  publisher={Sage Publications Sage CA: Thousand Oaks, CA}
}

@inproceedings{abbas2021university,
  title={University student surveys using chatbots: artificial intelligence conversational agents},
  author={Abbas, Noorhan and Pickard, Thomas and Atwell, Eric and Walker, Aisha},
  booktitle={International Conference on Human-Computer Interaction},
  pages={155--169},
  year={2021},
  organization={Springer}
}

@article{abbas2022online,
  title={Online chat and chatbots to enhance mature student engagement in higher education},
  author={Abbas, Noorhan and Whitfield, Jenni and Atwell, Eric and Bowman, Helen and Pickard, Thomas and Walker, Aisha},
  journal={International Journal of Lifelong Education},
  volume={41},
  number={3},
  pages={308--326},
  year={2022},
  publisher={Taylor \& Francis}
}

@inproceedings{martinez2024ai,
  title={AI-driven student assistance: chatbots redefining university support},
  author={Martinez-Requejo, S and Garc{\'\i}a, E Jimenez and Duarte, S Redondo and L{\'a}zaro, J Ruiz and Sanz, E Puertas and Vivas, G Mariscal},
  booktitle={INTED2024 Proceedings},
  pages={617--625},
  year={2024},
  organization={IATED}
}

@article{tanwar2024opinebot,
  title={Opinebot: Class feedback reimagined using a conversational llm},
  author={Tanwar, Henansh and Shrivastva, Kunal and Singh, Rahul and Kumar, Dhruv},
  journal={arXiv preprint arXiv:2401.15589},
  year={2024}
}

@article{xiao2020tell,
  title={Tell me about yourself: Using an AI-powered chatbot to conduct conversational surveys with open-ended questions},
  author={Xiao, Ziang and Zhou, Michelle X and Liao, Q Vera and Mark, Gloria and Chi, Changyan and Chen, Wenxi and Yang, Huahai},
  journal={ACM Transactions on Computer-Human Interaction (TOCHI)},
  volume={27},
  number={3},
  pages={1--37},
  year={2020},
  publisher={ACM New York, NY, USA}
}

@article{belhaj2021engaging,
  title={Engaging students to fill surveys using chatbots: University case study},
  author={Belhaj, Nadir and Hamdane, Abdemounaime and Chaoui, N El Houda and Chaoui, Houda and El Bekkali, Moulhime},
  journal={Indones. J. Electr. Eng. Comput. Sci},
  volume={24},
  number={1},
  pages={473--483},
  year={2021}
}

@article{zarouali2024comparing,
  title={Comparing chatbots and online surveys for (longitudinal) data collection: an investigation of response characteristics, data quality, and user evaluation},
  author={Zarouali, Brahim and Araujo, Theo and Ohme, Jakob and de Vreese, Claes},
  journal={Communication Methods and Measures},
  volume={18},
  number={1},
  pages={72--91},
  year={2024},
  publisher={Taylor \& Francis}
}

@inproceedings{njeguvs2021conversational,
  title={Conversational Survey Chatbot: User Experience and Perception},
  author={Njegu{\v{s}}, Angelina and others},
  booktitle={Sinteza 2021-International Scientific Conference on Information Technology and Data Related Research},
  pages={322--327},
  year={2021},
  organization={Singidunum University}
}

@inproceedings{kim2019comparing,
  title={Comparing data from chatbot and web surveys: Effects of platform and conversational style on survey response quality},
  author={Kim, Soomin and Lee, Joonhwan and Gweon, Gahgene},
  booktitle={Proceedings of the 2019 CHI conference on human factors in computing systems},
  pages={1--12},
  year={2019}
}

@article{lenzner2010cognitive,
  title={Cognitive burden of survey questions and response times: A psycholinguistic experiment},
  author={Lenzner, Timo and Kaczmirek, Lars and Lenzner, Alwine},
  journal={Applied cognitive psychology},
  volume={24},
  number={7},
  pages={1003--1020},
  year={2010},
  publisher={Wiley Online Library}
}

@article{kocaballi2019personalization,
  title={The personalization of conversational agents in health care: systematic review},
  author={Kocaballi, Ahmet Baki and Berkovsky, Shlomo and Quiroz, Juan C and Laranjo, Liliana and Tong, Huong Ly and Rezazadegan, Dana and Briatore, Agustina and Coiera, Enrico},
  journal={Journal of medical Internet research},
  volume={21},
  number={11},
  pages={e15360},
  year={2019},
  publisher={JMIR Publications Toronto, Canada}
}

@article{chen2024recent,
  title={Recent Trends in Personalized Dialogue Generation: A Review of Datasets, Methodologies, and Evaluations},
  author={Chen, Yi-Pei and Nishida, Noriki and Nakayama, Hideki and Matsumoto, Yuji},
  journal={arXiv preprint arXiv:2405.17974},
  year={2024}
}

@misc{reghunath2021expression,
  title={Expression of empathy in social virtual bots used for genetic counseling},
  author={Reghunath, Anagha},
  year={2021}
}

@article{pitts2007improving,
  title={Improving requirements elicitation: an empirical investigation of procedural prompts},
  author={Pitts, Mitzi G and Browne, Glenn J},
  journal={Information systems journal},
  volume={17},
  number={1},
  pages={89--110},
  year={2007},
  publisher={Wiley Online Library}
}

@article{zhou2025building,
  title={Building citizen trust to enhance satisfaction in digital public services: the role of empathetic chatbot communication},
  author={Zhou, Mingxi and Liu, Luning and Feng, Yuqiang},
  journal={Behaviour \& Information Technology},
  pages={1--20},
  year={2025},
  publisher={Taylor \& Francis}
}

@article{bai2019systematic,
  title={A systematic review of emoji: Current research and future perspectives},
  author={Bai, Qiyu and Dan, Qi and Mu, Zhe and Yang, Maokun},
  journal={Frontiers in psychology},
  volume={10},
  pages={2221},
  year={2019},
  publisher={Frontiers Media SA}
}

@inproceedings{thorat2020review,
  title={A review on implementation issues of rule-based chatbot systems},
  author={Thorat, Sandeep A and Jadhav, Vishakha},
  booktitle={Proceedings of the international conference on innovative computing \& communications (ICICC)},
  year={2020}
}

@inproceedings{maeng2021designing,
  title={Designing a chatbot for survivors of sexual violence: Exploratory study for hybrid approach combining rule-based chatbot and ml-based chatbot},
  author={Maeng, Wookjae and Lee, Joonhwan},
  booktitle={Proceedings of the Asian CHI Symposium 2021},
  pages={160--166},
  year={2021}
}

@article{melville2016conducting,
  title={Conducting sensitive interviews: A review of reflections},
  author={Melville, Angela and Hincks, Darren},
  journal={Law and Method},
  volume={1},
  number={1},
  pages={1--26},
  year={2016},
  publisher={Boom Juridisch}
}

@article{wheeless1977measurement,
  title={The measurement of trust and its relationship to self-disclosure},
  author={Wheeless, Lawrence R and Grotz, Janis},
  journal={Human Communication Research},
  volume={3},
  number={3},
  pages={250--257},
  year={1977},
  publisher={Wiley Online Library}
}

@article{kim2008keeping,
  title={Keeping users at the center: Developing a multimedia interface for informed consent},
  author={Kim, Loel and Young, Amanda J and Neimeyer, Robert A and Baker, Justin N and Barfield, Raymond C},
  journal={Technical Communication Quarterly},
  volume={17},
  number={3},
  pages={335--357},
  year={2008},
  publisher={Taylor \& Francis}
}

@article{chan2022challenges,
  title={The challenges in designing a prevention chatbot for eating disorders: observational study},
  author={Chan, William W and Fitzsimmons-Craft, Ellen E and Smith, Arielle C and Firebaugh, Marie-Laure and Fowler, Lauren A and DePietro, Bianca and Topooco, Naira and Wilfley, Denise E and Taylor, C Barr and Jacobson, Nicholas C},
  journal={JMIR Formative Research},
  volume={6},
  number={1},
  pages={e28003},
  year={2022},
  publisher={JMIR Publications Inc., Toronto, Canada}
}

@inproceedings{tang2024advancing,
  title={Advancing mental health pre-screening: A new custom gpt for psychological distress assessment},
  author={Tang, Jinwen and Shang, Yi},
  booktitle={2024 IEEE 6th International Conference on Cognitive Machine Intelligence (CogMI)},
  pages={162--171},
  year={2024},
  organization={IEEE}
}

@inproceedings{tang2024decoding,
  title={Decoding Linguistic Nuances in Mental Health Text Classification Using Expressive Narrative Stories},
  author={Tang, Jinwen and Guo, Qiming and Zhao, Yunxin and Shang, Yi},
  booktitle={2024 IEEE 6th International Conference on Cognitive Machine Intelligence (CogMI)},
  pages={207--216},
  year={2024},
  organization={IEEE}
}

@INPROCEEDINGS{tang2025layered,
  author={Tang, Jinwen and Guo, Qiming and Sun, Wenbo and Shang, Yi},
  booktitle={2025 IEEE Conference on Artificial Intelligence (CAI)}, 
  title={A Layered Multi-Expert Framework for Long-Context Mental Health Assessments}, 
  year={2025},
  volume={},
  number={},
  pages={435-440},
  doi={10.1109/CAI64502.2025.00080}}

@article{blanco2008mental,
  title={Mental health of college students and their non--college-attending peers: results from the national epidemiologic study on alcohol and related conditions},
  author={Blanco, Carlos and Okuda, Mayumi and Wright, Crystal and Hasin, Deborah S and Grant, Bridget F and Liu, Shang-Min and Olfson, Mark},
  journal={Archives of general psychiatry},
  volume={65},
  number={12},
  pages={1429--1437},
  year={2008},
  publisher={American Medical Association}
}

@article{song2024combining,
  title={Combining Hierachical VAEs with LLMs for clinically meaningful timeline summarisation in social media},
  author={Song, Jiayu and Chim, Jenny and Tsakalidis, Adam and Ive, Julia and Atzil-Slonim, Dana and Liakata, Maria},
  journal={arXiv preprint arXiv:2401.16240},
  year={2024}
}

@article{soman2025human,
  title={Human guided empathetic AI agent for mental health support leveraging reinforcement learning-enhanced retrieval-augmented generation},
  author={Soman, Gayathri and Judy, MV and Abou, Aadhil Muhammad},
  journal={Cognitive Systems Research},
  volume={90},
  pages={101337},
  year={2025},
  publisher={Elsevier}
}

@article{garg2023exploring,
  title={Exploring the role of ChatGPT in patient care (diagnosis and treatment) and medical research: A systematic review},
  author={Garg, Ravindra Kumar and Urs, Vijeth L and Agarwal, Akshay Anand and Chaudhary, Sarvesh Kumar and Paliwal, Vimal and Kar, Sujita Kumar},
  journal={Health Promotion Perspectives},
  volume={13},
  number={3},
  pages={183},
  year={2023}
}

@article{van2023clinical,
  title={Clinical text summarization: adapting large language models can outperform human experts},
  author={Van Veen, Dave and Van Uden, Cara and Blankemeier, Louis and Delbrouck, Jean-Benoit and Aali, Asad and Bluethgen, Christian and Pareek, Anuj and Polacin, Malgorzata and Reis, Eduardo Pontes and Seehofnerova, Anna and others},
  journal={Research square},
  pages={rs--3},
  year={2023}
}

@article{singhal2025toward,
  title={Toward expert-level medical question answering with large language models},
  author={Singhal, Karan and Tu, Tao and Gottweis, Juraj and Sayres, Rory and Wulczyn, Ellery and Amin, Mohamed and Hou, Le and Clark, Kevin and Pfohl, Stephen R and Cole-Lewis, Heather and others},
  journal={Nature Medicine},
  pages={1--8},
  year={2025},
  publisher={Nature Publishing Group US New York}
}

@article{guo2024large,
  title={Large language models for mental health applications: Systematic review},
  author={Guo, Zhijun and Lai, Alvina and Thygesen, Johan H and Farrington, Joseph and Keen, Thomas and Li, Kezhi and others},
  journal={JMIR mental health},
  volume={11},
  number={1},
  pages={e57400},
  year={2024},
  publisher={JMIR Publications Inc., Toronto, Canada}
}

@article{guo2024soullmate,
  title={SouLLMate: An Application Enhancing Diverse Mental Health Support with Adaptive LLMs, Prompt Engineering, and RAG Techniques},
  author={Guo, Qiming and Tang, Jinwen and Sun, Wenbo and Tang, Haoteng and Shang, Yi and Wang, Wenlu},
  journal={arXiv preprint arXiv:2410.16322},
  year={2024}
}

@article{guo2024soullmate2,
  title={Soullmate: An adaptive llm-driven system for advanced mental health support and assessment, based on a systematic application survey},
  author={Guo, Qiming and Tang, Jinwen and Sun, Wenbo and Tang, Haoteng and Shang, Yi and Wang, Wenlu},
  journal={arXiv preprint arXiv:2410.11859},
  year={2024}
}

@article{edition2013diagnostic,
  title={Diagnostic and statistical manual of mental disorders},
  author={Edition, Fifth and others},
  journal={Am Psychiatric Assoc},
  volume={21},
  number={21},
  pages={591--643},
  year={2013}
}

@article{Kroenke2009PHQ8,
  author    = {Kroenke, Kurt and Strine, Tara W. and Spitzer, Robert L. 
               and Williams, Janet B. W. and Berry, Joyce T. and Mokdad, Ali H.},
  title     = {{The PHQ-8 as a measure of current depression in the general population}},
  journal   = {Journal of Affective Disorders},
  volume    = {114},
  number    = {1--3},
  pages     = {163--173},
  year      = {2009},
  doi       = {10.1016/j.jad.2008.06.026}
}

@inproceedings{tang2025tigergpt,
  title={TigerGPT: A Theory-Driven AI Chatbot for Adaptive Campus Climate Surveys},
  author={Tang, Jinwen and Chen, Songxi and Shang, Yi},
  booktitle={2025 IEEE International Conference on Future Machine Learning and Data Science (FMLDS)},
  pages={668--673},
  year={2025},
  organization={IEEE}
}

@article{tang2025aura,
  title={AURA: A Reinforcement Learning Framework for AI-Driven Adaptive Conversational Surveys},
  author={Tang, Jinwen and Shang, Yi},
  journal={arXiv preprint arXiv:2510.27126},
  year={2025}
}

@inproceedings{li2016deep,
  title={Deep reinforcement learning for dialogue generation},
  author={Li, Jiwei and Monroe, Will and Ritter, Alan and Jurafsky, Dan},
  booktitle={Proceedings of the 2016 Conference on Empirical Methods in Natural Language Processing (EMNLP)},
  pages={1192--1202},
  year={2016},
  organization={ACL}
}

@inproceedings{su2021dynamic,
  title={Dynamic policy networks for task-oriented dialogue with reinforcement learning},
  author={Su, Pei-Hao and Gasic, Milica and Young, Steve and Gašić, Milica and Young, Steve},
  booktitle={Proceedings of the 59th Annual Meeting of the Association for Computational Linguistics (ACL)},
  pages={341--352},
  year={2021},
  organization={ACL}
}

@article{sun2023contextual,
  title={Contextual bandits for adaptive dialogue management},
  author={Sun, Yichao and Li, Xin and Zhou, Ke and Wang, Xiaodong},
  journal={IEEE/ACM Transactions on Audio, Speech, and Language Processing},
  volume={31},
  pages={1152--1163},
  year={2023},
  publisher={IEEE}
}

@book{sutton1998reinforcement,
  title={Reinforcement learning: An introduction},
  author={Sutton, Richard S and Barto, Andrew G and others},
  volume={1},
  number={1},
  year={1998},
  publisher={MIT press Cambridge}
}

@article{brysbaert2014concreteness,
  title={Concreteness ratings for 40 thousand generally known English word lemmas},
  author={Brysbaert, Marc and Warriner, Amy Beth and Kuperman, Victor},
  journal={Behavior research methods},
  volume={46},
  number={3},
  pages={904--911},
  year={2014},
  publisher={Springer}
}

@article{newman2003lying,
  title={Lying words: Predicting deception from linguistic styles},
  author={Newman, Matthew L and Pennebaker, James W and Berry, Diane S and Richards, Jane M},
  journal={Personality and social psychology bulletin},
  volume={29},
  number={5},
  pages={665--675},
  year={2003},
  publisher={Sage Publications}
}

@article{levine2002autobiographical,
  title={The functional neuroanatomy of episodic and semantic autobiographical remembering: a prospective functional MRI study},
  author={Levine, Brian and Turner, Gary R and Tisserand, Daniel and Hevenor, Stephen J and Graham, Susan J and McIntosh, Anthony R},
  journal={Journal of Cognitive Neuroscience},
  volume={14},
  number={6},
  pages={668--676},
  year={2002},
  publisher={MIT Press}
}

@article{holland2009measuring,
  title={Measuring response quality in open-ended questions: an 
         application to the American Community Survey},
  author={Holland, Jennifer L and Christian, Leah Melani},
  journal={Field Methods},
  volume={21},
  number={4},
  pages={373--390},
  year={2009},
  publisher={SAGE Publications}
}

@article{pennebaker2003linguistic,
  title={Linguistic styles: language use as an individual difference},
  author={Pennebaker, James W and Mehl, Matthias R and Niederhoffer, Kate G},
  journal={Journal of Personality and Social Psychology},
  volume={77},
  number={6},
  pages={1296--1312},
  year={2003},
  publisher={American Psychological Association}
}

@article{russell1980circumplex,
  title={A circumplex model of affect},
  author={Russell, James A},
  journal={Journal of Personality and Social Psychology},
  volume={39},
  number={6},
  pages={1161--1178},
  year={1980},
  publisher={American Psychological Association}
}

@inproceedings{hutto2014vader,
  title={VADER: A parsimonious rule-based model for sentiment analysis of 
         social media text},
  author={Hutto, Clayton J and Gilbert, Eric},
  booktitle={Proceedings of the 8th International Conference on Weblogs 
             and Social Media (ICWSM)},
  pages={216--225},
  year={2014}
}

@book{tourangeau2000psychology,
  title={The psychology of survey response},
  author={Tourangeau, Roger and Rips, Lance J and Rasinski, Kenneth},
  year={2000},
  publisher={Cambridge University Press}
}

@article{pennebaker1999linguistic,
  title={Linguistic styles: Language use as an individual difference},
  author={Pennebaker, James W and King, Laura A},
  journal={Journal of Personality and Social Psychology},
  volume={77},
  number={6},
  pages={1296--1312},
  year={1999},
  publisher={American Psychological Association}
}

@book{gudykunst1988culture,
  title={Culture and interpersonal communication},
  author={Gudykunst, William B and Ting-Toomey, Stella},
  year={1988},
  publisher={Sage Publications}
}

@article{warriner2013norms,
  title={Norms of valence, arousal, and dominance for 13,915 English lemmas},
  author={Warriner, Amy Beth and Kuperman, Victor and Brysbaert, Marc},
  journal={Behavior Research Methods},
  volume={45},
  number={4},
  pages={1191--1207},
  year={2013},
  publisher={Springer},
  doi={10.3758/s13428-012-0314-x}
}

@book{giles1991contexts,
  title={Contexts of Accommodation: Developments in Applied Sociolinguistics},
  author={Giles, Howard and Coupland, Justine and Coupland, Nikolas},
  year={1991},
  publisher={Cambridge University Press}
}

@incollection{searle1976classification,
  title={A classification of illocutionary acts},
  author={Searle, John R},
  booktitle={Language in Society},
  volume={5},
  number={1},
  pages={1--23},
  year={1976},
  publisher={Cambridge University Press}
}

\chapter*{VITA}
\addcontentsline{toc}{chapter}{VITA}

Jinwen Tang received dual Bachelor of Science degrees in Statistics and Psychology (Magna Cum Laude) from Iowa State University, USA, in 2015. She pursued interdisciplinary graduate studies at the University of Missouri-Columbia, earning a Master of Science in Computer Science, a Master of Arts in Statistics, and a Master of Arts in Psychology. She completed her Doctor of Philosophy in Computer Science in the Department of Electrical Engineering and Computer Science at the University of Missouri-Columbia in 2025. Her research interests include natural language processing, large language models, reinforcement learning, and AI for public well-being. During her doctoral study, she has published 8 peer-reviewed papers, including 5 as first author. She was the recipient of the Outstanding Doctoral Students Award from the College of Engineering at the University of Missouri in 2026.

\end{document}